\documentclass[5p,number,sort&compress,times,final]{elsarticle} % final settings
\usepackage[utf8]{inputenc}
\usepackage[hyphens,spaces,obeyspaces]{url}
\usepackage[hidelinks]{hyperref}
\usepackage{amsmath}
\usepackage[export]{adjustbox}
\usepackage{xcolor}
\usepackage{calc}
\newcommand{\beginappendix}{%
	\setcounter{table}{0}
	\renewcommand{\thetable}{A.\arabic{table}}%
	\setcounter{figure}{0}
	\renewcommand{\thefigure}{A.\arabic{figure}}%
}
\usepackage[binary-units]{siunitx}
\usepackage[
	textsize=footnotesize,
	colorinlistoftodos,
	obeyFinal
	]{todonotes}
	
\presetkeys{todonotes}{inline,backgroundcolor=green!50!white}{}

\usepackage{nameref}
\makeatletter
\newcommand*{\thissectiontitle}{\@currentlabelname}
\makeatother

\usepackage[
	acronym,
	nomain,
	nogroupskip,
	symbols,
	nonumberlist,
	nopostdot,
	automake,
	style=super]{glossaries} 
\usepackage{glossary-mcols}
\newglossary[slg]{symbolslist}{syi}{syg}{Symbols}

\makeglossaries
\usepackage[capitalize,noabbrev]{cleveref}

\usepackage{nameref,zref-xr}
\zxrsetup{toltxlabel}
\usepackage{float}
\usepackage{placeins}
\usepackage{wrapfig}
\usepackage[figuresleft]{rotating}
\usepackage{graphicx}
\graphicspath{{./figs/}}

\usepackage{caption}
\DeclareCaptionLabelFormat{continued}{#1~#2 (continuation)}
\usepackage[skip=1mm]{subcaption}

\newlength{\figurewidth}
\newlength{\imagewidth}
\newlength{\spacerwidth}

\newcount\nImageColumns
\newlength{\firsttablecolumnwidth}
\usepackage{array}
\usepackage{booktabs}
\usepackage{multirow}
\usepackage{threeparttable}

\usepackage{tabularx}
\newcolumntype{Y}{>{\centering\bfseries\arraybackslash}X}
\newcolumntype{Z}{>{\centering\arraybackslash}X}
\newcolumntype{L}{>{\raggedright\arraybackslash}X}
\newcolumntype{R}{>{\raggedleft\arraybackslash}X}
\usepackage{graphicx}
\usepackage{calc}
\usepackage[export]{adjustbox}
\newlength{\depthofsumsign}
\newcommand % \scalebar{magnification in kx}{imagewidth}{x in nm}
{
	\scalebar
}[3]{
	\hfill
	\adjustbox
	{
		margin=0 1pt 0 1pt,bgcolor=lightgray,fgcolor=black,margin*=0 0.5pt 0 0
	}{
		\parbox
		{
			{#2}/120750*{#3}*{#1}
		}{
			\centering \SI{#3}{\nm}
		}
	}
}

\newacronym{ANN}{ANN}{artificial neural network}
\newacronym{PSD}{PSD}{particle size distribution}
\newacronym{DPN}{DPN}{DeepParticleNet}
\newacronym{GPU}{GPU}{graphics processing unit}
\newacronym{CPU}{CPU}{central processing unit}
\newacronym{CNN}{CNN}{convolutional neural network}
\newacronym{SEM}{SEM}{scanning electron microscopy}
\newacronym{TEM}{TEM}{transmission electron microscopy}
\newacronym{ReLU}{ReLU}{rectified linear unit}
\newacronym[plural=ROI,firstplural=regions of interest (ROI)]{ROI}{ROI}{region of interest}
\newacronym{RPN}{RPN}{region proposal network}
\newacronym{UE}{UE}{ultimate erosion}
\newacronym{HT}{HT}{Hough transformation}
\newacronym{WST}{WST}{watershed transformation}
\newacronym{synthPIC}{synthPIC}{synthetic particle image creator}
\newacronym{PS}{PS}{ParticleSizer}
\newacronym{MAPE}{MAPE}{mean absolute percentage error}
\newacronym[first=\glstext{SGDM}]{SGDM}{SGDM}{stochastic gradient descent with momentum}
\newacronym{COCO}{COCO}{common objects in context}
\newglossaryentry{symb:loss_val}{
	name=\ensuremath{\mathcal{L}_\text{validation}},
	description={validation loss},
	sort=Lvalidation, type=symbolslist
}

\newglossaryentry{symb:loss_tr}{
	name=\ensuremath{\mathcal{L}_\text{training}},
	description={training loss},
	sort=Ltraining, type=symbolslist
}

\newglossaryentry{symb:lr}{
	name=\ensuremath{\alpha},
	description={learning rate},
	sort=a, type=symbolslist
}

\newglossaryentry{symb:lr_min}{
	name=\ensuremath{\alpha_\text{min}},
	description={minimum optimal learning rate},
	sort=amin, type=symbolslist
}

\newglossaryentry{symb:lr_max}{
	name=\ensuremath{\alpha_\text{max}},
	description={maximum optimal learning rate},
	sort=amax, type=symbolslist
}

\newglossaryentry{symb:standard_deviation}{
	name=\ensuremath{\sigma},
	description={standard deviation},
	sort=s, type=symbolslist
}

\newglossaryentry{symb:kullback_leibler_divergence}{
	name=\ensuremath{D_\text{KL}},
	description={Kullback-Leibler divergence},
	sort=DKL, type=symbolslist
}

\newglossaryentry{symb:probability_distribution_p}{
	name=\ensuremath{P},
	description={probability distribution},
	sort=P, type=symbolslist
}

\newglossaryentry{symb:probability_distribution_q}{
	name=\ensuremath{Q_i},
	description={probability distribution},
	sort=Q_i, type=symbolslist
}

\newglossaryentry{symb:feret_diameter}{
	name=\ensuremath{d_\text{Feret}},
	description={Feret diameter},
	sort=dFeret, type=symbolslist
}

\newglossaryentry{symb:geometric_mean_diameter}{
	name=\ensuremath{d_\text{g}},
	description={geometric mean diameter},
	sort=dg, type=symbolslist
}

\newglossaryentry{symb:geometric_standard_deviation}{
	name=\ensuremath{\sigma_\text{g}},
	description={geometric standard deviation},
	sort=sg, type=symbolslist
}

\newglossaryentry{symb:solidity}{
	name=\ensuremath{S},
	description={solidity},
	sort=S, type=symbolslist
}

\newglossaryentry{symb:area}{
	name=\ensuremath{A},
	description={area},
	sort=A, type=symbolslist
}

\newglossaryentry{symb:area_convex}{
	name=\ensuremath{A_\text{convex}},
	description={convex area},
	sort=Aconvex, type=symbolslist
}

\newglossaryentry{symb:number_primary_particles}{
	name=\ensuremath{N},
	description={number of primary particles},
	sort=N, type=symbolslist
}

\newglossaryentry{symb:number_dates}{
	name=\ensuremath{n},
	description={number of dates},
	sort=n, type=symbolslist
}

\newglossaryentry{symb:temperature_sinter}{
	name=\ensuremath{T_\text{sinter}},
	description={sintering temperature},
	sort=Ts, type=symbolslist
}

\newglossaryentry{symb:MAPE}{
	name=\ensuremath{\text{MAPE}},
	description={mean absolute percentage error},
	sort=MAPE, type=symbolslist
}

\newglossaryentry{symb:geometric_mean_diameter_percentage_error}{
	name=\ensuremath{\Delta d_\text{g,\%}},
	description={percentage error of the geometric mean diameter},
	sort=Deltadgpercentage, type=symbolslist
}

\newglossaryentry{symb:geometric_standard_deviation_percentage_error}{
	name=\ensuremath{\Delta \sigma_\text{g,\%}},
	description={percentage error of the geometric standard deviation},
	sort=Deltasigmagpercentage, type=symbolslist
}

\newglossaryentry{symb:number_primary_particles_percentage_error}{
	name=\ensuremath{\Delta N_\%},
	description={percentage error of the number of primary particles},
	sort=DeltaNpercentage, type=symbolslist
}

\newglossaryentry{symb:percentage_error}{
	name=\ensuremath{\Delta X_{\%,i}},
	description={percentage error},
	sort=DeltaXpercentagei, type=symbolslist
}

\newglossaryentry{symb:desired_value}{
	name=\ensuremath{T_i},
	description={desired value},
	sort=Ti, type=symbolslist
}

\newglossaryentry{symb:actual_value}{
	name=\ensuremath{X_i},
	description={actual value},
	sort=Xi, type=symbolslist
}

\newglossaryentry{symb:index}{
	name=\ensuremath{i},
	description={index},
	sort=i, type=symbolslist
}

\newglossaryentry{symb:training_set_size}{
	name=\ensuremath{n_\text{training}},
	description={number of training samples},
	sort=ntraining, type=symbolslist
}

\newglossaryentry{symb:total_training_time}{
	name=\ensuremath{t_\text{training}},
	description={total training time},
	sort=ttraining, type=symbolslist
}

\newglossaryentry{symb:average_precision}{
	name=\ensuremath{\text{AP}},
	description={average precision},
	sort=ap, type=symbolslist
}

\newglossaryentry{symb:intersection_over_union}{
	name=\ensuremath{\text{IoU}},
	description={intersection over union},
	sort=iou, type=symbolslist
}

\DeclareSIUnit\px{px}

\begin{document}
	
	%%%%%%%%%%%%%%%%%%%%%%%%%%%%%%%%%%%%%%%%%%%%%%%%%%%%%%
	% Frontmatter
	%%%%%%%%%%%%%%%%%%%%%%%%%%%%%%%%%%%%%%%%%%%%%%%%%%%%%%
	\begin{frontmatter}
		\journal{% !TeX spellcheck = en_US
Powder Technology}	 	 
		\title{% !TeX spellcheck = en_US
Image-Based Size Analysis of Agglomerated and Partially Sintered Particles via Convolutional Neural Networks}	
	 	 
		% !TeX spellcheck = en_US
\author{M. Frei\corref{mycorrespondingauthor}}
\cortext[mycorrespondingauthor]{Corresponding author. Tel.: +49 203 379--3621}
\ead{max.frei@uni-due.de}

\author{F. E. Kruis\corref{}}

\address{Institute of Technology for Nanostructures (NST) and Center for Nanointegration Duisburg-Essen (CENIDE)\\University of Duisburg-Essen, Duisburg, D-47057, Germany}
	 	 
		\begin{abstract}
			% !TeX spellcheck = en_US
There is a high demand for fully automated methods for the analysis of primary particle size distributions of agglomerated, sintered or occluded primary particles, due to their impact on material properties. Therefore, a novel, deep learning-based, method for the detection of such primary particles was proposed and tested, which renders a manual tuning of analysis parameters unnecessary.

As a specialty, the training of the utilized convolutional neural networks was carried out using only synthetic images, thereby avoiding the laborious task of manual annotation and increasing the ground truth quality. Nevertheless, the proposed method performs excellent on real world samples of sintered silica nanoparticles with various sintering degrees and varying image conditions.

In a direct comparison, the proposed method clearly outperforms two state-of-the-art methods for automated image-based particle size analysis (Hough transformation and the ImageJ ParticleSizer plug-in), thereby attaining human-like performance.

		\end{abstract}

		\begin{keyword}
			% !TeX spellcheck = en_US
imaging particle size analysis \sep 
agglomerate \sep 
convolutional neural network~(CNN) \sep 
Mask R-CNN \sep 
Hough transformation \sep 
ImageJ ParticleSizer
		\end{keyword}
	\end{frontmatter}

	%%%%%%%%%%%%%%%%%%%%%%%%%%%%%%%%%%%%%%%%%%%%%%%%%%%%%%
	% Mainmatter
	%%%%%%%%%%%%%%%%%%%%%%%%%%%%%%%%%%%%%%%%%%%%%%%%%%%%%%
	
	%\listoftodos

	% !TeX spellcheck = en_US
\section{Introduction}
Powders play an important role in chemical industry. In Europe, approximately \SI{60}{\percent} of the products of this branch of industry are powders themselves and another \SI{20}{\percent} of the products require powders during their production \cite{Schulze.2014}. Accordingly, particle measurement technologies, e.g. for the determination of \glspl{PSD}, are of vital importance for the chemical industry.

The distinction of objects on images and the image background, a process referred to as image segmentation, is an important requirement for imaging particle analysis techniques. The simplest way to do so is to create a binary image, where black pixels represent the image background, while white pixels represent the sought-after objects. Subsequently, the sizes of the objects can be attained by determining the numbers of pixels of connected white regions. It is plain to see that overlapping or occluded objects, for instance caused by agglomeration, are a major source of error for this approach, because they impede the detection of the individual objects. This is a severe problem, because the detection of overlapping and occluded objects plays an important role in many applications, such as the determination of the primary \gls{PSD} of agglomerated particles on microscopic images \cite{Schneider.2015} as well as bubbles in multi-phase reactors \cite{Rodrigues.2003} or the detection and characterization of fibers \cite{Azari.2014}. 

Often, the solutions for such problems are either highly customized and thus difficult to reuse and adapt or extremely laborious (e.g. manual analysis) and therefore expensive.

While humans are very good at recognizing overlapping objects as individual objects, this task is usually much harder for algorithms, because they lack the flexibility to adapt to the specific image conditions. Therefore, we propose the utilization of \glspl{ANN} and more specifically a \gls{CNN}, we named \gls{DPN}, which can autonomously learn to detect individual objects based on a large number (usually a few dozens or hundreds) of already annotated images via supervised learning. In case of a necessary adaption to new imaging conditions it suffices to adjust the training data and to retrain the \gls{DPN}.

The fact that the \gls{DPN} heavily depends on training data is one of its major advantages, because it minimizes the amount of a priori knowledge which is necessary to achieve a reliable object detection. However, at the same time, it is also its largest disadvantage because already evaluated training data is often not available and has to be created manually, which is a very laborious task. Therefore, we propose the synthesis of the images and ground truths needed for the training. Prior to the image synthesis, the characteristics of real images are analyzed and used to yield lifelike synthetic images with known ground truths, thereby making a manual annotation of images obsolete. A certain degree of realism of the synthesized images is very important to guarantee a realistic training of the \gls{DPN}. Otherwise, the application to real images might yield insufficient results.

The combination of a self-learning method and a customized synthesis of its training data allow a high degree of specialization, while still maintaining high levels of adaptability and robustness.

\subsection{Automated Image-Based Particle Sizing Methods}

In the context of this work, it is sensible to categorize established and innovative methods for the determination of the \glspl{PSD} of agglomerated and occluded particles into three groups: conventional, \gls{ANN}-based and \gls{CNN}-based methods. 

\paragraph{Conventional Methods} Conventional methods utilize classic image processing techniques for the identification of the individual primary particles. Among the most popular of these methods are the \gls{WST} \cite{Cheng.2009,Jung.2010,Shu.2013}, \gls{UE} \cite{Park.2013,Wang.2016} and the \gls{HT} \cite{Kruis.1994,Ballard.1981,Merlin.1975,Xu.1990}. \gls{WST}- and \gls{UE}-based methods are usually fast and feature a small memory footprint. However, they often tend to over-segment images, due to their noise susceptibility. In contrast, \gls{HT}-based methods are considerably slower and consume much more memory. In return, they are more robust towards noise and other image distortions. One major disadvantage that conventional methods share is the fact that they usually feature one or more parameters, which need to be set by the user according to the imaging conditions, to achieve a high precision. Therefore, a change in the imaging conditions, e.g. due to an operator change, may result in faulty analysis results and the necessity to retune the parameters.

\paragraph{\glsentryshort{ANN}-based Methods} In contrast to conventional methods, \gls{ANN}-based methods \cite{Frei.2018,Hamzeloo.2014,Ko.2011} can help to avoid image specific parameters. They are therefore usually fully automatic and more robust to changes of the imaging conditions. However, they often use shape descriptors as inputs, which have to be identified and implemented during the design of the \glspl{ANN}. Apart from the increased design-effort, the use of shape descriptors also leads to a severe loss of information. While, on the one hand, this reduction is necessary to make the use of \glspl{ANN} feasible in the first place, it can also severely impede the detection accuracy.

\paragraph{\glsentryshort{CNN}-based Methods} The main advantage of \glspl{CNN} for image-based particle analysis is the fact that, unlike \glspl{ANN}, they can not only be trained on feature interpretation but also feature extraction. \glspl{CNN}-based methods are therefore often referred to as end-to-end methods. They take the raw data, i.e. images, as input and output the sought-after measurands \cite{Goodfellow.2016}. Current \gls{CNN}-based particle analysis methods focus either on the shape-discrimination of particles, e.g. agglomerate/non-agglomerate \cite{Mehle.2017} or on the identification and subsequent \gls{PSD} measurement of single particles \cite{Heimowitz.2018}. In contrast, within this publication, a method for the \gls{CNN}-based \gls{PSD} measurement of agglomerated and even partially sintered primary particles will be presented. 
	% !TeX spellcheck = en_US
\section{Theory of Convolutional Neural Networks}
The proposed method is based on \glspl{CNN}, which are a special form of \glspl{ANN}. Therefore, the fundamental principles of \glspl{CNN} shall be explained briefly within the following sections. For a comprehensive introduction to \glspl{ANN} and \glspl{CNN}, please refer to Kriesel \cite{Kriesel.2007} and Goodfellow et al. \cite{Goodfellow.2016}, respectively. Additionally, a very sophisticated \gls{CNN} architecture, the Mask R-CNN, which was first introduced by He et al. \cite{He.2017}, will be elaborated upon, as it is the basis for the proposed method.

\subsection{Convolutional Neural Networks}
\glspl{CNN} introduce a new kind of layer to \glspl{ANN}: The eponymous convolutional layer, which uses a convolution as activation function and matrices (also referred to as filter kernels) as weights. These properties make \glspl{CNN} very effective for image processing, because they can learn complicated image filtering sequences to solve a given task by adjusting their filter kernels during the training \cite{Goodfellow.2016}.

Image filtering, i.e. convolving an image with a filter to extract a feature map, is a standard procedure in conventional image processing. However, usually, the utilized filters have to be engineered manually for a certain application (e.g. edge detection) \cite{Gonzalez.2004}. This optimization can be automated very effectively using \glspl{CNN}. Another advantage of \glspl{CNN} is the fact that they learn multiple filters per layer and combine very simple filters (e.g. for oriented edges) of lower layers to filters of increasing complexity in higher layers. Therefore, they can extract very complicated features (e.g. letters or faces) \cite{Lee.2011}. Due to this strategy, \glspl{CNN} usually apply a drastically increased number of layers in comparison to conventional \glspl{ANN}, which inspired the term deep learning \cite{Deng.2013}. 

\subsection{Mask R-CNN}

\begin{figure}
	\centering
		\includegraphics[width=\figurewidth]{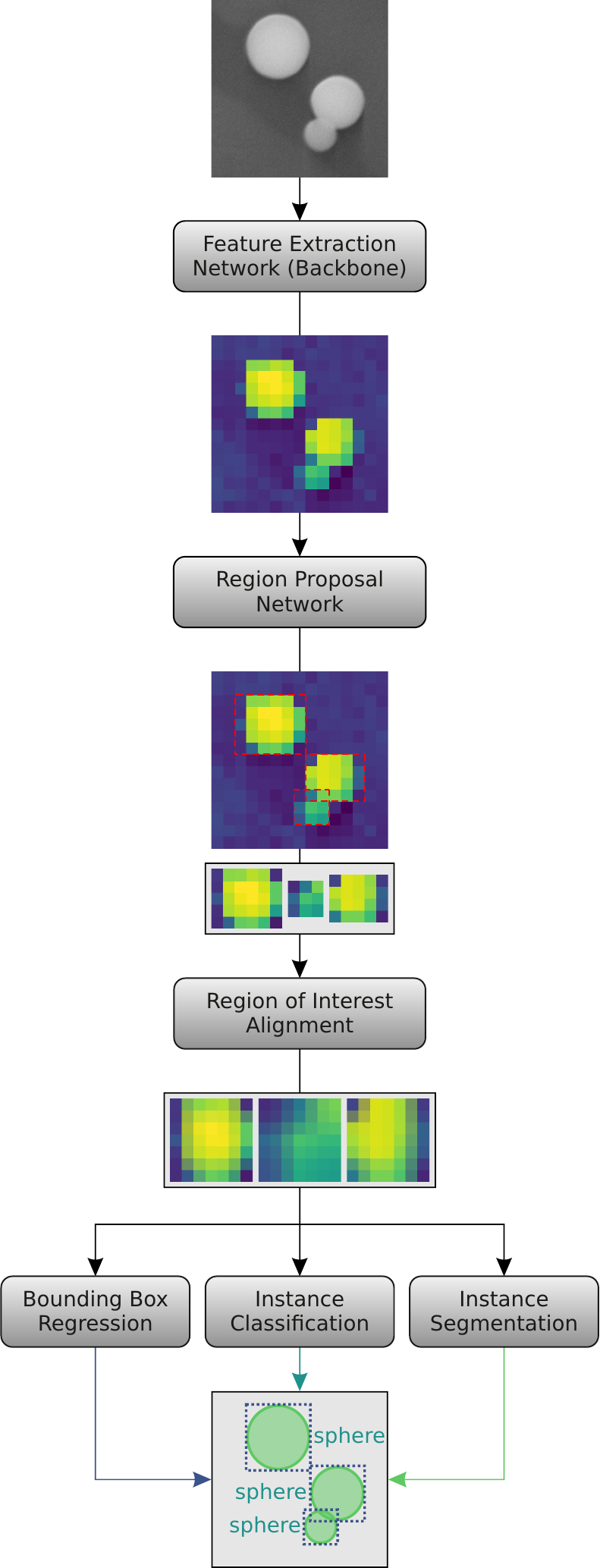}
	\caption{Mask R-CNN structure (inspired by \cite{Hui.2018}).}
	\label{fig:MaskRCNN}
\end{figure}

The Mask R-CNN \cite{He.2017} architecture is a very sophisticated example of a \gls{CNN}, which consists of six components (see \cref{fig:MaskRCNN}).

\paragraph{Feature Extraction Network (Backbone)}
The feature extraction network, also referred to as backbone, of a Mask R-CNN is a \gls{CNN}, used to extract prominent features over the entirety of an input image. One advantage of the Mask R-CNN architecture is the fact that the backbone is easily interchangeable, which substantially facilitates transfer learning\footnote{Transfer learning means that, at the beginning of the training, the weights of an \gls{ANN} are not initialized randomly but with weights that resulted from a training on a different task (e.g. the classification of everyday objects). Like that, knowledge gathered during the training on one task is transferred to another task.}.

\paragraph{Region Proposal Network}
The \gls{RPN} is a \gls{CNN} that proposes bounding boxes of \glspl{ROI}, by evaluating the feature map, which was output by the feature extraction network preceding the \gls{RPN}. Subsequently, the extracted regions of the feature map are used for all tasks involved in the instance segmentation, i.e. the segmentation, classification and bounding box regression of each instance, after the extracted regions have been refined by \gls{ROI} alignment. The fact that a single feature map is used for all tasks of the instance segmentation greatly improves the training and inference speed of the Mask R-CNN architecture.

\paragraph{Region of Interest Alignment}
The resolution of the feature map is usually significantly smaller than the one of the input image, due to one or more downsampling steps in the backbone. Masks generated based on the low resolution features can therefore be misaligned with the objects on the input image. This issue was fixed by the authors of the Mask R-CNN architecture by introducing \gls{ROI} alignment. This means that before the extraction of the \glspl{ROI}, the feature map is bilinearly interpolated to improve the alignment of the utilized \glspl{ROI} of the feature map with the actual objects on the input image.

\paragraph{Bounding Box Regression Block}
The bounding boxes of the proposed \glspl{ROI} usually do not fit the objects on the original image perfectly. Therefore, a small \gls{ANN} consisting of fully-connected layers is used for the regression of the bounding boxes, based on the previously extracted feature map, thereby improving their accuracy.

\paragraph{Instance Classification Block}
In the instance classification block, each instance is classified, again using a small \gls{ANN} of fully-connected layers with the respective \gls{ROI} of the feature map as input.

\paragraph{Instance Segmentation Block}
The segmentation of the instances featured in the \glspl{ROI} is realized by performing a pixel-based binary classification, based on the previously extracted feature map, with help of a small \gls{CNN}. During the classification, each pixel is assigned one of the classes \emph{object} or \emph{background}.

	% !TeX spellcheck = en_US
\section{Method}
\begin{figure}
	\centering
	\includegraphics[width=\figurewidth]{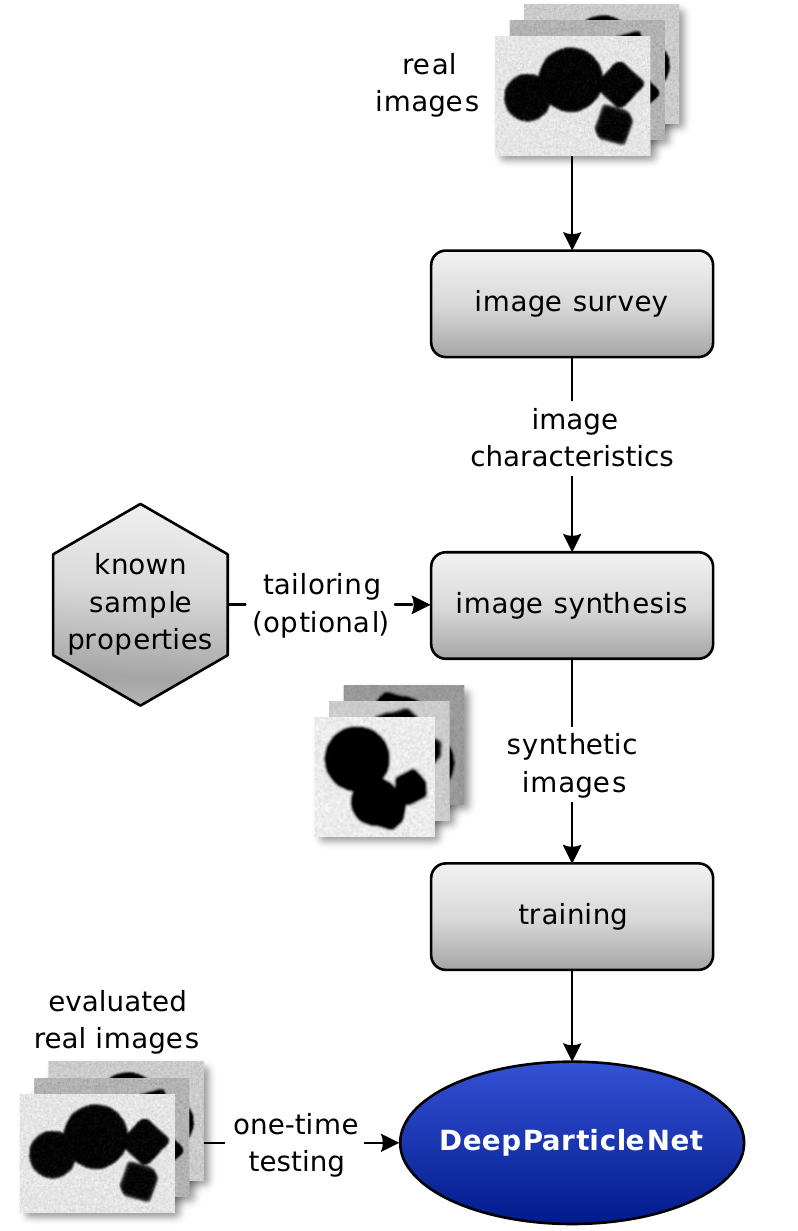}
	\caption{Workflow of the \gls{DPN} training phase.}
	\label{fig:Workflow-training}
\end{figure}

The workflow of the proposed method, where the Mask R-CNN is applied to synthetic and real particle images, consists of two phases: the training and the application phase.
\paragraph{Training Phase}
During the training phase (see \cref{fig:Workflow-training}), real images -- e.g. \gls{SEM} or \gls{TEM} images -- are surveyed for their characteristic properties (e.g. imaging method, particle shape, agglomerate size, noise and blur levels, contrast, etc.). Optionally, these observations can be supplemented with a priori knowledge about the sample at hand (e.g. approximate range of the geometric mean diameter and geometric standard deviation), to specialize the \gls{DPN} during the training and yield better analysis results in the application phase. The gathered information is used to set the parameters of the image synthesis, after which the synthetic images are used for the training of the \gls{DPN}. After the \gls{DPN} has been trained on the synthetic images, it is tested based on a small set of already evaluated real images, to prove that it can be applied to new and therefore unknown samples in the application phase.

\begin{figure}
	\centering
	\includegraphics[width=\figurewidth]{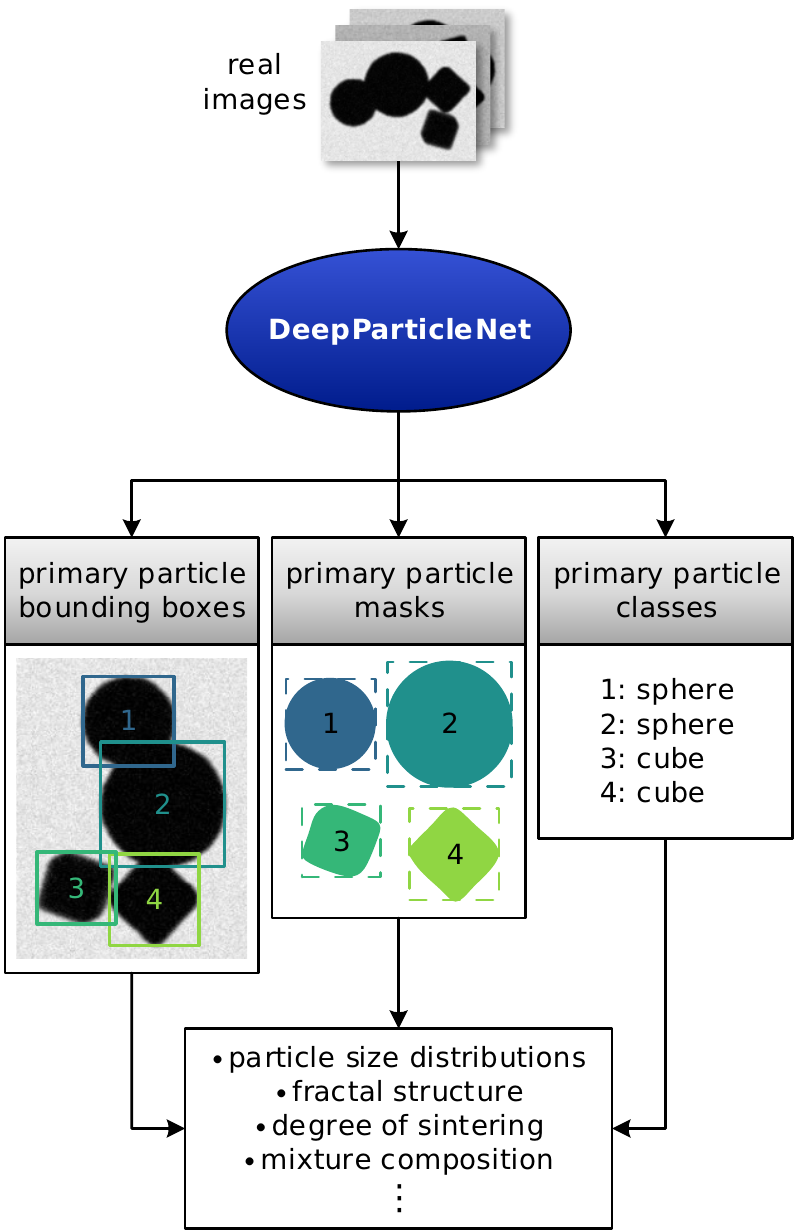}
	\caption{Workflow of the \gls{DPN} application phase.}
	\label{fig:Workflow-application}
\end{figure}

\paragraph{Application Phase}
During the application phase (see \cref{fig:Workflow-application}), the \gls{DPN} outputs three kinds of information for each detected primary particle: a bounding box, a mask and a class. These attributes allow the determination of a huge variety of properties of the particle ensemble, e.g. \gls{PSD}, fractal structure, degree of sintering, mixture composition and many more.
\newline
\newline
While the training of the \gls{DPN} usually takes a longer time (in the range of hours) and requires one or more \glspl{GPU}, the application to images can be done on a \gls{CPU} within seconds (see \cref{sec:AnalysisSpeed}).

\subsection{Training, Validation and Test Data}
\label{sec:Training-Validation-and-Test-Data}

For the training of an \gls{ANN}, usually three kinds of data sets are needed: a training set, a validation set, to verify that the \gls{ANN} is not being overfit during the training and a test set, to check the generalization abilities of the \gls{ANN} on previously unseen data after the training. For the proposed method, the training set as well as the validation set are synthetically generated to avoid the laborious manual annotation of the large number of images needed for the training and validation. However, for the test set, it is essential to use real images, to prove that the \gls{ANN} is able to make correct predictions for the data, which the \gls{ANN} is actually going to be used on. For the publication at hand, \gls{SEM} images were used for the testing of the proposed method. However, also images from \gls{TEM} or other imaging methods could be used.

\paragraph{Training and Validation Data -- Image Synthesis} A key feature of the proposed method is the fact that only synthetic images are used for the training of the \gls{DPN}, so that there is no need for a laborious manual evaluation of real images, to get a ground truth, which is essential for supervised learning. 

\begin{figure}
	\centering
	\includegraphics[width=\figurewidth]{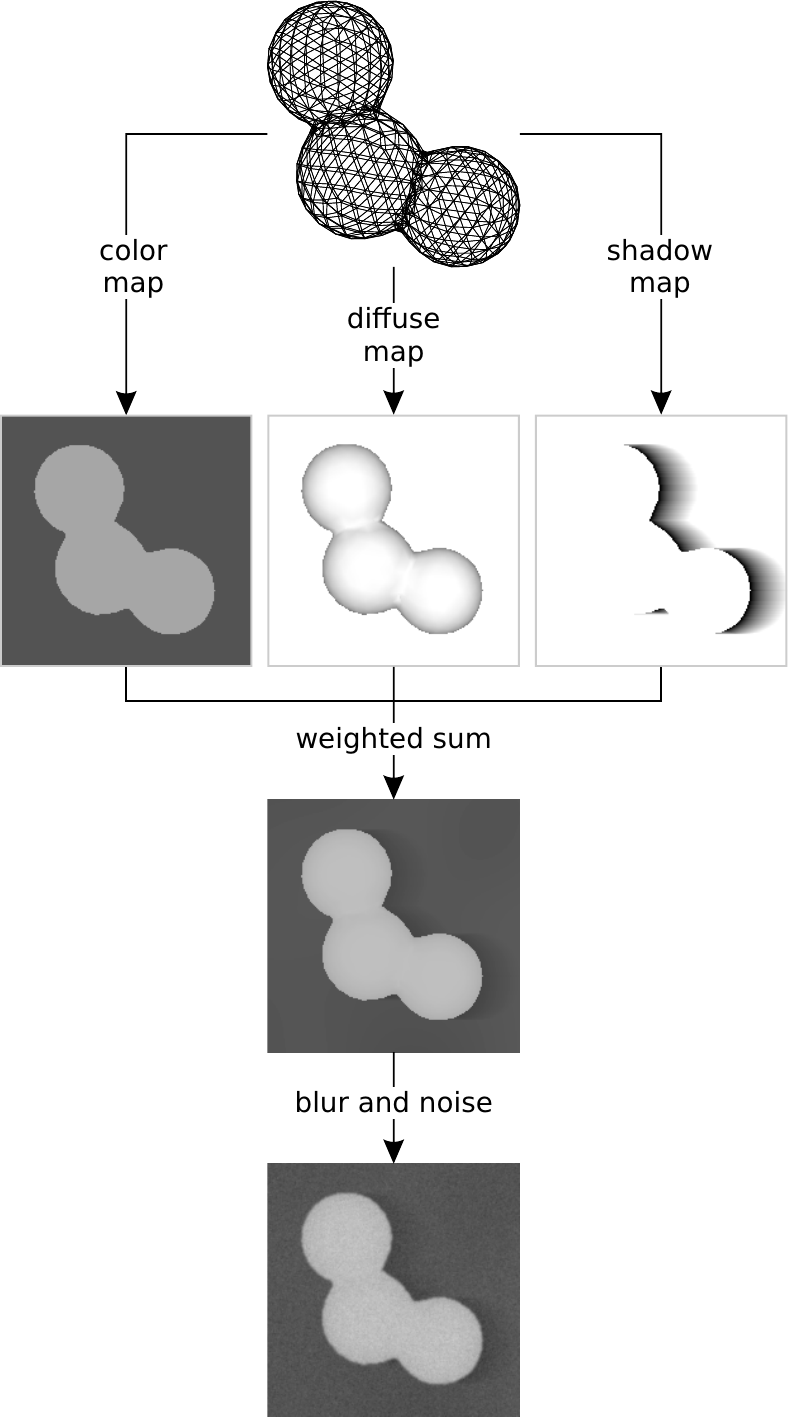}
	\caption{Elementary steps of the image synthesis.}
	\label{fig:ImageSynthesis}
\end{figure}

The image synthesis is a three step process (see \cref{fig:ImageSynthesis}):
\begin{itemize}
	\item Meshing: During the meshing\footnote{In computer graphics, a mesh is a collection of triangles in a three-dimensional space, which form the surface of an object.}, the three-dimensional geometry of a set of particles is generated. The shape of the resulting mesh depends on a variety of primary particle parameters, such as their primitive shape and their \gls{PSD}, as well as agglomerate properties, such as the agglomerate size distribution, the degree of sintering and the agglomeration mode. It is noteworthy that it is not the goal of the meshing to emulate the physical formation but rather the visual appearance of particles and agglomerates.
	\item Rendering: The rendering process transfers the mesh into a set of pixel-based intensity maps. Each of these maps represents an image feature, such as color, diffuseness or shadowing. Analogously to the meshing, the rendering does not try to simulate the image formation on a physical basis, but rather aims at a life-like emulation of the resulting image.
	\item Compositing: For the compositing, a weighted sum of the previously rendered intensity maps is calculated to form an intermediate image. Subsequently, characteristic image distortions such as blur and noise are added to the image, to increase its life-likeness.
\end{itemize}

Apart from the reduced effort, a key advantage of the image synthesis is the perfect knowledge about elsewise unknown features of the training data, such as the three-dimensional agglomerate geometry. This knowledge can then be used to improve the quality of the ground truth and thereby the training of the \gls{DPN}. In the studies at hand, the quality of the ground truth was improved drastically by the incorporation of occlusion (see \cref{fig:Occlusion}), a very important phenomenon for nontransparent materials, which is often ignored during manual evaluations. 

\nImageColumns = 4
\setlength{\spacerwidth}{1mm}
\setlength{\imagewidth}{(\figurewidth-\spacerwidth*(\nImageColumns-1))/\nImageColumns}

\begin{figure}
	\setlength{\tabcolsep}{0mm} % 
	\centering
	\begin{tabularx}{\figurewidth}{ZZZZ}
		\begin{subfigure}{\imagewidth}
			\includegraphics[width=\textwidth]{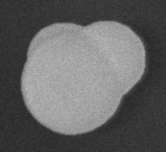}
			\caption{}
		\end{subfigure}
		&
		\begin{subfigure}{\imagewidth}
			\includegraphics[width=\textwidth]{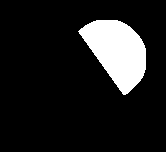}
			\caption{}
		\end{subfigure}
		&
		\begin{subfigure}{\imagewidth}
			\includegraphics[width=\textwidth]{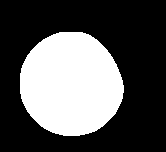}
			\caption{}
		\end{subfigure}
		&
		\begin{subfigure}{\imagewidth}
			\includegraphics[width=\textwidth]{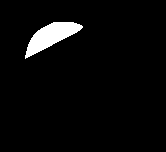}
			\caption{}
		\end{subfigure}
		\\
	\end{tabularx}
	\caption{Synthetic image of an agglomerate (a) and associated ground truth, i.e. convex occlusion masks of the primary particles (b-d).}
	\label{fig:Occlusion}
\end{figure}

To facilitate the large scale image synthesis, a toolbox -- the \gls{synthPIC}\footnote{The \gls{synthPIC} toolbox will be made available on the following website in the near future: \url{https://github.com/maxfrei750/synthPIC4Matlab}} -- was especially implemented. With help of the \gls{synthPIC} toolbox, a mixture of synthetic \gls{SEM} images, with varying degrees of sintering as well as agglomerate and primary particle sizes was generated (see \cref{fig:SyntheticImages-Examples}), to mimic the diversity of the real \gls{SEM} images of the test set (see \cref{fig:RealImages-Examples}).

% !TeX spellcheck = en_US
\setlength{\spacerwidth}{1mm}
\begin{figure*}
		\begin{subfigure}{0.5\textwidth-\spacerwidth}
			\setlength{\tabcolsep}{0mm} % 
			\centering
				\begin{tabularx}{\textwidth}{LR}
					\includegraphics[width=\linewidth-0.5\spacerwidth]{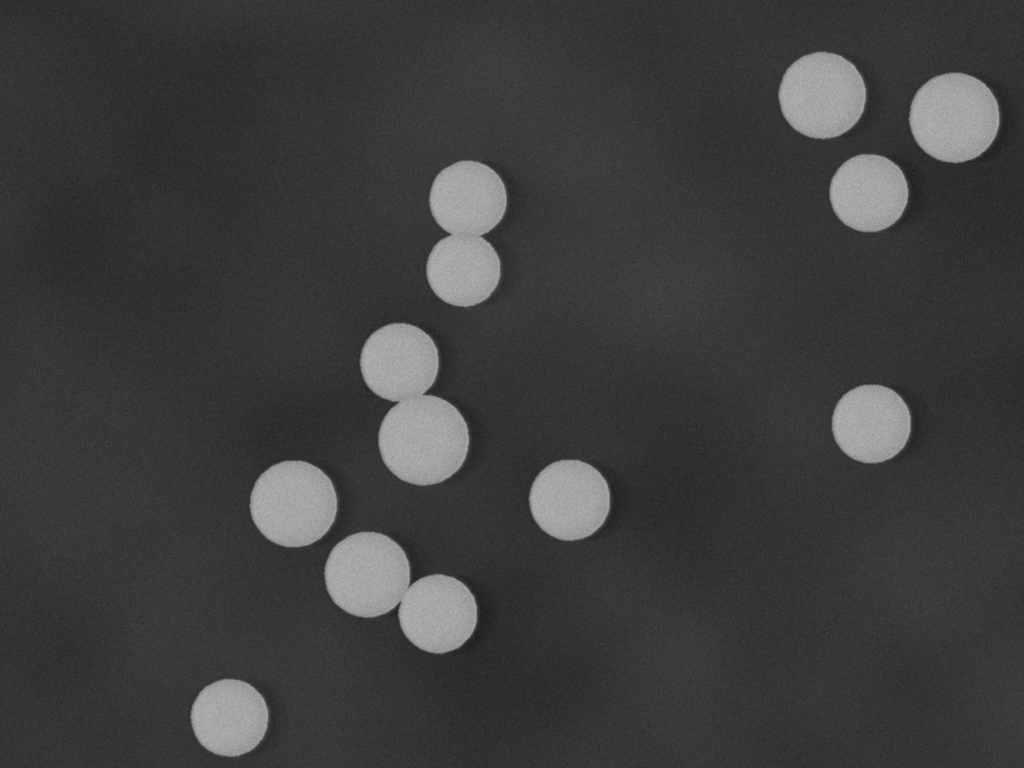}
					&
					\includegraphics[width=\linewidth-0.5\spacerwidth]{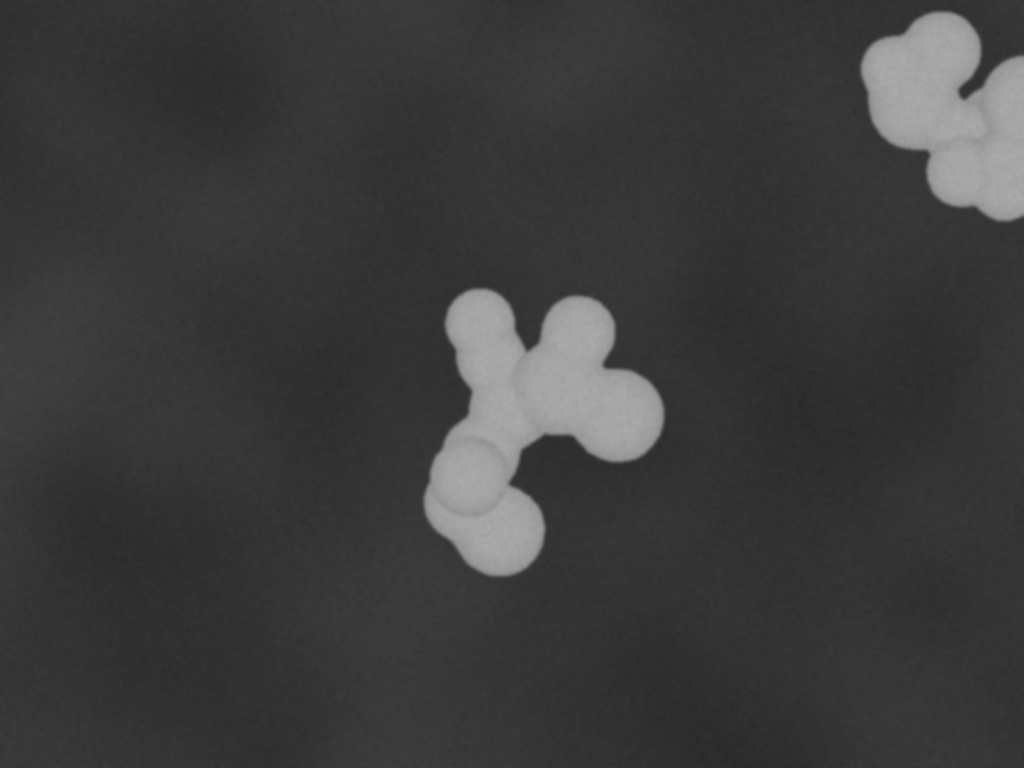}
					\\
					\includegraphics[width=\linewidth-0.5\spacerwidth]{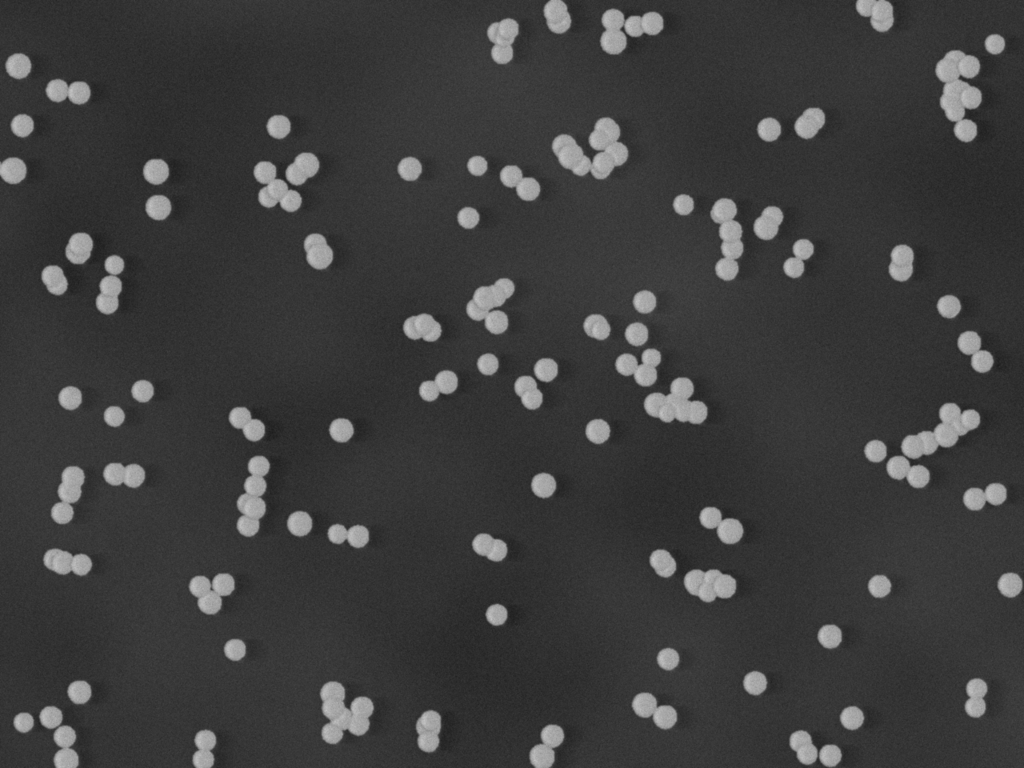}
					&
					\includegraphics[width=\linewidth-0.5\spacerwidth]{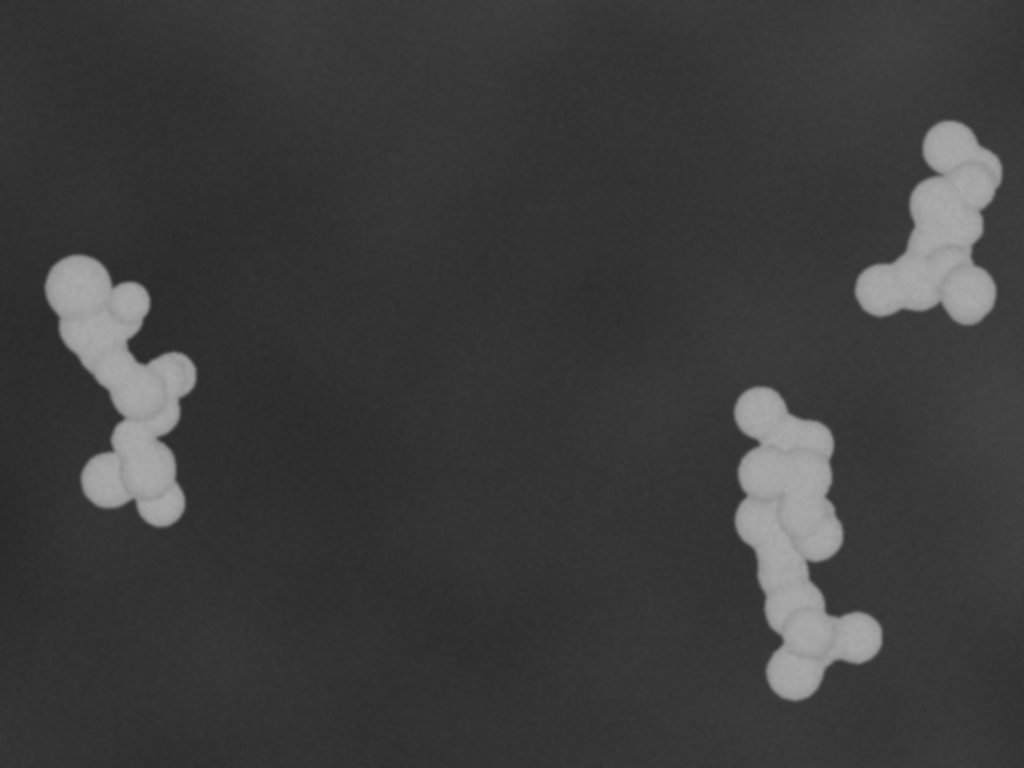}
					\\
				\end{tabularx}
			\caption{}
			\label{fig:SyntheticImages-Examples}
		\end{subfigure}
		\hfill
		\begin{subfigure}{0.5\textwidth-\spacerwidth}
			\setlength{\tabcolsep}{0mm} % 
			\centering
				\begin{tabularx}{\textwidth}{LR}
					\includegraphics[width=\linewidth-0.5\spacerwidth]{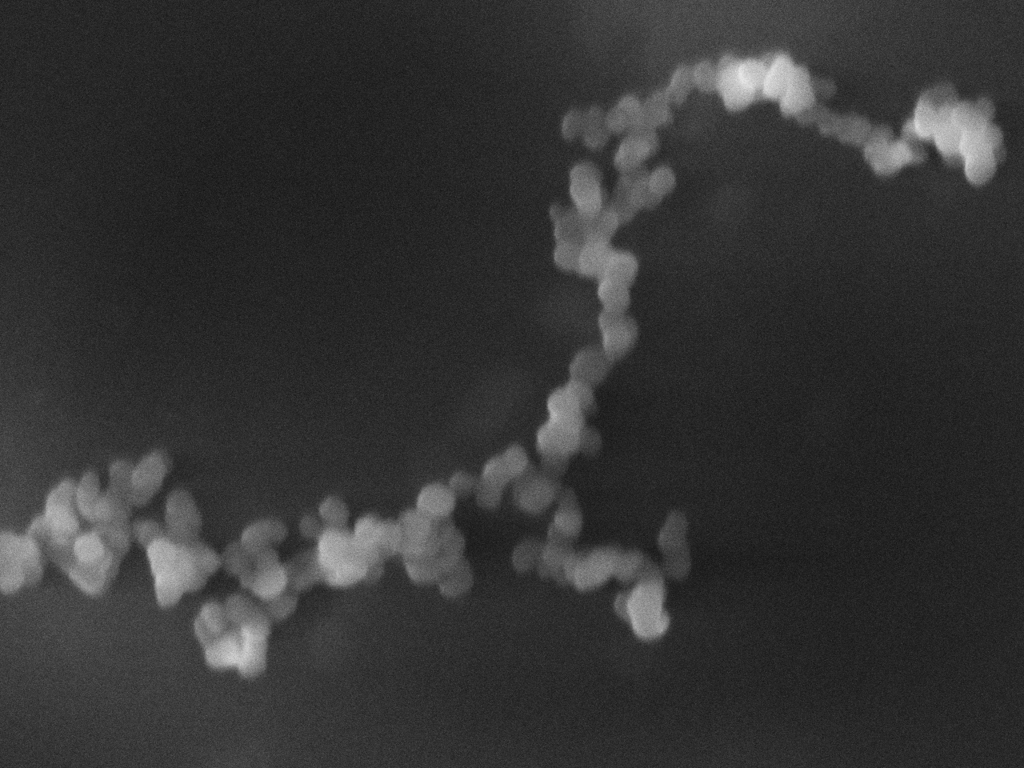}
					&
					\includegraphics[width=\linewidth-0.5\spacerwidth]{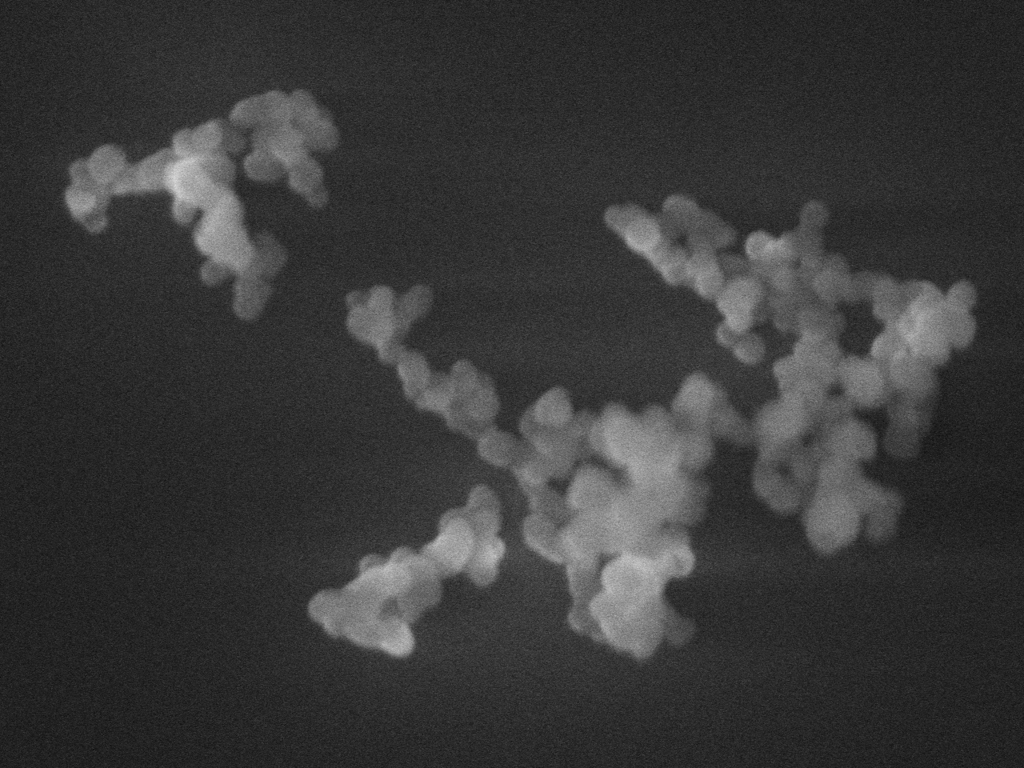}
					\\
					\includegraphics[width=\linewidth-0.5\spacerwidth]{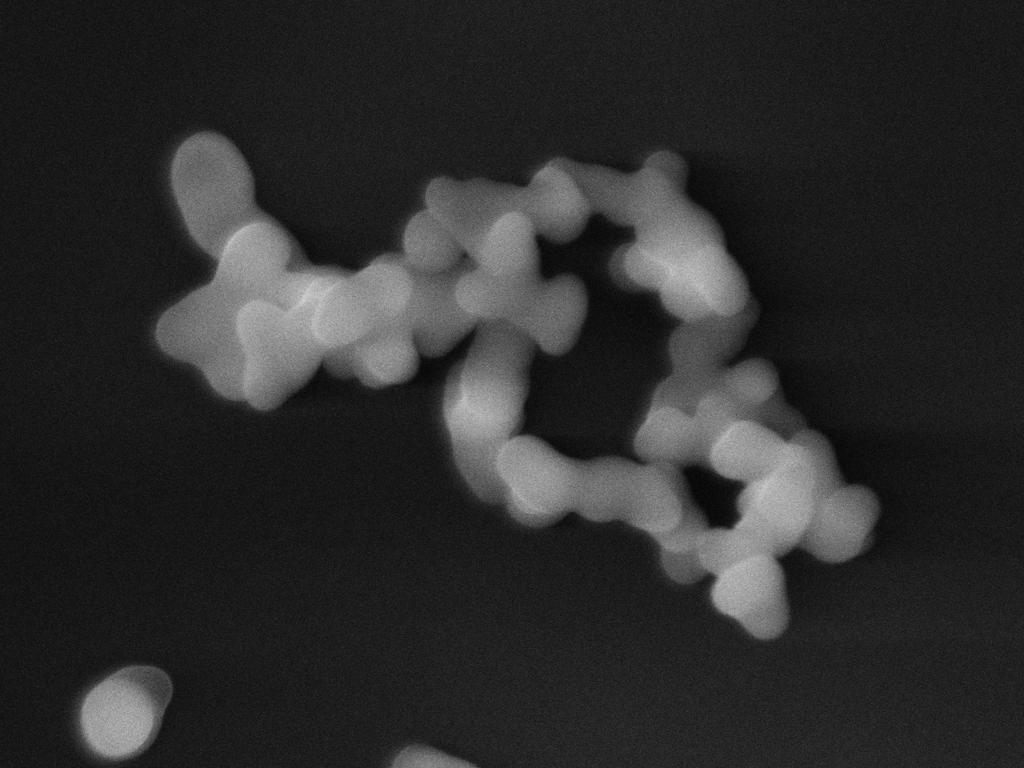}
					&
					\includegraphics[width=\linewidth-0.5\spacerwidth]{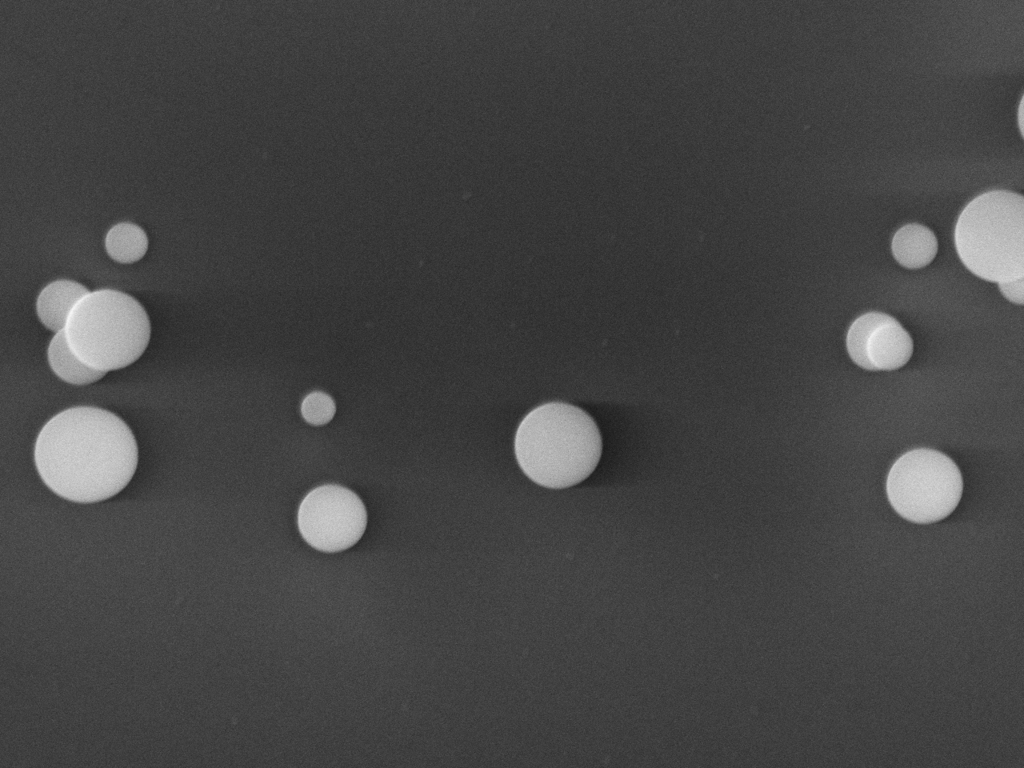}
					\\
				\end{tabularx}
			\caption{}
			\label{fig:RealImages-Examples}
		\end{subfigure}
	\caption{Examples of synthetic images used for the training and validation (a) and real images from the 10 data sets used for the testing (b) of the \gls{DPN}. For a complete overview of the 10 test sets please refer to the Supplementary Materials (Figure SM.1).}
\end{figure*}

\paragraph{Test Data -- \glsentrytitlecase{SEM}{long}} As test data, real \gls{SEM} images of agglomerated and partially sintered silica nanoparticles were used (see \cref{fig:RealImages-Examples}). The images resulted from a sintering study, performed by Babick et al. \cite{Babick.2018}.

During the sintering study, particles of a certain size were generated and subsequently sintered at different temperatures in the range of \SIrange{1000}{1500}{\celsius} (see Supplementary Materials Table SM.1). As a result, 10 samples which exhibit increasing degrees of sintering were produced and subsequently analyzed with \gls{SEM}\footnote{JEOL JSM-7500F Field Emission Scanning Electron Microscope}. For the samples with sintering temperatures in the range of \SIrange{1000}{1300}{\celsius} and \SIrange{1350}{1500}{\celsius} magnifications of \SI{100}{\kilo\times} and \SI{50}{\kilo\times} were used respectively, due to the primary particle size increasing with the sintering temperature. In total, the test data consists of 10 data sets, with a total of 99 images (see Supplementary Materials Table SM.2), featuring 6126 primary particles in 699 agglomerates (according to a manual evaluation). Each image has a resolution of \SIrange[range-phrase=$\times$]{768}{1024}{\px}.

A very important feature of the test data is its high level of heterogeneity with respect to the image conditions (e.g. brightness, contrast, noise, blur and background texture), due to varying sample and instrument conditions as well as varying \gls{SEM} operators. Usually these factors severely impede the reliability of imaging particle size analysis methods. Therefore, good results on heterogeneous test data would indicate a high robustness and flexibility of the proposed method.

\paragraph{Similarity of Training and Test Data}
There are two levels of similarity with respect to the training and the test data: On the one hand, there is the visual likeness of the synthetic and real images and on the other hand, there is the possible similarity of the distributions of the target measurand, i.e. the primary particle size of the depicted particles.

Concerning the two kinds of possible similarity of the utilized training and test data, we face a dilemma. The synthesized images must exhibit a certain life-likeness, so that the \gls{DPN} can pick up characteristic features of the training data, which also apply to the test data. This includes, at least to a certain extent, also the primary particle sizes. However, the training and test data may not be too similar, so that the \gls{DPN} learns to generalize and does not suffer from a data bias.

\begin{figure}
	\centering
	\includegraphics[width=\figurewidth]{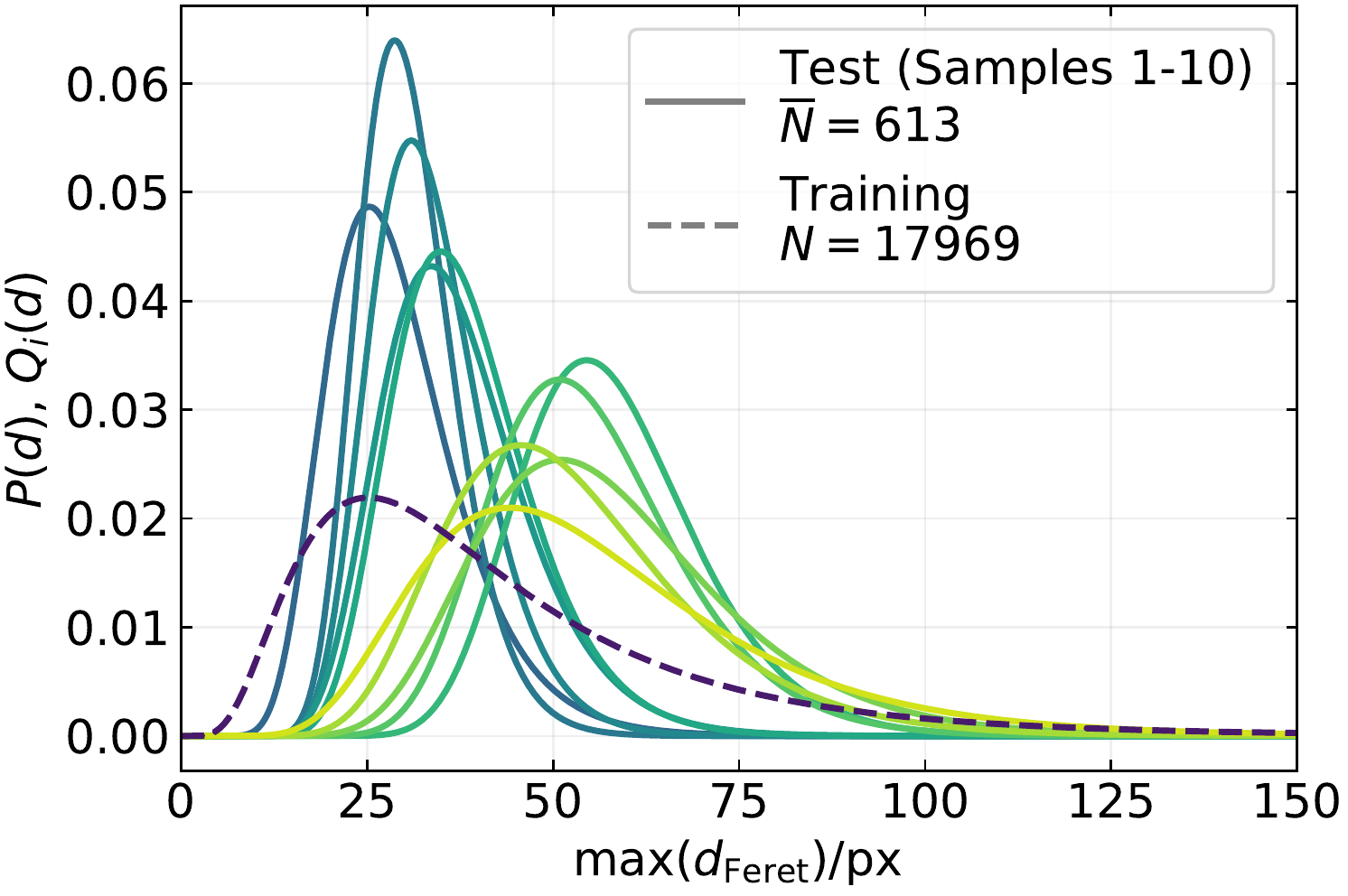}
	\caption{Comparison of the distributions \gls{symb:probability_distribution_p} and \gls{symb:probability_distribution_q} of the maximum Feret diameter $\max(\gls{symb:feret_diameter})$ of the primary particles of the training set and the 10 test sets, respectively. \gls{symb:number_primary_particles} is the number of primary particles.}
	\label{fig:DataBias}
\end{figure}

To study the similarities of the training set and the 10 test sets with regard to the measurand, the primary \gls{PSD} \gls{symb:probability_distribution_p} of the training set was compared to the primary \glspl{PSD} \gls{symb:probability_distribution_q} of the test sets (see \cref{fig:DataBias}). As equivalent diameter, the maximum Feret diameter\footnote{The Feret diameter \gls{symb:feret_diameter} of a non-circular object depends on the direction of the measurement. Therefore, the maximum Feret diameter is the maximum of all the Feret diameters of an object. It equals the largest possible euclidean distance of two points on the outline of an object.} $\max(\gls{symb:feret_diameter})$ of the primary particle masks was used, to account for the occlusions in case of the training set (see paragraph \emph{Training and Validation Data -- Image Synthesis}). 

The comparison yielded two important qualitative insights:
\begin{itemize}
	\item The \gls{PSD} of the training set covers the full range of all individual test set \glspl{PSD} and is therefore much broader.
	\item The similarity of the training set \gls{PSD} and the individual test set \glspl{PSD} is rather low. 
\end{itemize}

For a better, quantitative, comparability of the similarities of the \glspl{PSD} of the training and the test sets, it is helpful to introduce a similarity or inversely a divergence measure. A common measure for the divergence of two probability distributions \gls{symb:probability_distribution_p} and \gls{symb:probability_distribution_q} is the Kullback-Leibler divergence \cite{MacKay.2016}:
\begin{equation}
	\gls{symb:kullback_leibler_divergence}(\gls{symb:probability_distribution_p} \parallel \gls{symb:probability_distribution_q}) = \sum_{x} \gls{symb:probability_distribution_p}(x) \log\left(\frac{\gls{symb:probability_distribution_p}(x)}{\gls{symb:probability_distribution_q}(x)}\right),
\end{equation}
where, for the case at hand, \gls{symb:probability_distribution_p} and \gls{symb:probability_distribution_q} are the primary \glspl{PSD} of the training set and one of the test sets respectively.\footnote{For the calculation, bins where either $\gls{symb:probability_distribution_p}(d)=0$ or $\gls{symb:probability_distribution_q}(d)=0$ were excluded.} The Kullback-Leibler divergence yields values which range from 0 for identical to 1 for completely different probability distributions.

The comparisons of the \glspl{PSD} of the training with the test sets yield Kullback-Leibler divergences in the range from 0.432 to 0.870 (see Supplementary Materials Table SM.3), which indicate low levels of similarity. It can therefore be concluded that the risk of a data bias is low.

\subsection{DeepParticleNet}
\label{sec:DeepParticleNet}
The \gls{DPN} is the \gls{CNN} at the core of the proposed method, which is trained on the detection of individual primary particles. Its architecture was inspired by the Mask R-CNN architecture, developed by He et al. \cite{He.2017} and based on an implementation of Abdulla \cite{Abdulla.2017}, realized with Keras \cite{Chollet.2015} and TensorFlow \cite{Abadi.2015}, controlled by Python \cite{Python.2019}.

\subsubsection{Architecture}
As backbone for the \gls{DPN}, a ResNet architecture \cite{He.2015} was chosen, because it offers several benefits for the task at hand: Firstly, it allows the use of very deep networks, without facing problems concerning vanishing gradients, i.e. that the training stagnates, due to the involved gradients becoming very small \cite{He.2015}. Secondly, the ResNet architecture offers high accuracies at comparably low computational costs \cite{Canziani.2016}. Thirdly, there are multiple variants of the ResNet architecture with increasing numbers of layers, which allows a choice between higher accuracies and lower computational costs. Last but not least, due to its popularity, there are numerous sets of weights available for the various ResNet models, which result from extensive trainings on large data sets of everyday objects (e.g. ImageNet \cite{Russakovsky.2014} and \gls{COCO} \cite{Lin.2014}) and can therefore be used for transfer learning.

\begin{figure}[t]
	\centering
	\includegraphics[width=\figurewidth]{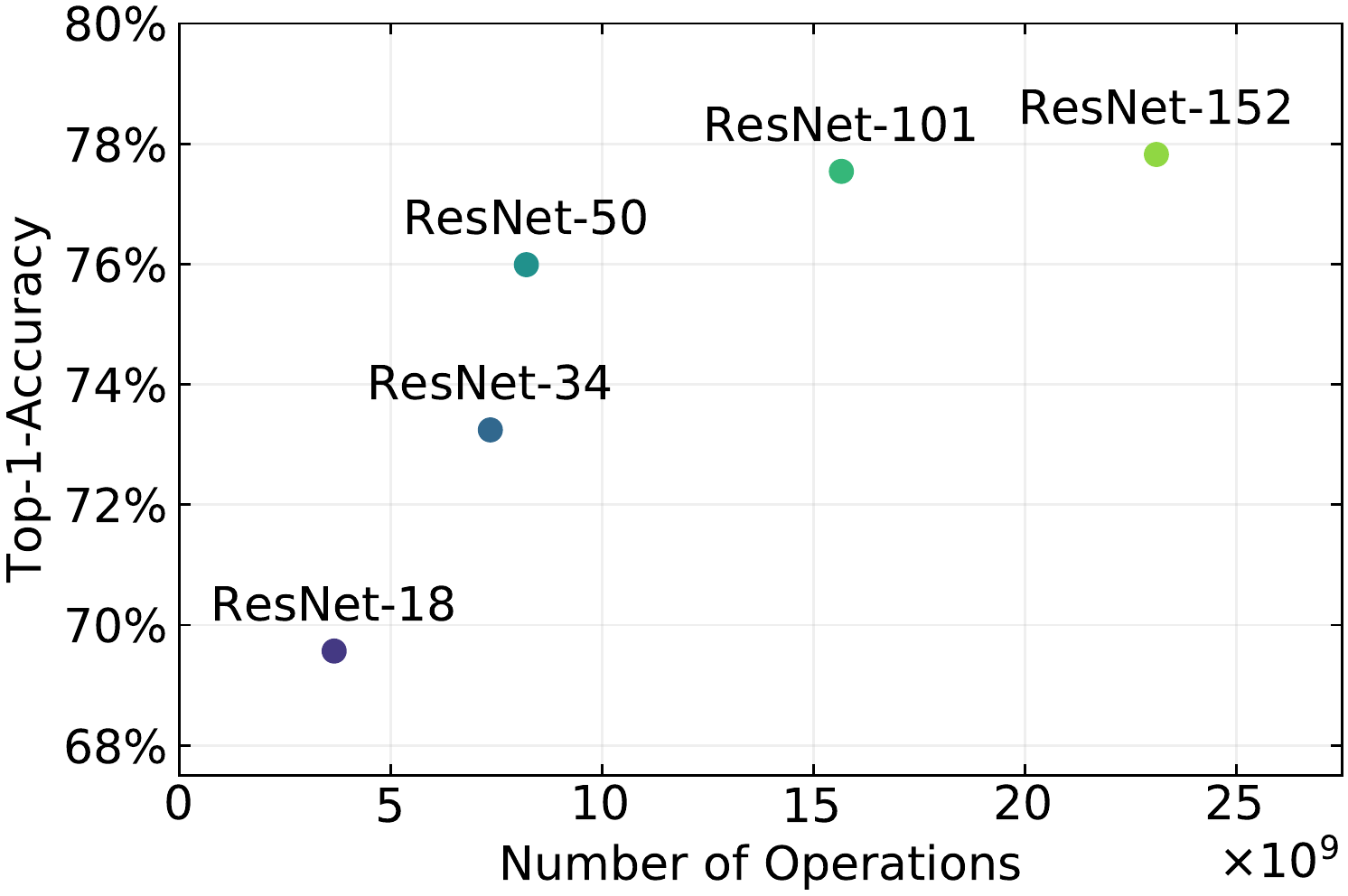}
	\caption{Top-1-accuracies of different ResNet variants, when being trained on the ImageNet data set, versus their computational costs, i.e. the number of operations, needed for a single forward pass (data from \cite{He.2015}, inspired by \cite{Canziani.2016}).}
	\label{fig:Resnet-Comparison}
\end{figure}

\paragraph{Choice of a ResNet Variant as Backbone} As mentioned before, there are multiple variants of the ResNet architecture (see \cref{fig:Resnet-Comparison}). The top-1-accuracy\footnote{The top-1-accuracy is a common measure for classification tasks (such as the ImageNet competition). It is calculated as the percentage of correctly classified images, under the condition that the method is only allowed to make a single prediction per image.} of the models, when being trained on the ImageNet data set, increases with an increasing number of layers. However, this also increases their computational costs. For the proposed method, ResNet-50 and ResNet-101 were chosen as possible candidates for the backbone of the \gls{DPN}, due to their high accuracies at reasonable computational costs. 

\begin{figure}
	\centering
	\includegraphics[width=\figurewidth]{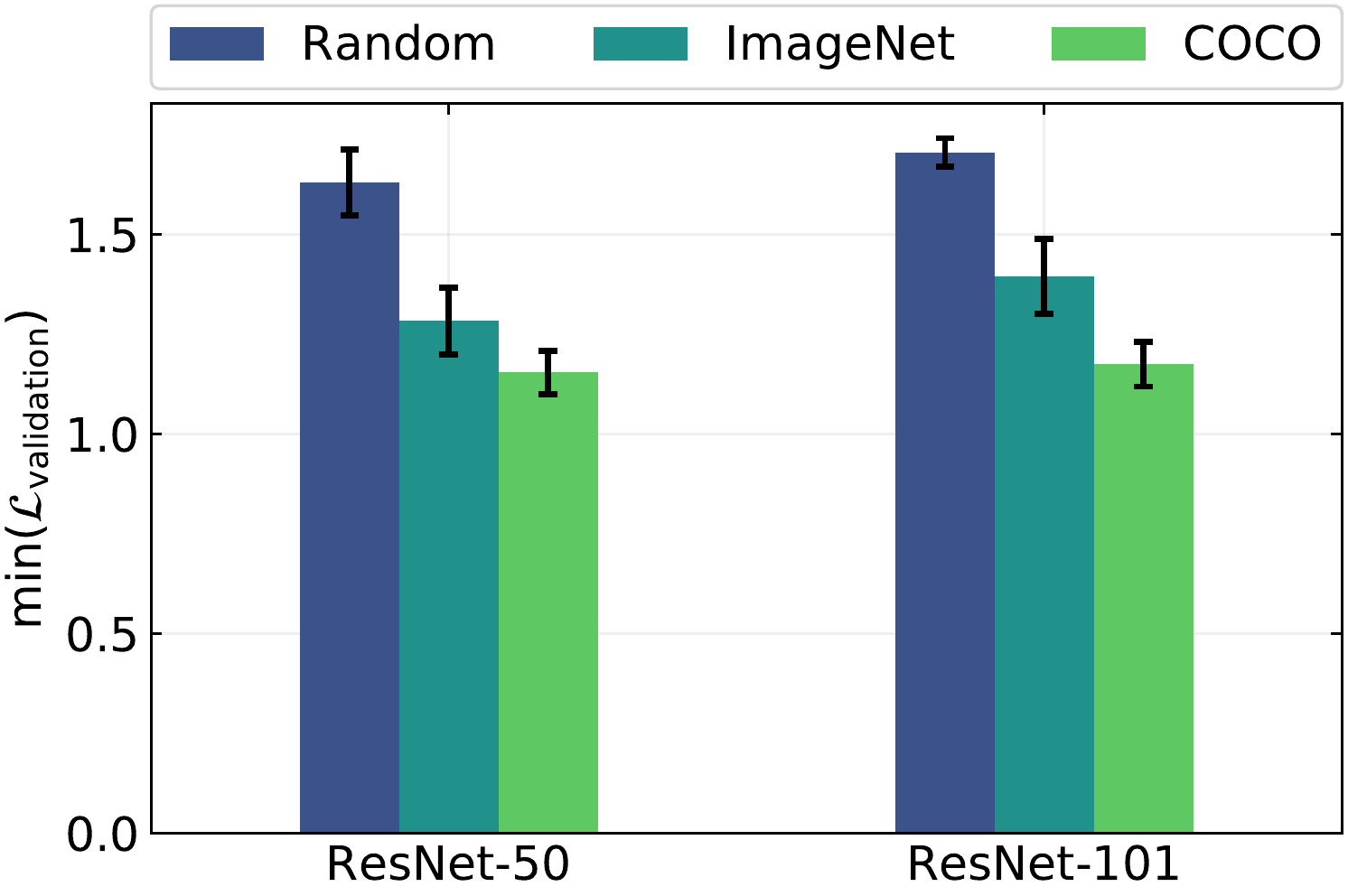}
	\caption{Comparison of the minimum validation losses \gls{symb:loss_val} of \glspl{DPN} on the validation data set, with ResNet-50 and -101 models, initialized with different weights, as backbones (3 repetitions, error bars represent $\pm\gls{symb:standard_deviation}$).}
	\label{fig:Resnet-Comparison-50vs101}
\end{figure}

For the final decision concerning the backbone of the \gls{DPN}, the losses \gls{symb:loss_val} of the two candidate models on the validation data set were compared (training conditions: see Supplementary Materials Table SM.4; see also \cref{sec:Training}). On the one hand for randomly initialized weights and on the other hand for initializations with weights resulting from previous trainings on the ImageNet and the \gls{COCO} data set (see \cref{fig:Resnet-Comparison-50vs101}). 

The comparison yielded three important insights:
\begin{itemize}
	\item Transfer learning is a very effective means to yield lower validation losses.
	\item Weights resulting from a previous training on the \gls{COCO} data set yield lower validation losses than weights resulting from a previous training on the ImageNet data set.
	\item \glspl{DPN} with ResNet-50 backbones offer slightly lower minimum validation losses, compared to \glspl{DPN} with ResNet-101 backbones. However, the differences are negligible.
\end{itemize}

The last of the three insights is rather counterintuitive, considering the higher performance of the ResNet-101 backbone on the ImageNet data set (see \cref{fig:Resnet-Comparison}). However, a possible explanation might be the fact that the additional layers of the ResNet-101 backbone extract increasingly intricate image features. While this might be beneficial for the classification of very complex objects like animals, it might be superfluous for the detection of rather simple objects like primary particles, which mainly consist of basal features like straight and curved edges. Therefore, additional layers might rather increase the noise than extract helpful image features.

Ultimately, the ResNet-50 model initialized with weights resulting from a previous training on the \gls{COCO} data set was used as backbone for the \gls{DPN}, due to it exhibiting the highest performance at the lowest computational cost.

\subsubsection{Training}
\label{sec:Training}

From an application-oriented point of view, the training of \glspl{ANN} is the most elaborate task for the application of deep learning methods to new problems. An architecture that is in principle suitable for the solution of a given task, might yield insufficient results, when being applied to new kinds of data. Therefore, this section will present state-of-the-art concepts and strategies for the training of \glspl{ANN}, as a guide line, to enable the reader to apply the proposed method to his or her own data. 

\paragraph{Early Stopping} Early stopping is a very simple, yet effective regularization method, i.e. a means to avoid overfitting \cite{Goodfellow.2016}. During the training of \glspl{ANN}, it is a very common phenomenon that the validation loss reaches a minimum while the training loss still decreases. Therefore, by stopping the training early, i.e. as soon as the validation loss starts to increase again, we can not only save on training time but also reach a higher level of generalization and thus a better performance during the deployment of the \gls{ANN}. However, due to the noisy nature of the validation loss, especially for the small batch sizes\footnote{The batch size is the number of samples (in the case at hand images) that is processed simultaneously by an \gls{ANN}. It is therefore a very important hyperparameter to control the memory consumption.} often used in deep learning, it is sensible not to stop the training immediately when encountering a potentially merely local minimum of the validation loss. Instead, it is sensible to wait for a predefined number of epochs, to ensure that the validation loss really does not decrease any further. The number of epochs to be waited is called early stopping patience. For the publication at hand, multiple early stopping patiences were tested (see Supplementary Materials Figure SM.2 and Table SM.5) and an early stopping patience of 10 epochs was used, as a compromise between training time and minimum validation loss.

\paragraph{Optimization of the Learning Rate}
The learning rate \gls{symb:lr} is the most important hyperparameter for the training of \glspl{CNN} \cite{Smith.2015}. Often, the optimal value for the learning rate is determined by rather brute force, using grid or random search. However, these simple approaches are very time consuming. Therefore, for the proposed method, a cyclical learning rate strategy was tested, as described by Smith \cite{Smith.2015}, using code of Kenstler \cite{Kenstler.2017}. This means that during the training, the learning rate was not held constant but varied cyclically within an optimal learning rate range.

\begin{figure}
	\centering
	\includegraphics[width=\figurewidth]{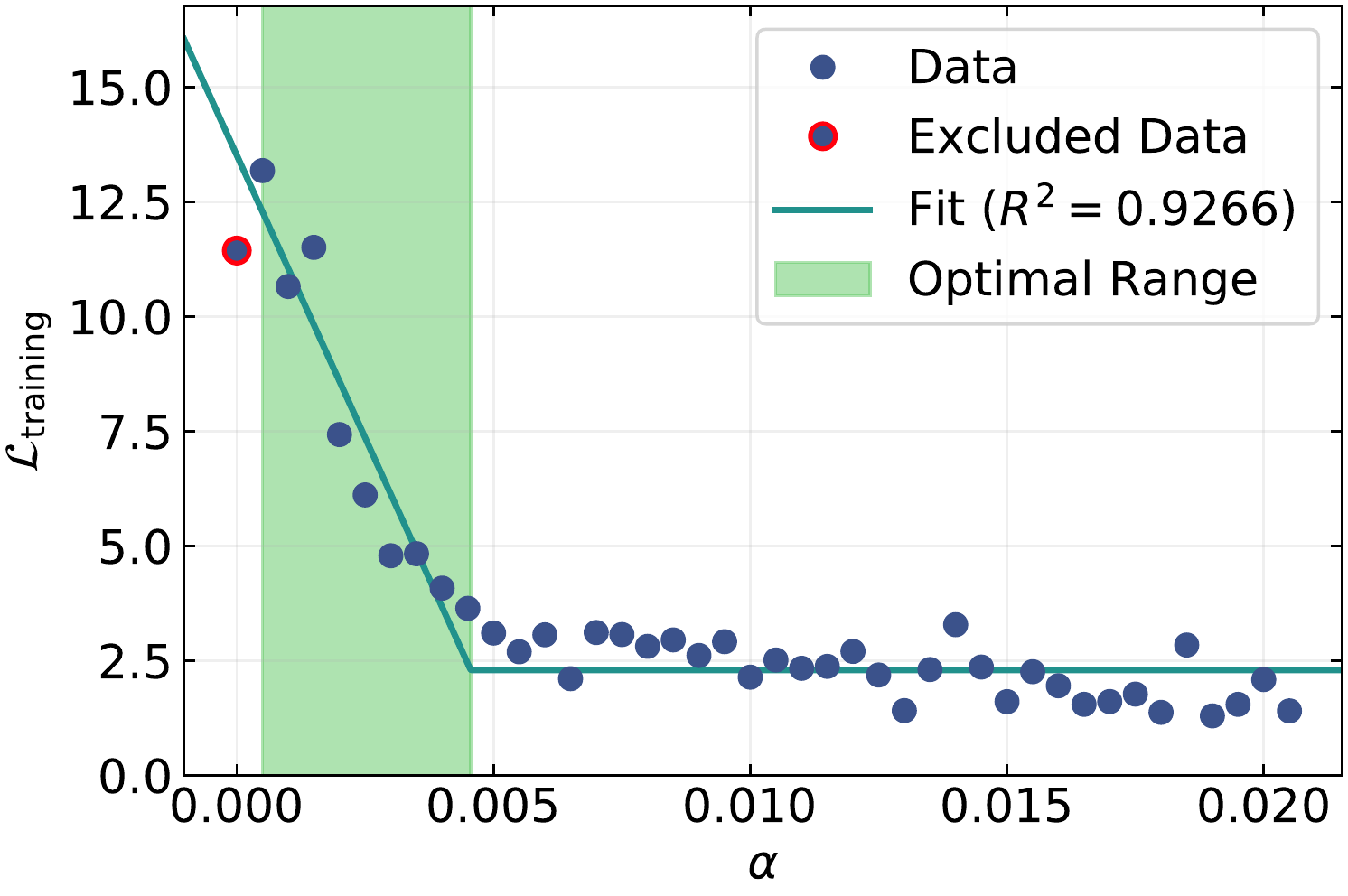}
	\caption{Learning rate range test -- Training loss \gls{symb:loss_tr} versus the learning rate \gls{symb:lr}.}
	\label{fig:LearningRateRangeTest}
\end{figure}

According to Smith \cite{Smith.2015}, a simple test can be utilized to find the optimal learning rate range. The learning rate is linearly increased and the training loss of the trained \gls{CNN} is monitored. The optimal learning rate ranges from \gls{symb:lr_min}, the point of the initial training loss maximum, to \gls{symb:lr_max}, where the training loss no longer decreases (see \cref{fig:LearningRateRangeTest}).

To introduce a reproducible measure for \gls{symb:lr_max}, the data from the learning rate range test was fit with a function of the form
\begin{equation}
	f(\gls{symb:lr}) = \max{(m\gls{symb:lr}+b,c)}
\end{equation}
where data points with learning rates below \gls{symb:lr_min} were excluded from the fit. Subsequently, the crossover of the two line segments of $f(\alpha)$ at 
\begin{equation}
	\gls{symb:lr_max} = \frac{c-b}{m}
\end{equation}
was used as upper boundary \gls{symb:lr_max} of the optimal learning rate range.

\begin{figure}[h]
	\centering
	\includegraphics[width=\figurewidth]{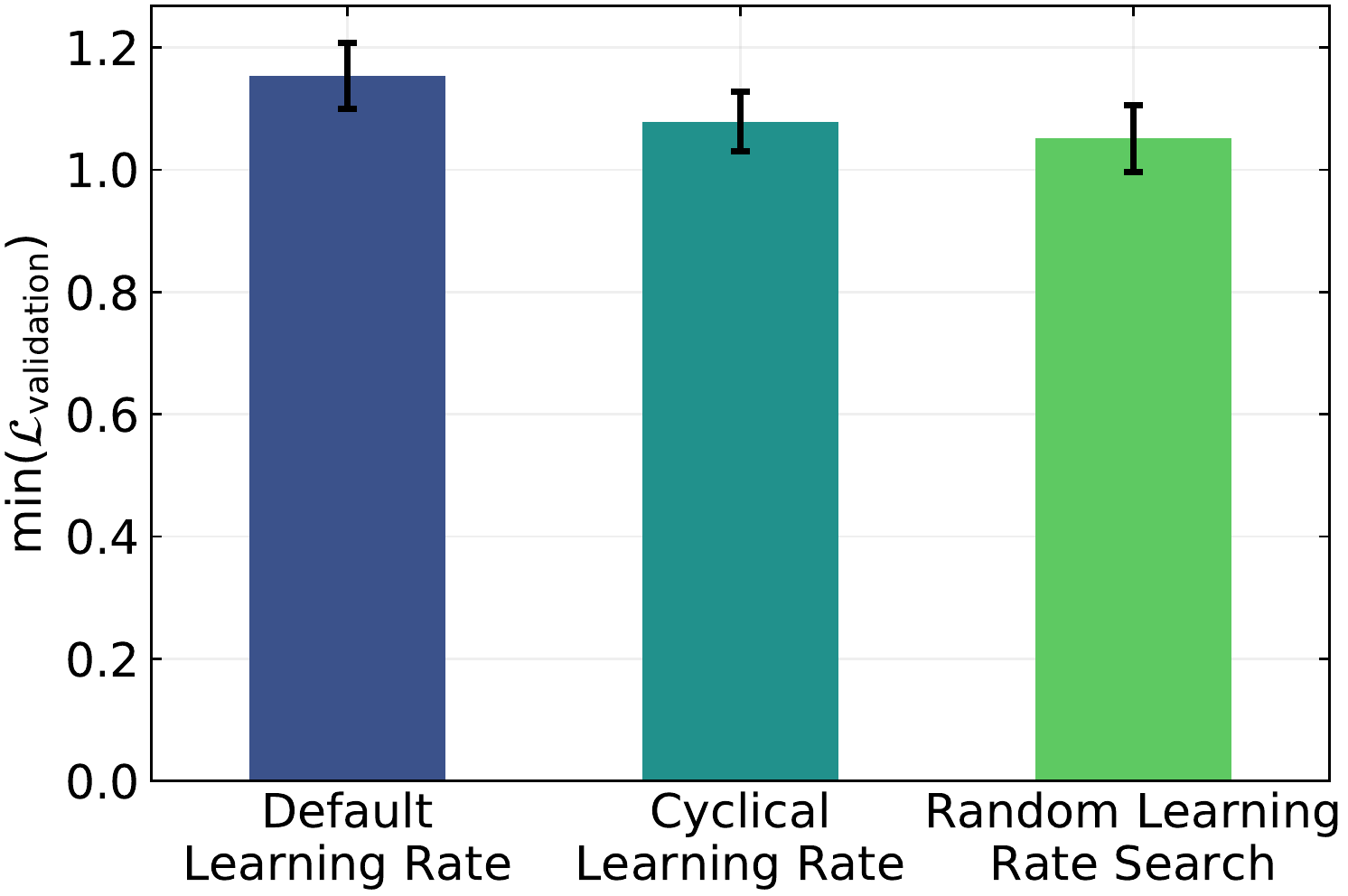}
	\caption{Comparison of minimum validation losses \gls{symb:loss_val} achieved with different learning rate optimization strategies (3 repetitions, error bars represent $\pm\gls{symb:standard_deviation}$).}
	\label{fig:LearningRateMethodComparison}
\end{figure}

Using the learning rate range determined in this way and the recommendation of Smith \cite{Smith.2015} concerning the optimal cycle length (see Supplementary Materials Table SM.6), a training utilizing a cyclical learning rate policy was carried out. The resulting minimum validation loss was compared to that achieved using a random learning rate search (see Supplementary Materials Figure SM.3) and the default learning rate (0.001, according to \cite{Abdulla.2017}), respectively (see \cref{fig:LearningRateMethodComparison}).

The comparison shows that the default learning rate does not yield optimal results. Both the cyclical learning rate strategy as well as the random learning rate search can improve the minimum validation loss, with the latter performing slightly better. However, the random learning rate search consumes a substantially larger amount of training time and is very susceptible to small changes concerning the randomly chosen learning rate (see Supplementary Materials Figure Sm.3). Therefore, the cyclical learning rate strategy was utilized for the further experiments.

\paragraph{Optimal Number of Training Images} The fact that the proposed method utilizes synthetic images for the training of the \gls{DPN} allows us to be very generous with respect to the number of utilized training images. However, if the training was to be performed on real images, e.g. to further improve the performance of the proposed method, the training data would have to be annotated manually, which is a very laborious task. Therefore, to give an orientation on how many training images are actually necessary, the influence of the number of training images on the minimum validation loss of the \gls{DPN} was examined (see \cref{fig:NumberOfTrainingSamplesAndAugmentation}; training conditions: see Supplementary Materials Table SM.8).
\begin{figure}
	\centering
	\includegraphics[width=\figurewidth]{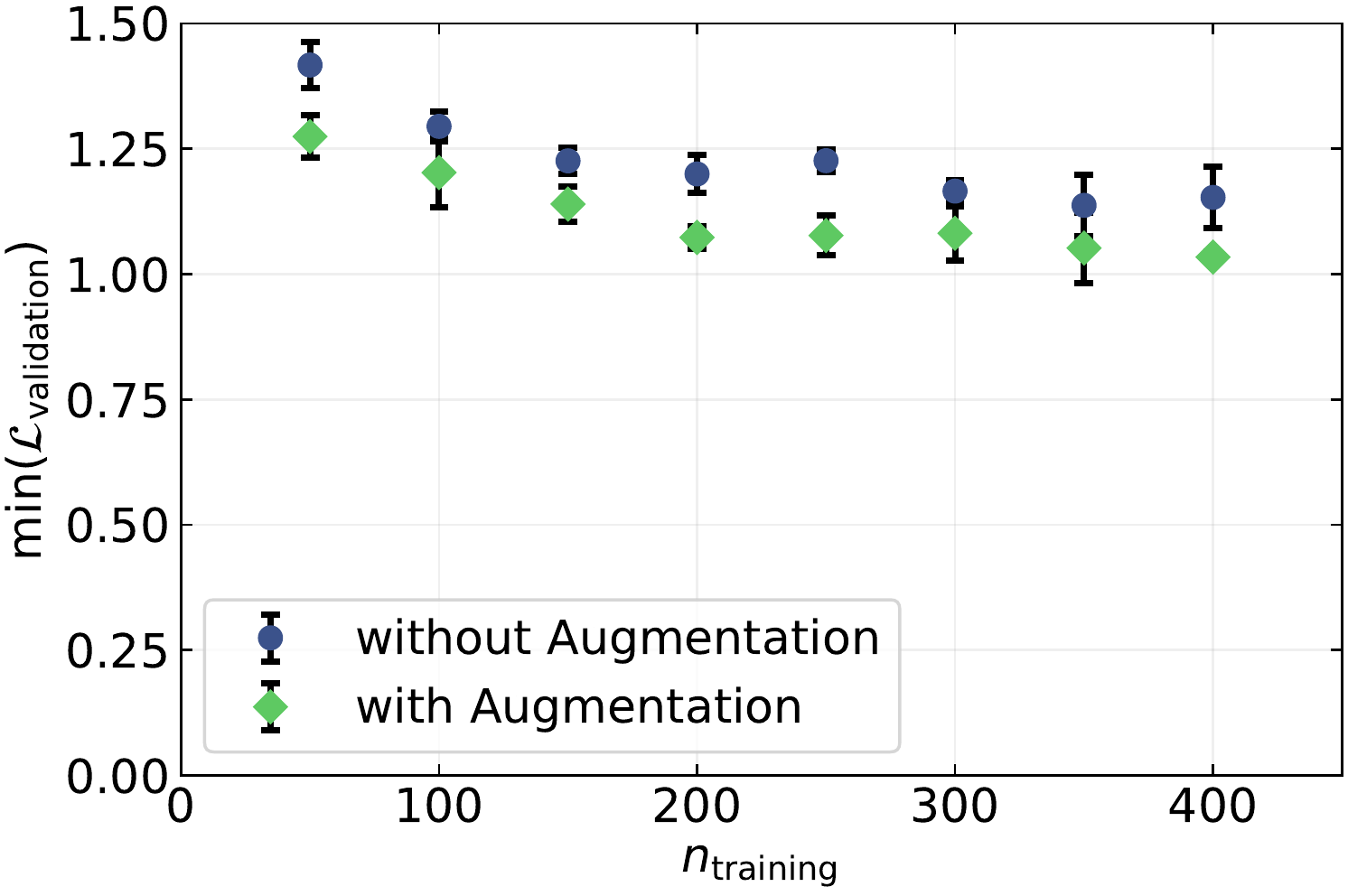}
	\caption{Minimum validation losses \gls{symb:loss_val} versus the number of images in the training set \gls{symb:training_set_size}, with and without image augmentation (3 repetitions, error bars represent $\pm\gls{symb:standard_deviation}$).}
	\label{fig:NumberOfTrainingSamplesAndAugmentation}
\end{figure}

The minimum validation loss decreases with an increase of the training set size up to a number of 300 images. However, beyond this point additional images do not further improve the performance of the \gls{DPN}, so that it reaches a validation loss threshold. 

\paragraph{Image Augmentation} A probate means to decrease the necessary number of training images is image augmentation, i.e. that training images are being reused after they have been transformed (e.g. rotated) or distorted (e.g. blurred). Like that, the number of training images can effectively be increased and the generalization capabilities of the model are enhanced, due to the training being more versatile \cite{Goodfellow.2016}.

For an evaluation of the potential benefits of image augmentation for the given use case, the influence of the number of training images on the minimum validation loss of the \gls{DPN} was reexamined, using a selection of up to two, from a list of five, image augmentations (see Appendix \cref{app:tab:Augmentation}; training conditions: see Supplementary Materials Table SM.9). 

The results of the investigation (see \cref{fig:NumberOfTrainingSamplesAndAugmentation}) clearly show that image augmentation can reduce the required number of training images. If image augmentation is used, the original validation loss threshold of the \gls{DPN} (at 300 training images) is already reached at 150 training images and the new validation loss threshold is encountered already at 200 images and is approximately \SI{8}{\percent} lower. Therefore, image augmentation was used for all further experiments.

\paragraph{Final Training} The final \gls{DPN}, which was used to generate the results presented in \cref{sec:Results}, was designed and trained based on the insights of \cref{sec:DeepParticleNet}. However, the early stopping patience was increased to 20, to further improve the minimum validation loss of the model. Additionally, three training runs were carried out and the model featuring the absolute minimum validation loss ($\gls{symb:loss_val}=0.955$) was selected. Overall, the minimum validation loss was reduced by approximately \SI{46}{\percent}, in comparison to the worst-case scenario\footnote{ResNet-101 backbone with randomly initialized weights, constant default learning rate, 400 training samples and no image augmentation (see \cref{fig:Resnet-Comparison-50vs101}).}, by careful choice of the backbone and the application of the training strategies presented within this section. 

	% !TeX spellcheck = en_US
\section{Results}
\label{sec:Results}
To asses the eligibility of the proposed method for image-based particle size measurements, the final \gls{DPN}\footnote{The source code of the specific version of the \gls{DPN} toolbox, the final model and the training, validation and test data sets, used for the training of the \gls{DPN} as well as the generation of the results presented in this section are available via the following link:\\ \url{https://github.com/maxfrei750/DeepParticleNet/releases/v1.0}\\\\Additionally, the training, validation and test data sets are part of the BigParticle.Cloud (\url{https://bigparticle.cloud}).} (training coditions: see Appendix \cref{app:tab:FinalTrainingConditions}) was tested on 10 different test sets, consisting of real \gls{SEM} images (see \cref{sec:Training-Validation-and-Test-Data}). In the process, the detection quality, the accuracy of the \gls{PSD} measurement as well as the analysis speed was surveyed. Additionally, the performance of the proposed method was compared to those of two prominent, already established methods -- the MATLAB version of the \gls{HT} \cite{MATLAB.2018b} and the ImageJ \gls{PS} plug-in \cite{Schneider.2012,Wagner.2017} -- in the aforementioned disciplines.

The non-default parameters of the \gls{HT} and the ImageJ \gls{PS} that were used during the comparison are given in the Supplementary Materials, Tables SM.11 and SM.12, respectively.

\subsection{Solidity as a Measure for the Sintering Degree}
The main difference between the 10 test samples is their increasing sintering degree. Therefore, it is  necessary to quantify the sintering degree to study its influence on the performance of the three tested methods. For this purpose, the solidity \gls{symb:solidity} was utilized as an easily obtainable measure for the sintering degree. It is defined as the ratio of the area \gls{symb:area}, i.e. the number of pixels, of an object to the area of its convex hull \gls{symb:area_convex} (see \cref{fig:AreaAndConvexArea}) \cite{MATLAB.2018a}: 

\begin{equation}
\gls{symb:solidity} = \frac{\gls{symb:area}}{\gls{symb:area_convex}}
\end{equation}

\nImageColumns = 3
\setlength{\spacerwidth}{1mm}
\setlength{\imagewidth}{(\figurewidth-\spacerwidth*(\nImageColumns-1))/\nImageColumns}

\begin{figure}
	\setlength{\tabcolsep}{0mm} % 
	\centering
	\begin{tabularx}{\figurewidth}{ZZ}
		\begin{subfigure}{\imagewidth}
			\includegraphics[width=\textwidth]{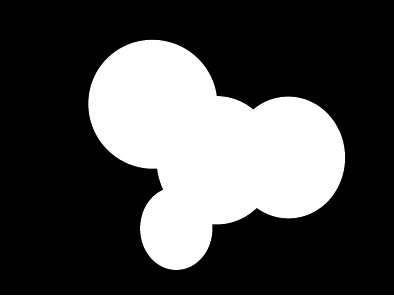}
			\caption{}
		\end{subfigure}
		&
		\begin{subfigure}{\imagewidth}
			\includegraphics[width=\textwidth]{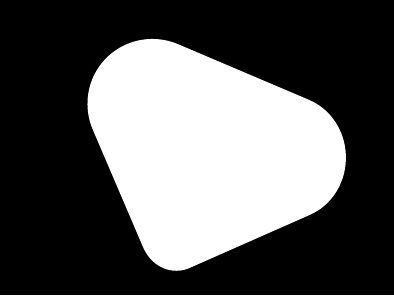}
			\caption{}
		\end{subfigure}
		\\
	\end{tabularx}
	\caption{Area \gls{symb:area} (a) and convex area \gls{symb:area_convex} (b) of an agglomerate.}
	\label{fig:AreaAndConvexArea}
\end{figure}
\begin{figure}
	\centering
	\includegraphics[width=\figurewidth]{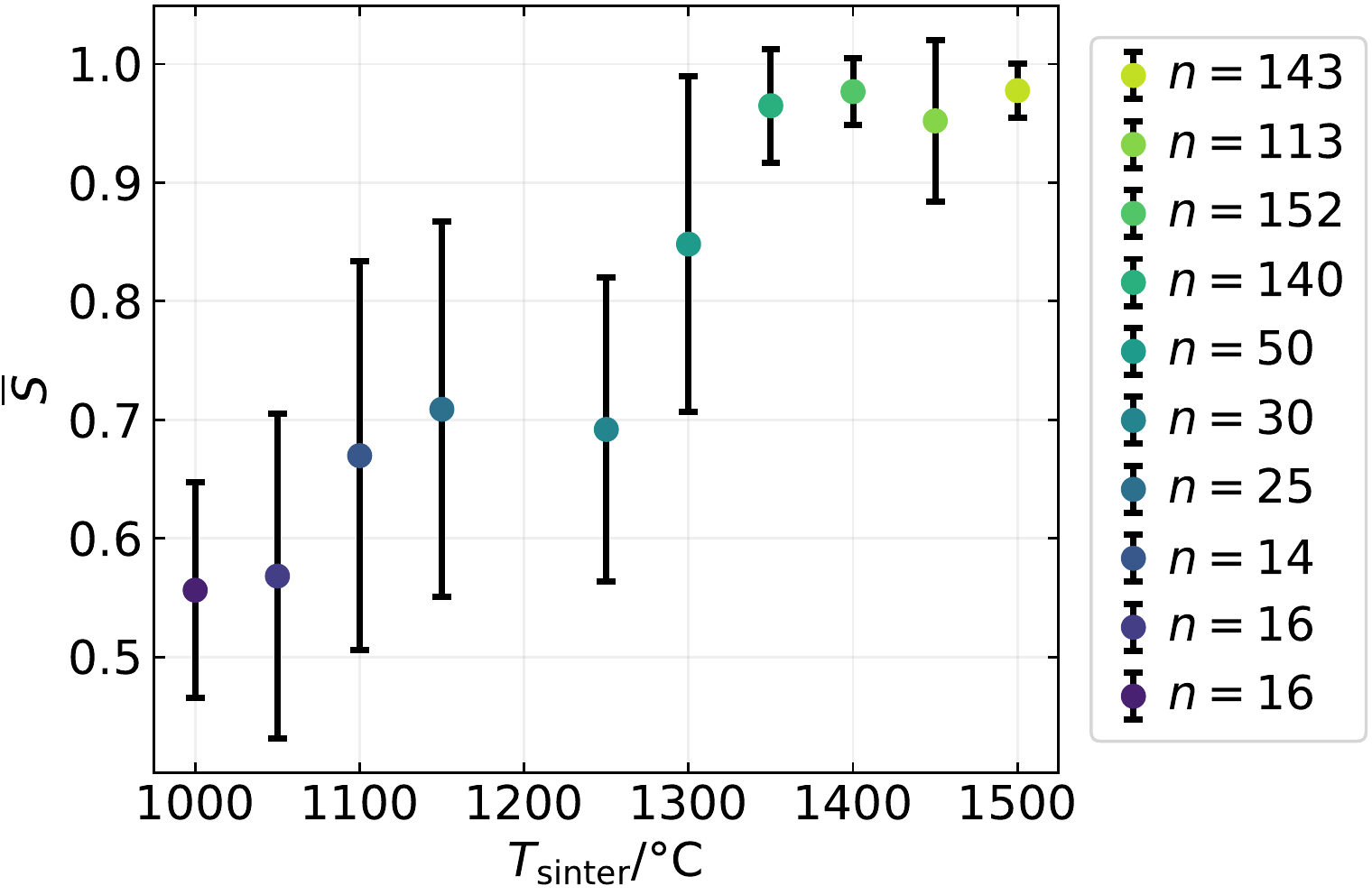}
	\caption{Mean solidities $\overline{\gls{symb:solidity}}$ of the agglomerates of the samples of the test set versus the respective sintering temperatures \gls{symb:temperature_sinter}. \gls{symb:number_dates} is the number of dates, i.e. the number of agglomerates in the respective sample, used for the calculation of the mean value. Error bars represent $\pm\gls{symb:standard_deviation}$.}
	\label{fig:Solidity-vs-Temperature}
\end{figure}

\cref{fig:Solidity-vs-Temperature} depicts the mean solidities of the agglomerates of the 10 test sets versus the respective sintering temperatures. As expected, the solidity increases with rising temperatures until it cannot increase any further because all primary particles are already sintered.

\subsection{Detection}
For some applications, e.g. structure analysis, not only the reliable detection of the primary particle sizes but also of their position is elemental. Therefore, the detection quality of the proposed method was assessed both qualitatively and quantitatively.
 
\subsubsection{Qualitative Evaluation}
\cref{fig:Detection-Comparison} depicts a comparison of original images from the 10 test sets and the corresponding detections by the proposed method, the \gls{HT} and the ImageJ \gls{PS}. Although a comparison based on images can only be qualitative, there are already some important insights to be gained.

% Large figure
% !TeX spellcheck= en_US
\setlength{\firsttablecolumnwidth}{4mm}
\nImageColumns = 4
\setlength{\spacerwidth}{1mm}
\setlength{\imagewidth}{(\textheight-\spacerwidth*(\nImageColumns-1)-\firsttablecolumnwidth)/\nImageColumns}

\begin{sidewaysfigure*}
	\centering
	\setlength{\tabcolsep}{0mm} % 
	\begin{tabularx}{\textwidth}{m{\firsttablecolumnwidth}*\nImageColumns{>{\centering\arraybackslash}m{\imagewidth+\spacerwidth}}}
		&
		Original &
		Proposed Method &
		Hough Transformation &
		ImageJ ParticleSizer \\
		
		\rotatebox{90}{Sample 1} &
		\includegraphics[width=\imagewidth]{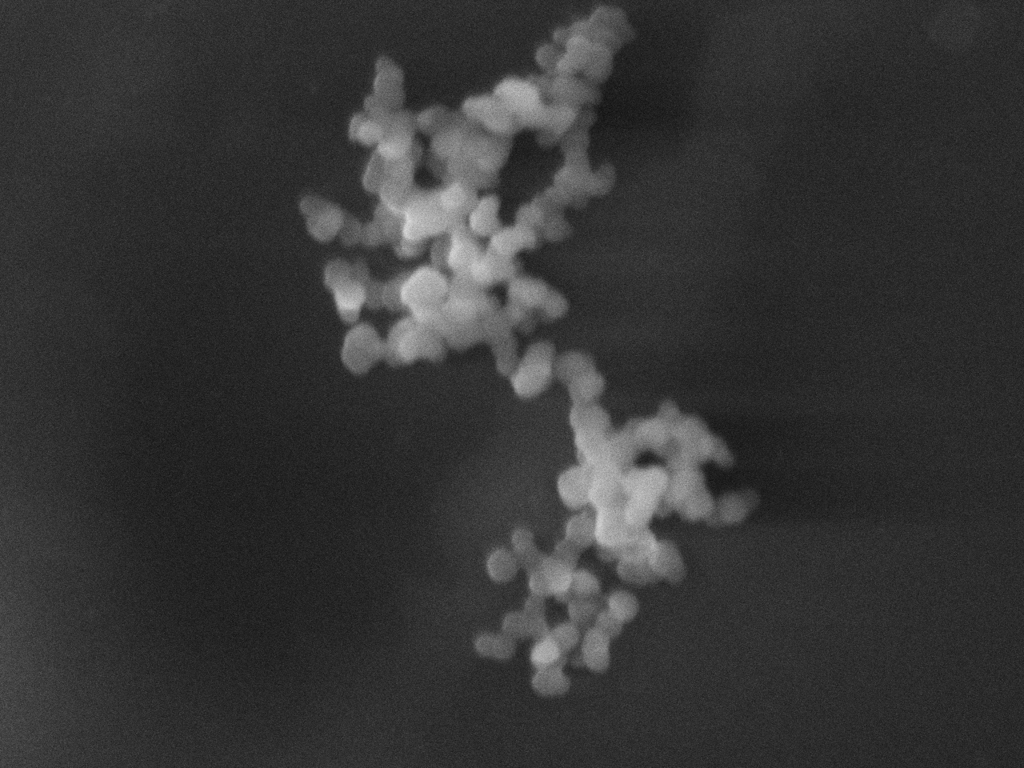} &
		\includegraphics[width=\imagewidth]{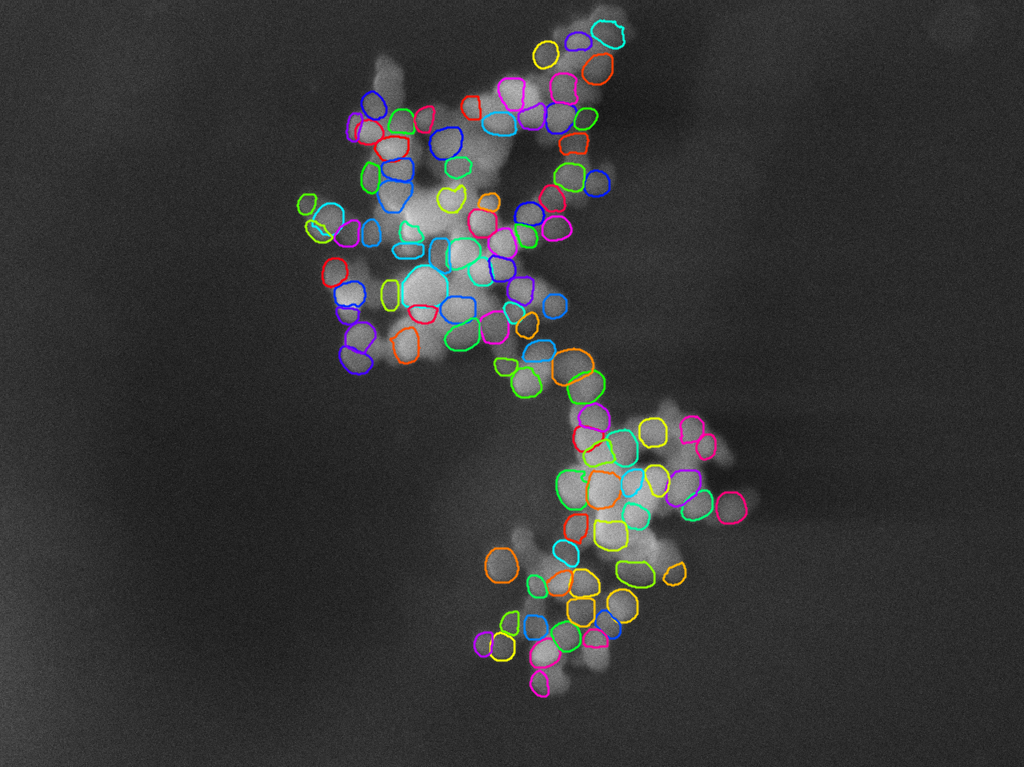} &
		\includegraphics[width=\imagewidth]{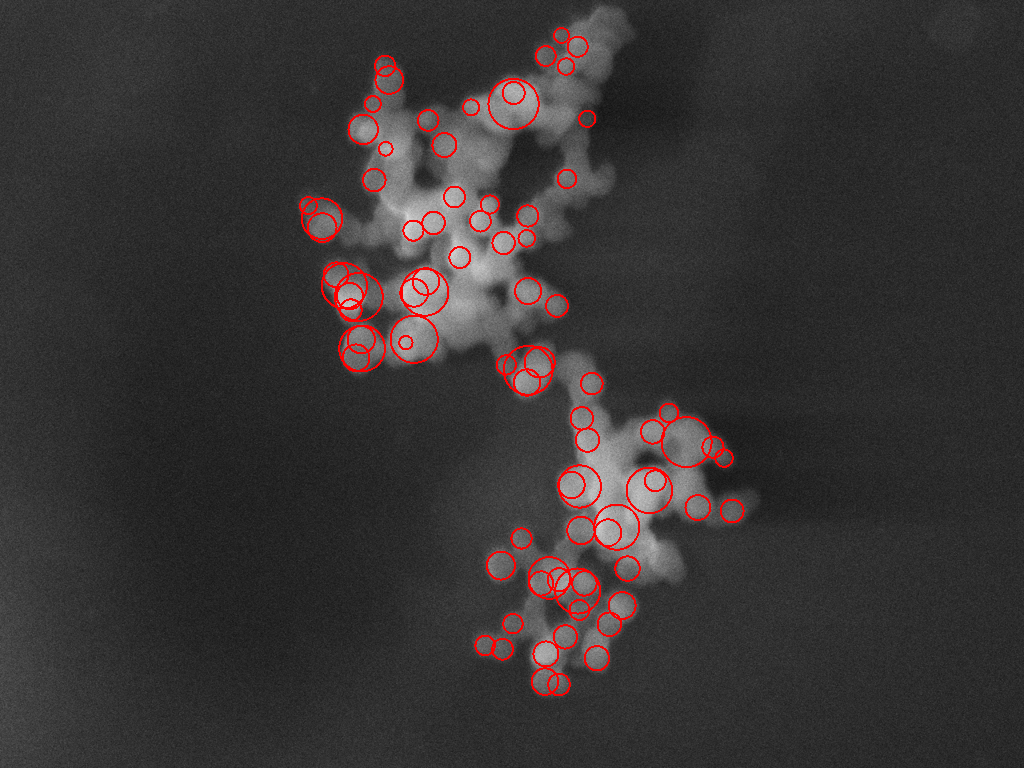} &
		\includegraphics[width=\imagewidth]{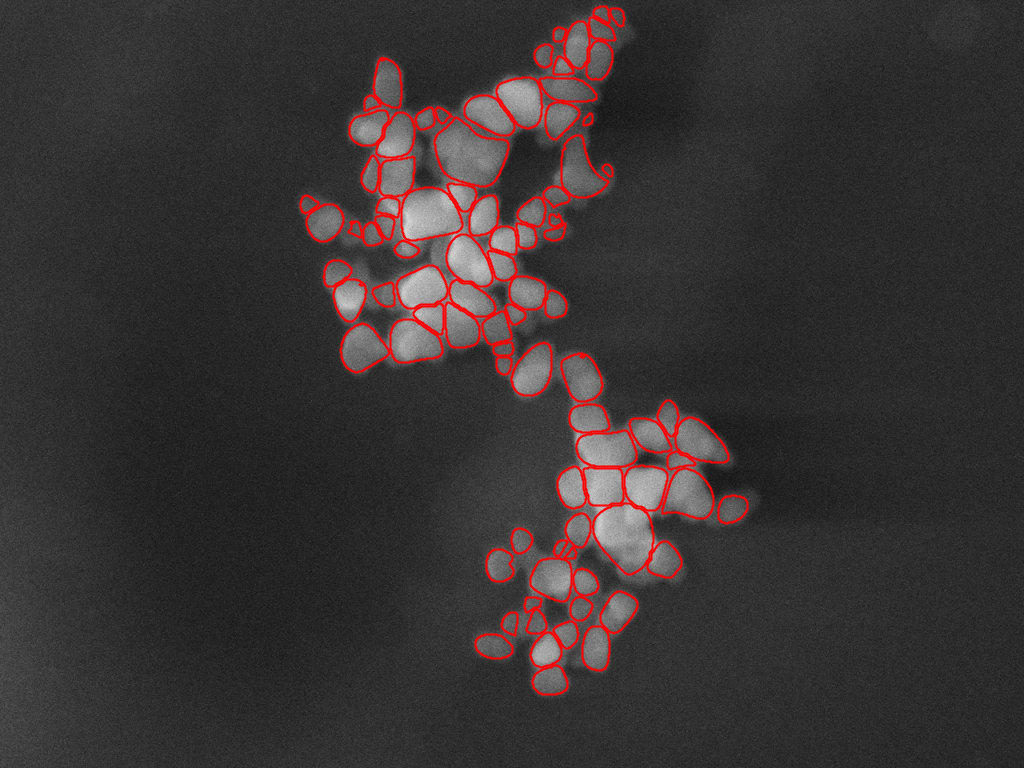} \\

		\rotatebox{90}{Sample 4} &
		\includegraphics[width=\imagewidth]{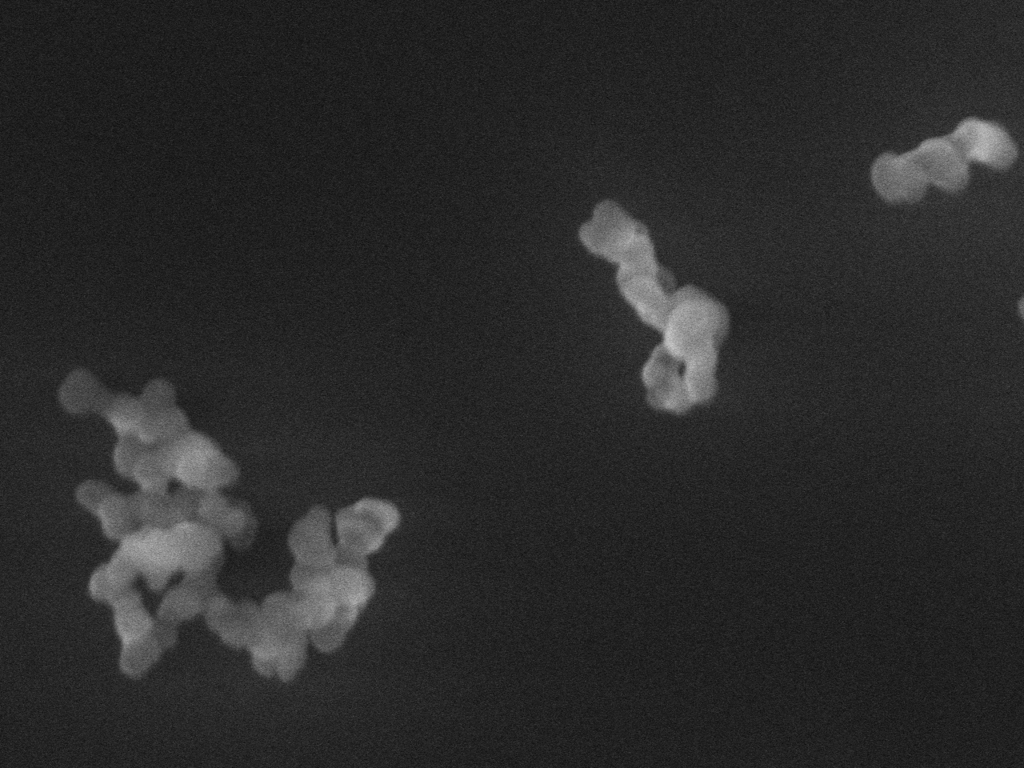} &
		\includegraphics[width=\imagewidth]{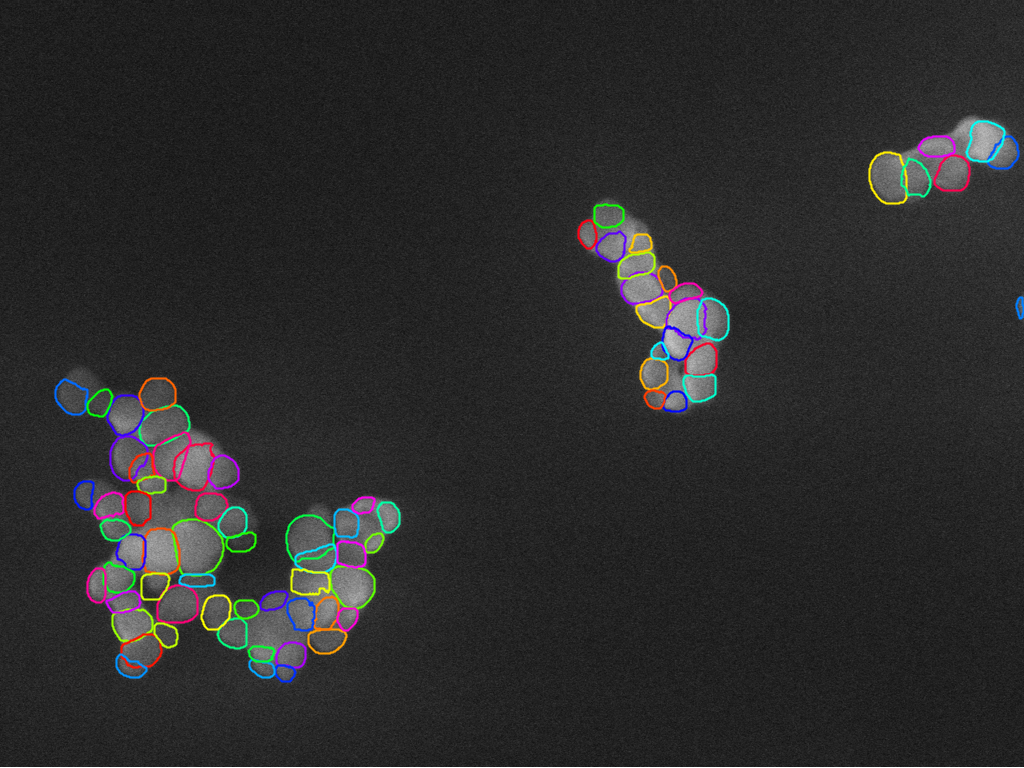} &
		\includegraphics[width=\imagewidth]{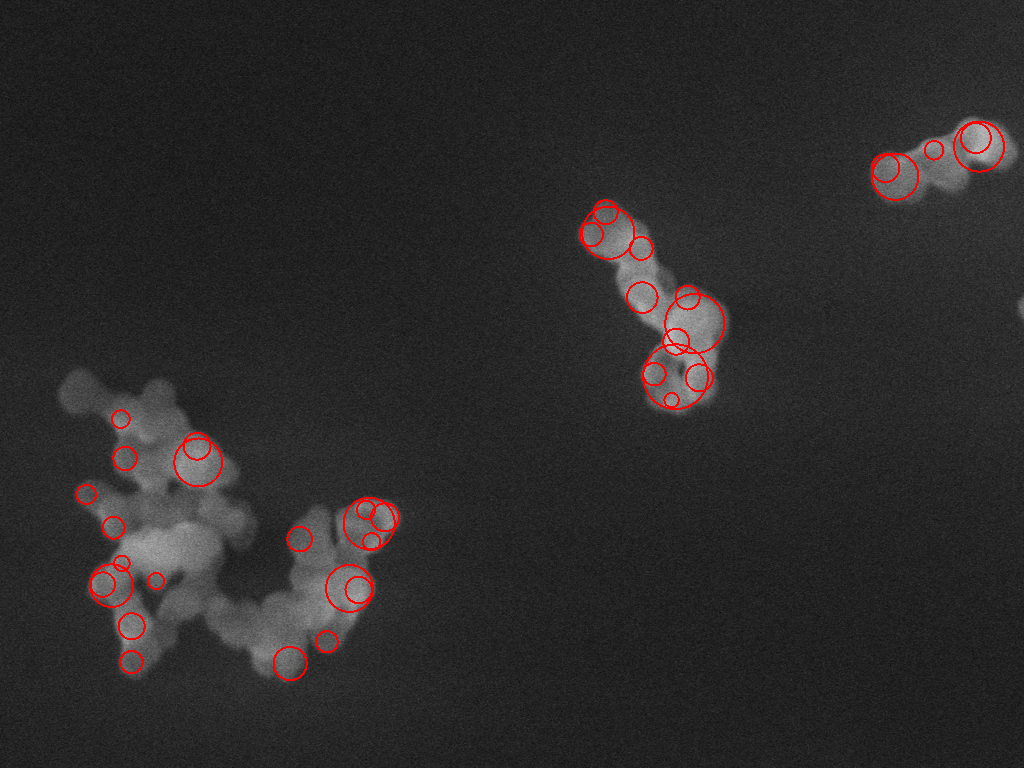} &
		\includegraphics[width=\imagewidth]{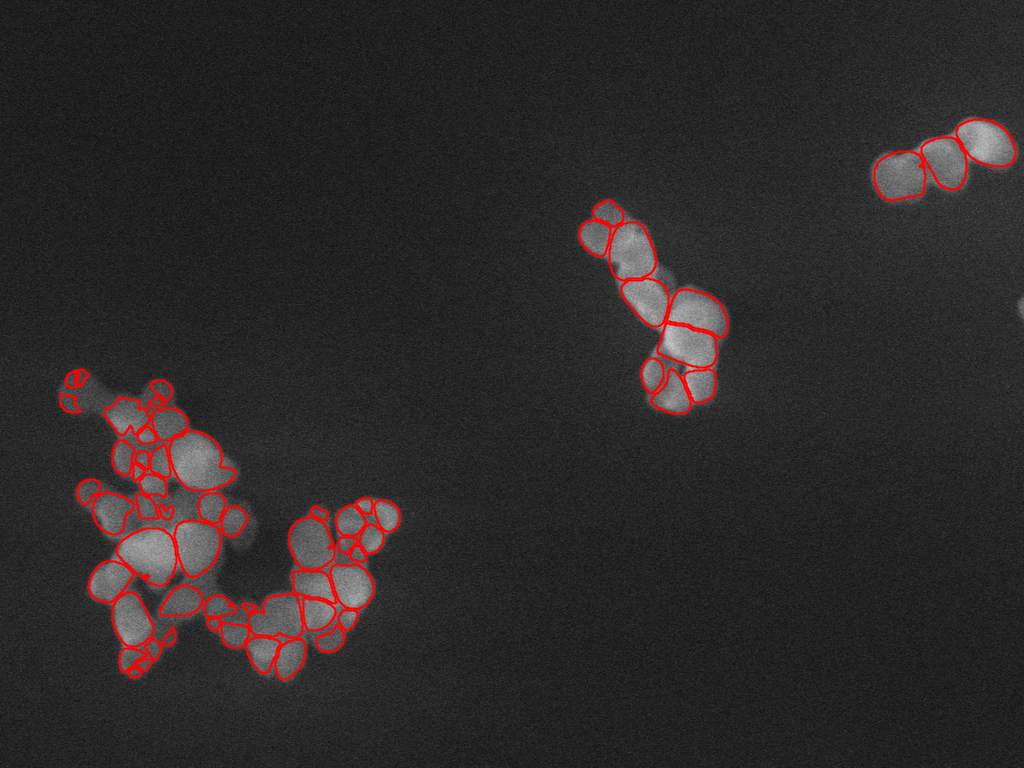} \\
		
		\rotatebox{90}{Sample 6} &
		\includegraphics[width=\imagewidth]{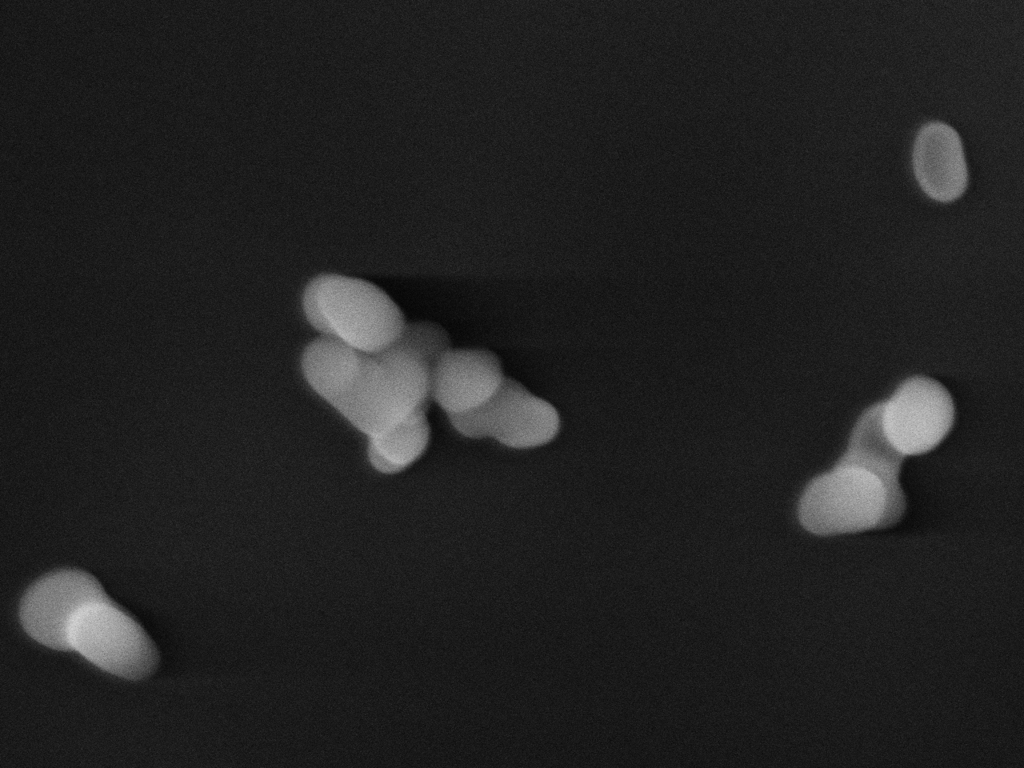} &
		\includegraphics[width=\imagewidth]{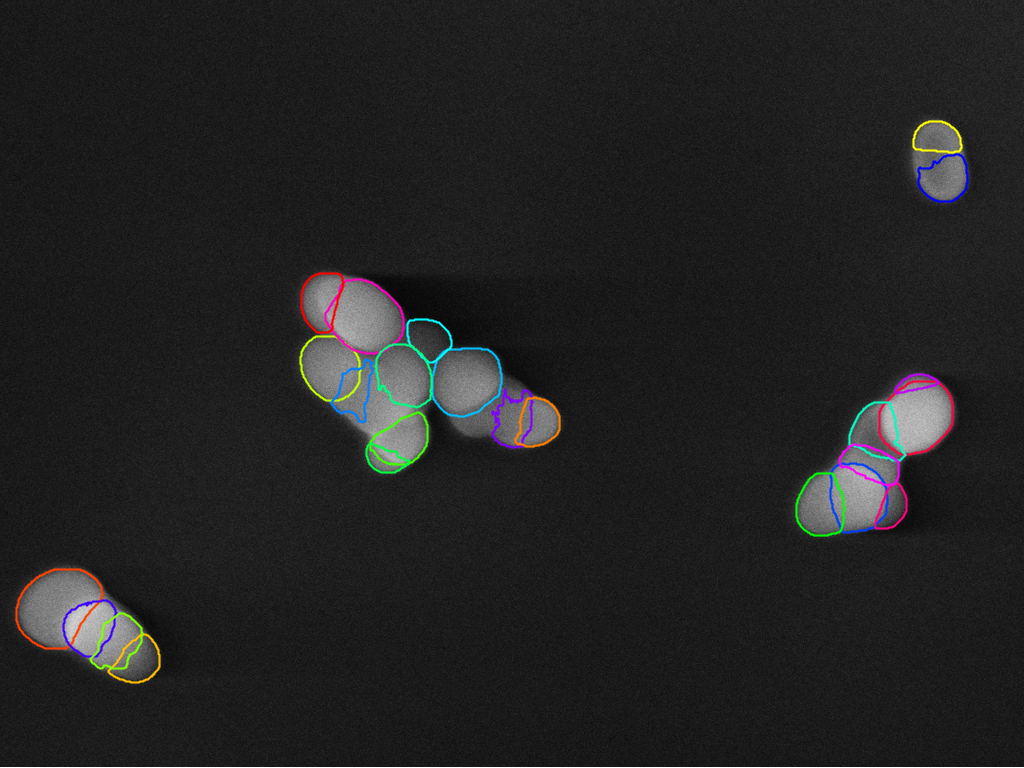} &
		\includegraphics[width=\imagewidth]{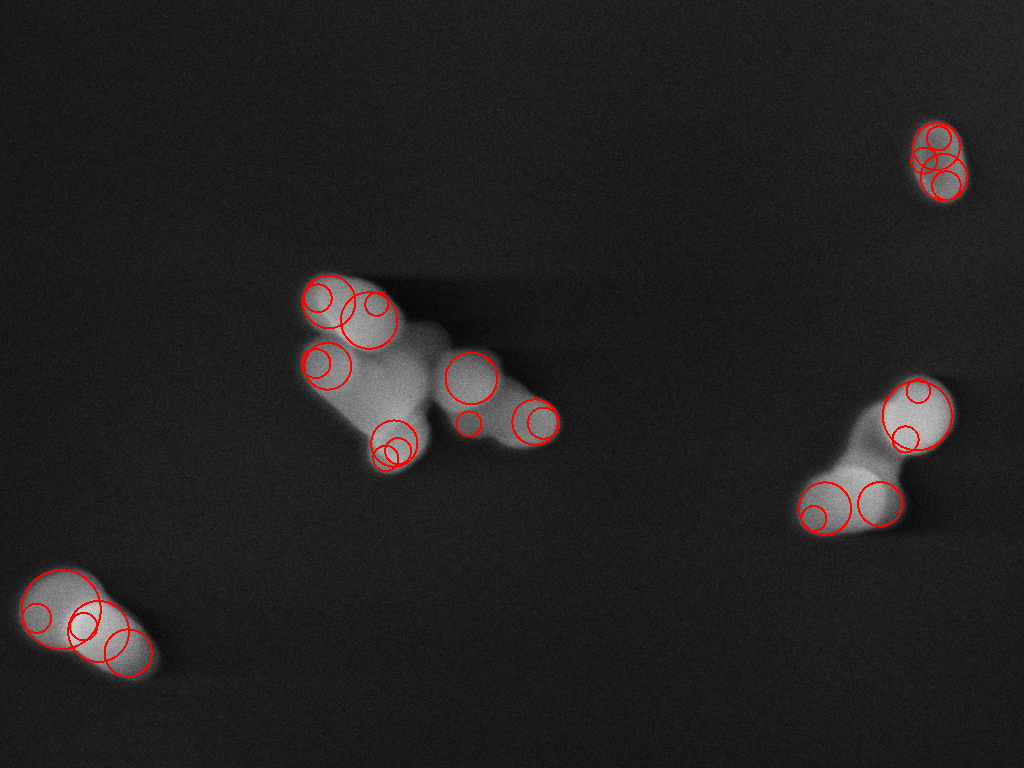} &
		\includegraphics[width=\imagewidth]{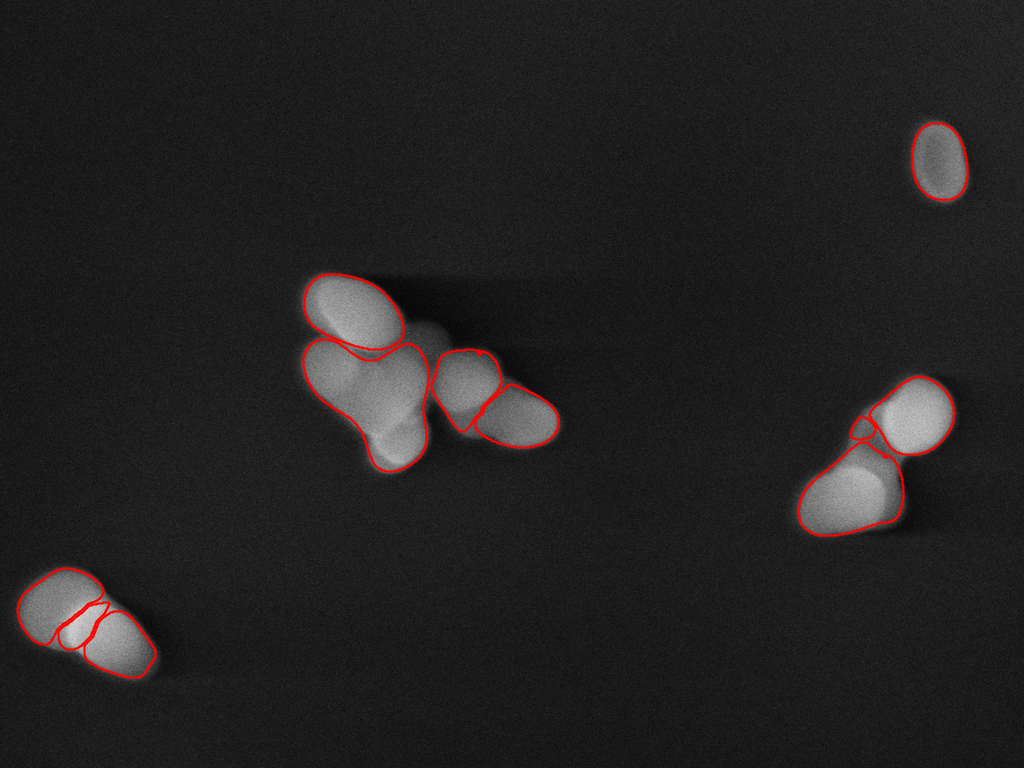} \\
		
		\rotatebox{90}{Sample 10} &
		\includegraphics[width=\imagewidth]{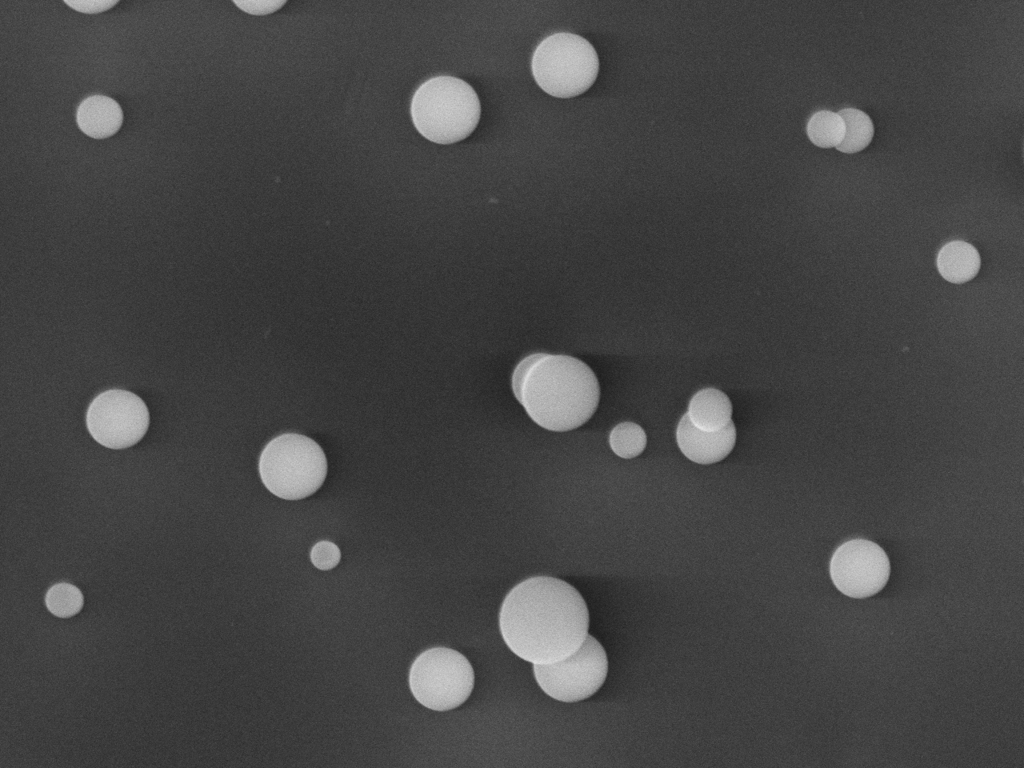} &
		\includegraphics[width=\imagewidth]{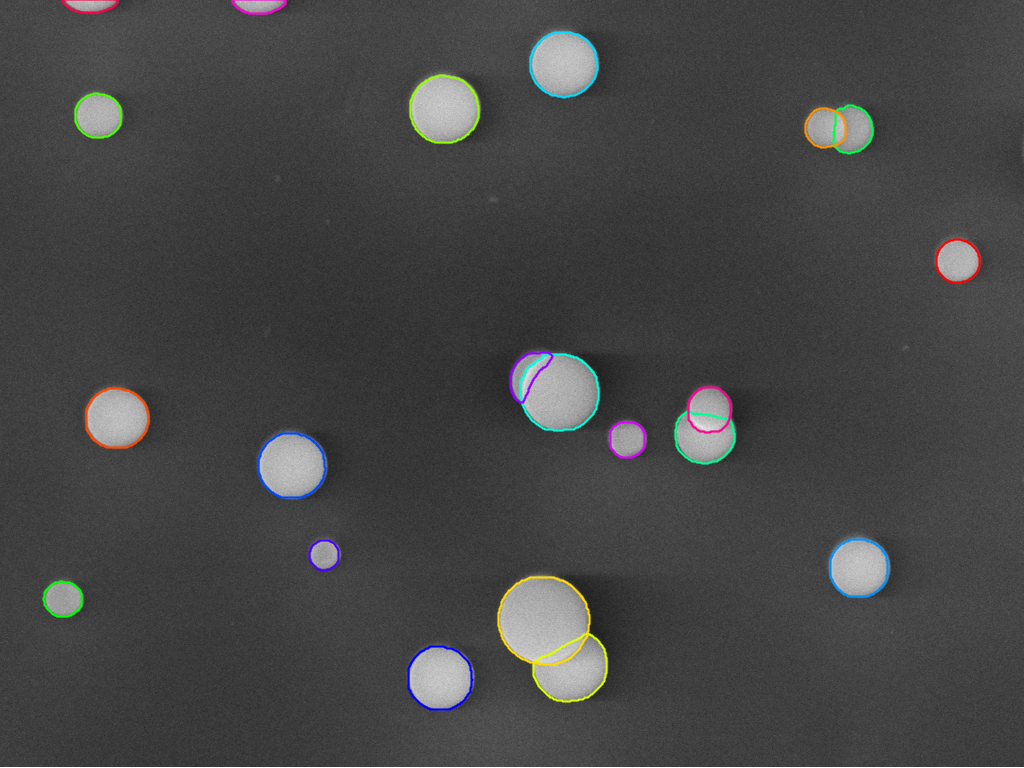} &
		\includegraphics[width=\imagewidth]{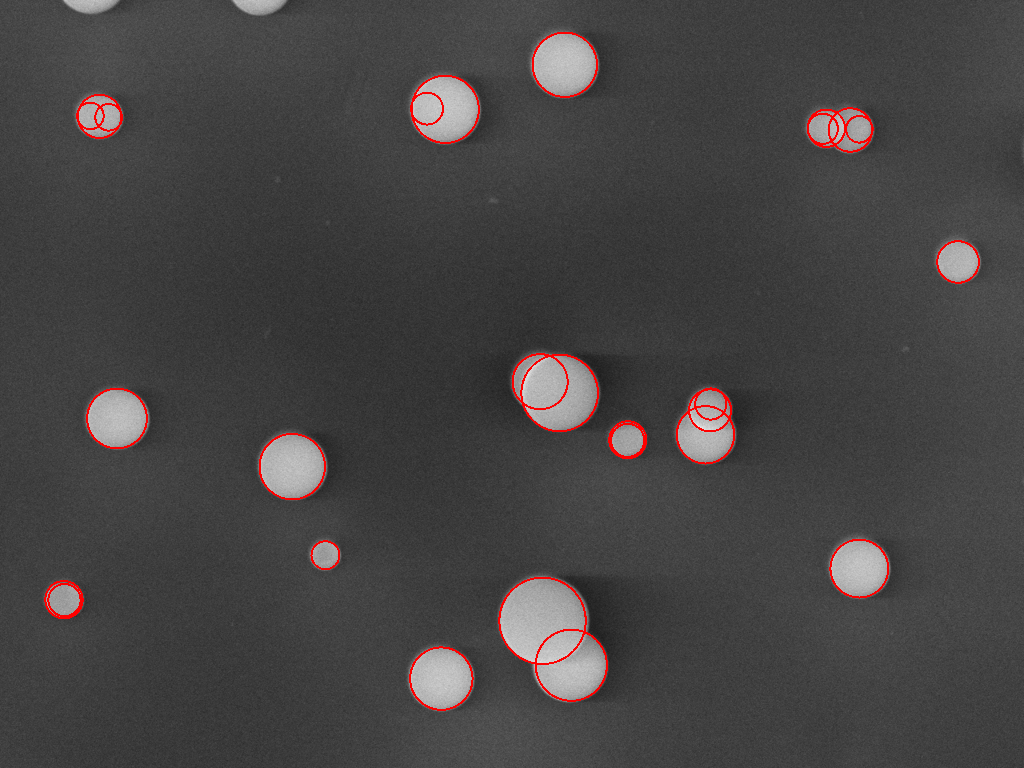} &
		\includegraphics[width=\imagewidth]{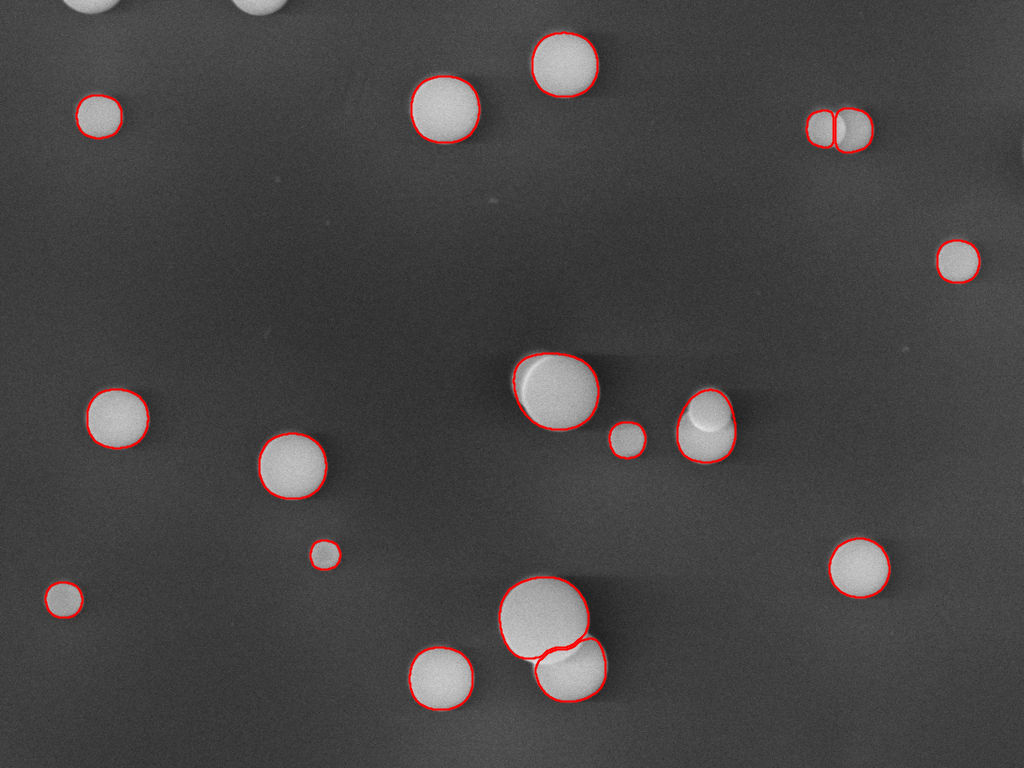} \\
	\end{tabularx}
	\caption{Comparison of randomly picked original images and the corresponding detections of the proposed method, the \gls{HT} and the ImageJ \gls{PS} from 4 of the 10 test sets. For a complete overview of the 10 test sets please refer to the Supplementary Materials (Figure SM.5).}
	\label{fig:Detection-Comparison} 
\end{sidewaysfigure*}

\paragraph{Human Perspective} Even as a human, it is difficult to decide where individual primary particles begin and end. This applies especially to the samples \numrange{1}{7}, which feature a lot of sintering necks. Therefore, there is a very important insight, to keep in mind for the later comparison of the tested methods, when the manual evaluation is used as a reference: For some samples, there is no such thing as a definitive ground truth.

\paragraph{Proposed Method} The proposed method marks only the visible parts of primary particles and yields very plausible results for all the samples. A few primary particles that feature a strong blur or a very low contrast are omitted by the analysis. The detected primary particles feature a medium to high amount of circularity, which is plausible, considering the nature of the sintering process. 

\paragraph{\glsentrytitlecase{HT}{long}} The \gls{HT} omits many primary particles when being confronted with rather fractal agglomerates (see samples \numrange{1}{6}). Due to the underlying measuring principle, it only yields perfectly circular detections, which often feature a high degree of overlap.

\paragraph{\glsentrylong{PS}} The most apparent difference of the detections of the ImageJ \gls{PS} to those of the other two tested methods is the fact that particles touching the border of the image are ignored during the analysis. Unfortunately, this feature cannot be disabled in the options. Apart from that, the ImageJ \gls{PS} plausibly fills the area of the agglomerates with detections and exhibits very few omittances and no overlap. However, the detections are often quite bulky, non-circular and concave, which is implausible, considering the nature of the sintering process.

\paragraph{Intermediate Result}
All three methods offer a reliable detection of primary particles for the samples \numrange{8}{10}, which are already highly sintered and therefore very circular. For the samples \numrange{1}{7}, based on the first impression, the proposed method seems to offer the most reliable, intuitive and physically plausible detection, followed by the ImageJ \gls{PS} and the \gls{HT}. Due to the fact that these impressions can only be qualitative, the detection quality will be quantified within the next section.

\subsubsection{Quantitative evaluation}
\begin{figure}
	\centering
	\includegraphics[width=\figurewidth]{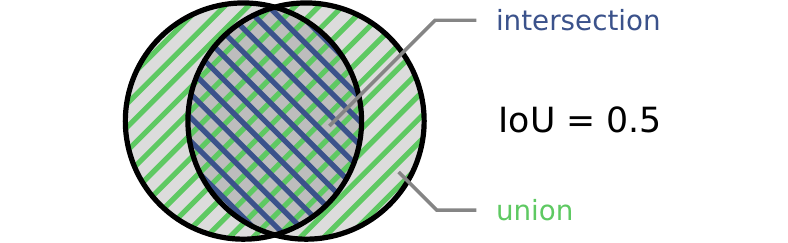}
	\caption{Illustration of the intersection over union metric.}
	\label{fig:IntersectionOverUnion}
\end{figure}

A common metric for the detection quality is the average precision, which is basically the percentage of correct detections. The criterion, whether a detection can be considered correct is usually the intersection over union (\gls{symb:intersection_over_union}) of the object to be detected and the predicted detection \cite{Lin.2014}. As the name implies, the \gls{symb:intersection_over_union} is the ratio of the areas of the intersection and the union of object and prediction (see \cref{fig:IntersectionOverUnion}). Therefore, the average precision can be defined at multiple \gls{symb:intersection_over_union} thresholds. Commonly used thresholds are $\gls{symb:intersection_over_union}=0.5$ and $\gls{symb:intersection_over_union}=0.75$, which yield $\gls{symb:average_precision}_{50}$ and $\gls{symb:average_precision}_{75}$, respectively. To lessen the impact of the \gls{symb:intersection_over_union} threshold, the mean of the average precision at multiple \gls{symb:intersection_over_union} thresholds, usually in the stepped range [\num{0.50}:\num{0.05}:\num{0.95}], can be calculated, which is simply referred to as \gls{symb:average_precision} \cite{COCOConsortium.2019}.

\begin{figure}
	\centering
	\includegraphics[width=\figurewidth]{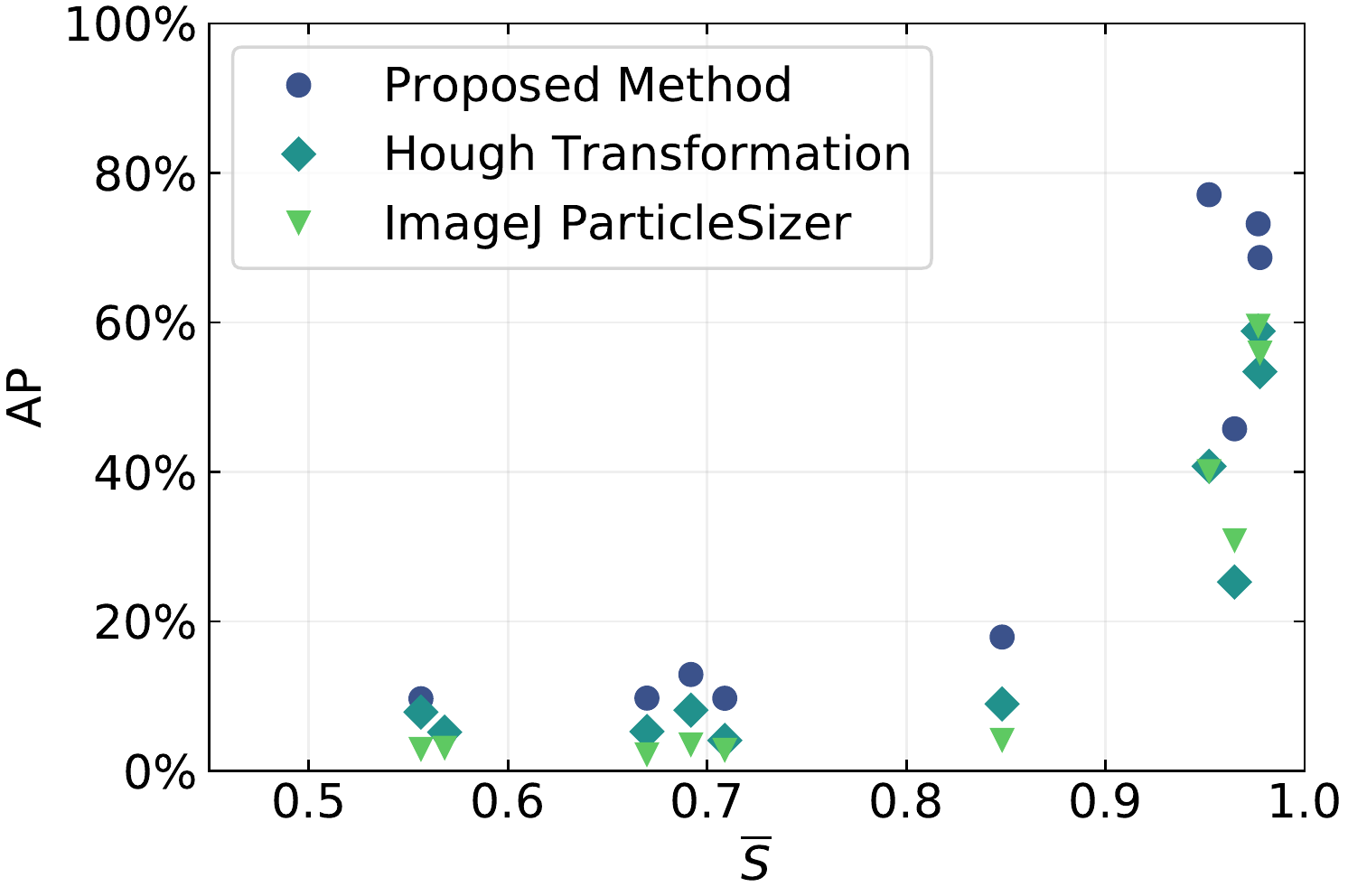}
	\caption{Average precisions \gls{symb:average_precision} retrieved via the proposed method, the \gls{HT} and the ImageJ \gls{PS} versus the mean solidities $\overline{\gls{symb:solidity}}$ of the 10 test samples.}
	\label{fig:AveragePrecisionComparison}
\end{figure}

\cref{fig:AveragePrecisionComparison} depicts a comparison of the average precisions of the three tested methods for the 10 test samples (for comparisons of the $\gls{symb:average_precision}_{50}$ and $\gls{symb:average_precision}_{75}$, please see Supplementary Materials Figure SM.4). As ground truth, the results from a manual analysis were used.
The quantitative evaluation of the detection quality confirms the qualitative impressions from the previous section. For all 10 test samples, the proposed method outperforms the other two tested methods. Just as it might be expected, test samples with lower solidities are far more challenging for the tested methods than samples featuring higher solidities.

\subsection{Primary \glsentrytitlecase{PSD}{long} Measurement}
Depending on the application, the capability of the three tested methods to correctly reproduce the underlying primary \glspl{PSD} of the test samples is more important than the perfect detection of every primary particle. Therefore, the primary \glspl{PSD}, as measured by each of the three methods were compared to the manually determined primary \gls{PSD}, for each of the 10 test samples (see Supplementary Materials Figure SM.6).

As equivalent diameter, the maximum Feret diameter was used as a compromise between the perfectly circular detections of the manual analysis and the \gls{HT} on the one hand and the irregularly shaped detections of the proposed method and the ImageJ \gls{PS} on the other hand.

\paragraph{Proposed Method} The \glspl{PSD} retrieved via the proposed method agree very well with the manually determined \glspl{PSD} for all the samples except for sample~6. However, this sample is very challenging for the \gls{HT} and the ImageJ \gls{PS} as well. Of the three tested methods, the proposed method still shows the highest agreement with the manual analysis results.

\paragraph{\glsentrytitlecase{HT}{long}} The \gls{HT} is clearly biased towards a primary particle size of approximately \SIrange{25}{30}{\px}, which results in rather strong deviations from the manual analysis. Unfortunately, the reasons for this behavior remain unclear.

\paragraph{\glsentrylong{PS}} The ImageJ \gls{PS} shows a good agreement with the manual analysis for the samples \numrange{8}{10}, i.e. samples which mainly feature agglomerates with small numbers of rather circular primary particles. For the other samples, the \glspl{PSD} retrieved via the ImageJ \gls{PS} differ from the manually determined \glspl{PSD}, especially with respect to the geometric standard deviation.

\paragraph{Intermediate Result} The proposed method clearly outperforms the ImageJ \gls{PS} and the \gls{HT}. Particularly its consistently high reliability across nearly all the test samples is remarkable.

\subsubsection{Accuracy versus Sintering Degree}
To survey the impact of the sintering degree on the \gls{PSD} measurement accuracy of the three tested methods, the geometric mean diameter \gls{symb:geometric_mean_diameter}, the geometric standard deviation \gls{symb:geometric_standard_deviation} and the primary particle number \gls{symb:number_primary_particles} were determined for the \glspl{PSD} of the 10 test samples, retrieved via the three tested methods. Subsequently, the percentage errors \gls{symb:geometric_mean_diameter_percentage_error}, \gls{symb:geometric_standard_deviation_percentage_error} and \gls{symb:number_primary_particles_percentage_error} of these three parameters were determined according to the following equation:
\begin{equation}
\gls{symb:percentage_error}=\frac{\gls{symb:actual_value}-\gls{symb:desired_value}}{\gls{symb:desired_value}} \cdot 100\%,
\end{equation}
where \gls{symb:index} is the sample index, \gls{symb:desired_value} is the desired value as determined via manual analysis and \gls{symb:actual_value} is the actual value retrieved via one of the tested methods.

\begin{figure}[p]
	\centering
	\begin{subfigure}{\figurewidth}
		\includegraphics[width=\figurewidth]{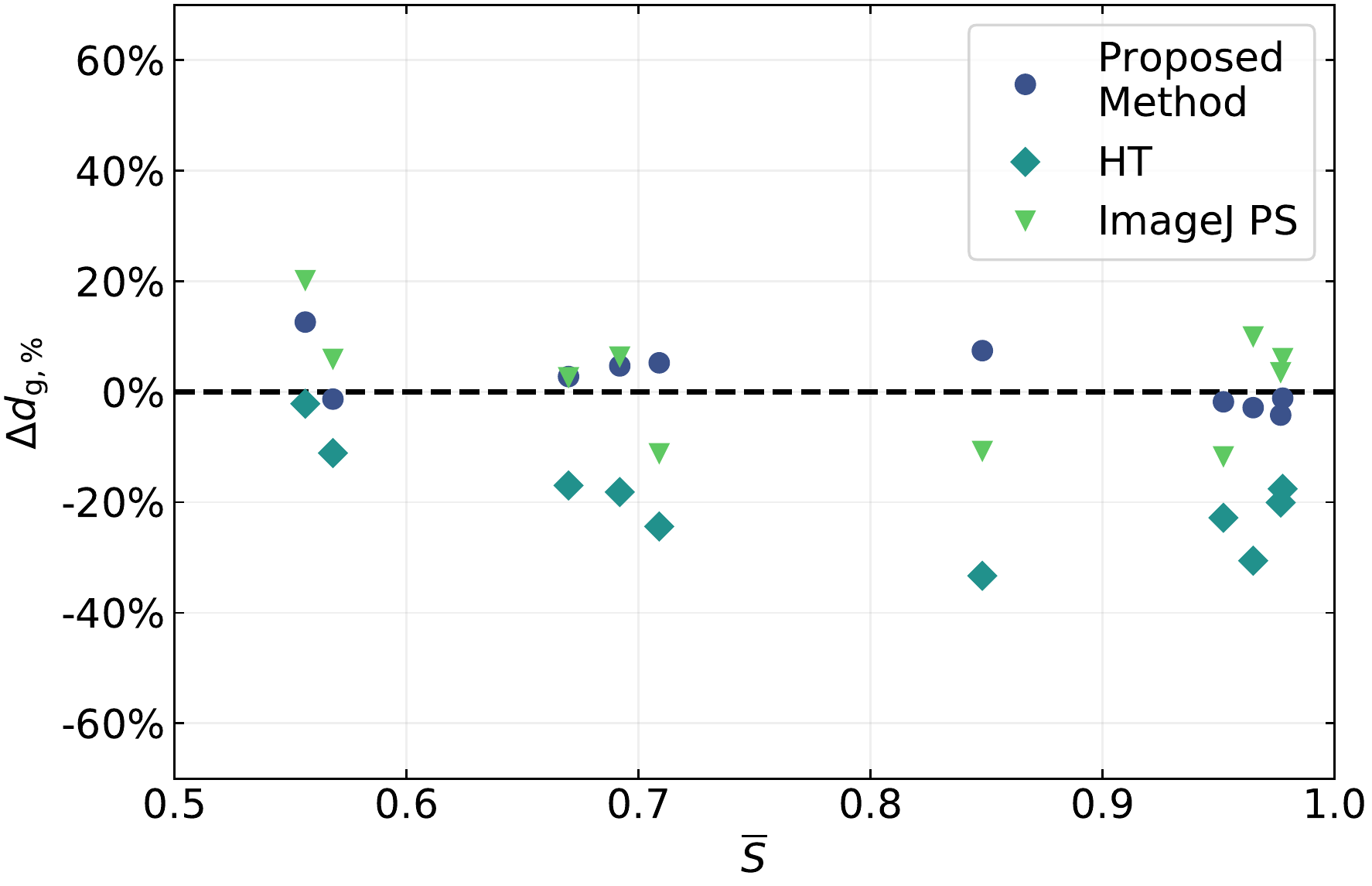}
		\caption{}
		\label{fig:d_g_vs_S_comparison}
	\end{subfigure}\\%
	\vspace{4ex}
	\begin{subfigure}{\figurewidth}
		\includegraphics[width=\figurewidth]{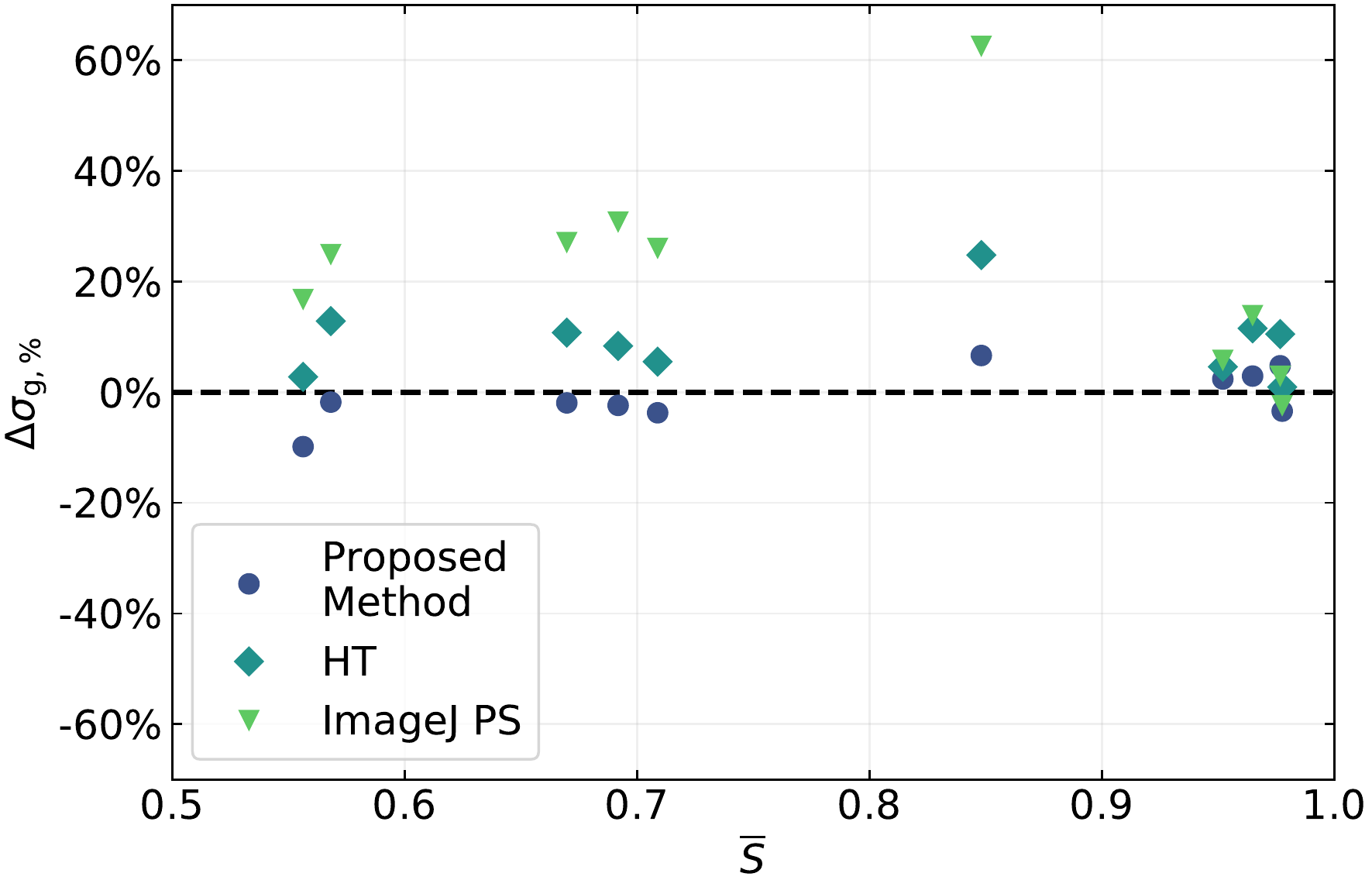}
		\caption{}
		\label{fig:s_g_vs_S_comparison}
	\end{subfigure}\\%
	\vspace{4ex}
	\begin{subfigure}{\figurewidth}
		\includegraphics[width=\figurewidth]{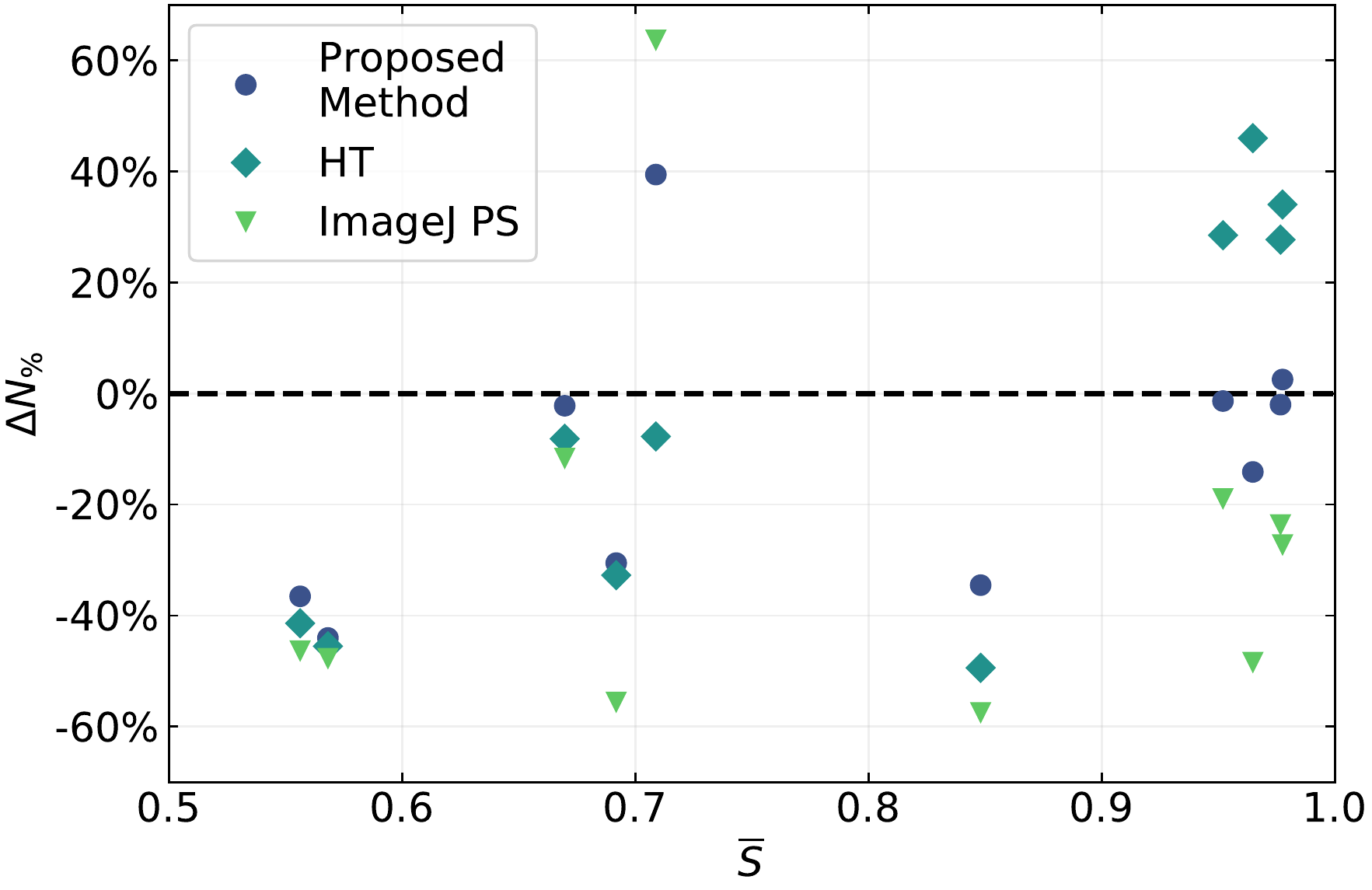}
		\caption{}
		\label{fig:N_vs_S_comparison}
	\end{subfigure}\\%
	\caption{Percentage errors of the geometric mean diameters \gls{symb:geometric_mean_diameter_percentage_error} (a), the geometric standard deviations \gls{symb:geometric_standard_deviation_percentage_error} (b) and the numbers of primary particles \gls{symb:number_primary_particles_percentage_error} (c) retrieved via the proposed method, the \gls{HT} and the ImageJ \gls{PS} versus the mean solidities $\overline{\gls{symb:solidity}}$ of the 10 test samples.}
	\label{fig:PSD_parameter_error_comparison}
\end{figure}

Finally, the percentage errors \gls{symb:geometric_mean_diameter_percentage_error}, \gls{symb:geometric_standard_deviation_percentage_error} and \gls{symb:number_primary_particles_percentage_error} of each of the three tested methods for each of the 10 test samples, were plotted versus the corresponding mean solidities $\overline{\gls{symb:solidity}}$ (see \cref{fig:PSD_parameter_error_comparison}).

\paragraph{Geometric Mean Diameter} The geometric mean diameters determined via the proposed method (see \cref{fig:d_g_vs_S_comparison}) exhibit only small deviations from the manual analysis (\SIrange{-4}{13}{\percent}), thereby outperforming the \gls{HT} (\SIrange{-33}{-2}{\percent}) and the ImageJ \gls{PS} (\SIrange{-12}{20}{\percent}) by a large margin. While the errors of the proposed method and the ImageJ \gls{PS} are symmetrical and mostly independent from the mean solidity, the \gls{HT} has a tendency to underestimate the geometric mean diameter, especially for samples of high mean solidity.

\paragraph{Geometric Standard Deviation} The percentage errors of the proposed method regarding the geometric standard deviation (\SIrange{-10}{5}{\percent}, see \cref{fig:s_g_vs_S_comparison}) are similar to those of the geometric mean diameter and thereby again clearly outperform the \gls{HT} (\SIrange{1}{25}{\percent}) and the ImageJ \gls{PS} (\SIrange{-2}{62}{\percent}). While the proposed method again exhibits rather symmetrical errors, the ImageJ \gls{PS} and the \gls{HT} both have a tendency to overestimate the geometric standard deviation. Furthermore, the impact of the mean solidity is small for the proposed method and the \gls{HT} but clearly noticeable for the ImageJ \gls{PS}.

\paragraph{Primary Particle Number} Also with respect to the determination of the number of primary particles (see \cref{fig:N_vs_S_comparison}), the proposed method exhibits smaller errors (\SIrange{-44}{39}{\percent}) than the \gls{HT} (\SIrange{-49}{46}{\percent}) and the ImageJ \gls{PS} (\SIrange{-58}{64}{\percent}). In general, the errors of all three tested methods are significantly larger for the determination of the primary particle number, than they are for the determination of the geometric mean diameter and the geometric standard deviation. Independent of the solidity, all three tested methods have a tendency to underestimate the primary particle number, except for the \gls{HT}, which overestimates the primary particle number for samples of large solidity. 

\paragraph{Intermediate Result} The survey of the impact of the solidity on the \gls{PSD} measurement capabilities of the three tested methods demonstrates that the proposed method clearly outperforms the ImageJ \gls{PS} as well as the \gls{HT} and is very robust towards the sintering degree of the analyzed samples.

\subsubsection{Overall Accuracy}
To allow for a better comparison of the accuracies of the three tested methods, it is useful to describe the error of each method with help of a single characteristic number per sought-after \gls{PSD} parameter (geometric mean diameter, geometric standard deviation and primary particle number) that gives an intuition of the performance of the tested methods across all 10 test sets. A suitable characteristic number for the given use case is the \gls{MAPE}, which is defined as \cite{Hyndman.2006}:
\begin{equation}
\gls{symb:MAPE} = \frac{1}{\gls{symb:number_dates}}\cdot\sum_{\gls{symb:index}=1}^{\gls{symb:number_dates}}|\gls{symb:percentage_error}|,
\end{equation}
where \gls{symb:index} is the sample index and \gls{symb:number_dates} is the number of dates, i.e. the number of test sets.

Advantageous properties of the \gls{MAPE} are its easy interpretability and the fact that negative and positive errors do not mutually compensate.

\begin{figure}
	\centering
	\includegraphics[width=\figurewidth]{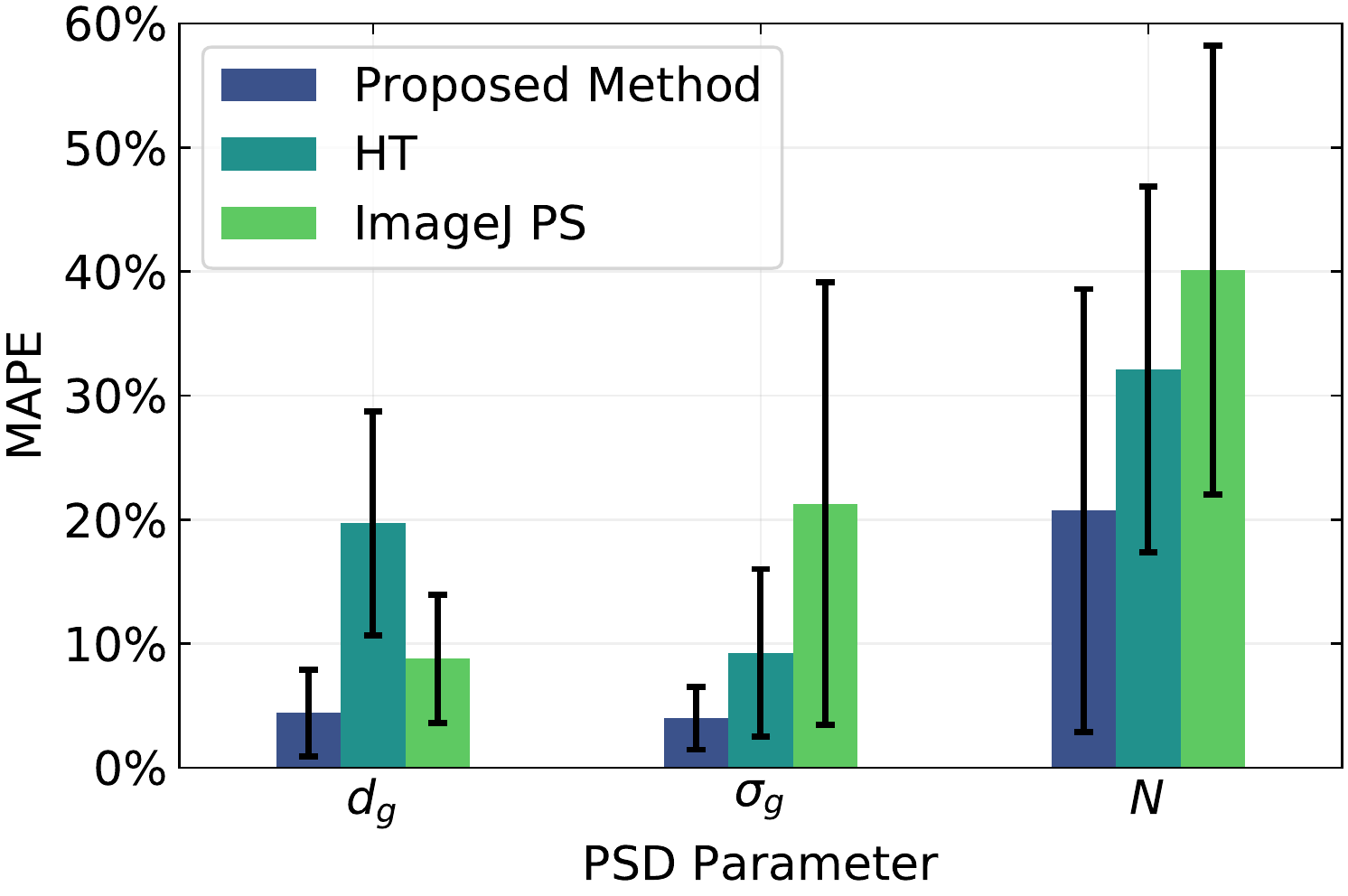}
	\caption{Mean absolute percentage errors \gls{symb:MAPE} of the \gls{PSD} parameters \gls{symb:geometric_mean_diameter}, \gls{symb:geometric_standard_deviation} and \gls{symb:number_primary_particles} across the 10 test sets, as retrieved via the proposed method, the \gls{HT} and the ImageJ \gls{PS} (error bars represent $\pm\gls{symb:standard_deviation}$ of the absolute percentage errors).}
	\label{fig:MAPE_comparison}
\end{figure}

\cref{fig:MAPE_comparison} depicts a comparison of the \glspl{MAPE} of the geometric mean diameter, the geometric standard deviation and the primary particle number, across all 10 test samples for the three tested methods. It can be clearly seen that the proposed method is not only superior to the \gls{HT} and the ImageJ \gls{PS} for each of the \gls{PSD} parameters, but also that it is the only tested method that is capable of a reliable determination of both the geometric mean diameter \emph{and} the geometric standard deviation. It thereby achieves accuracies at a near-human level (approximately \SI{4}{\percent} error for the geometric mean diameter and \SI{1}{\percent} error for the geometric standard deviation \cite{Frei.2018}).

\subsection{Analysis Speed}
\label{sec:AnalysisSpeed}

\begin{figure}
	\centering
	\includegraphics[width=\figurewidth]{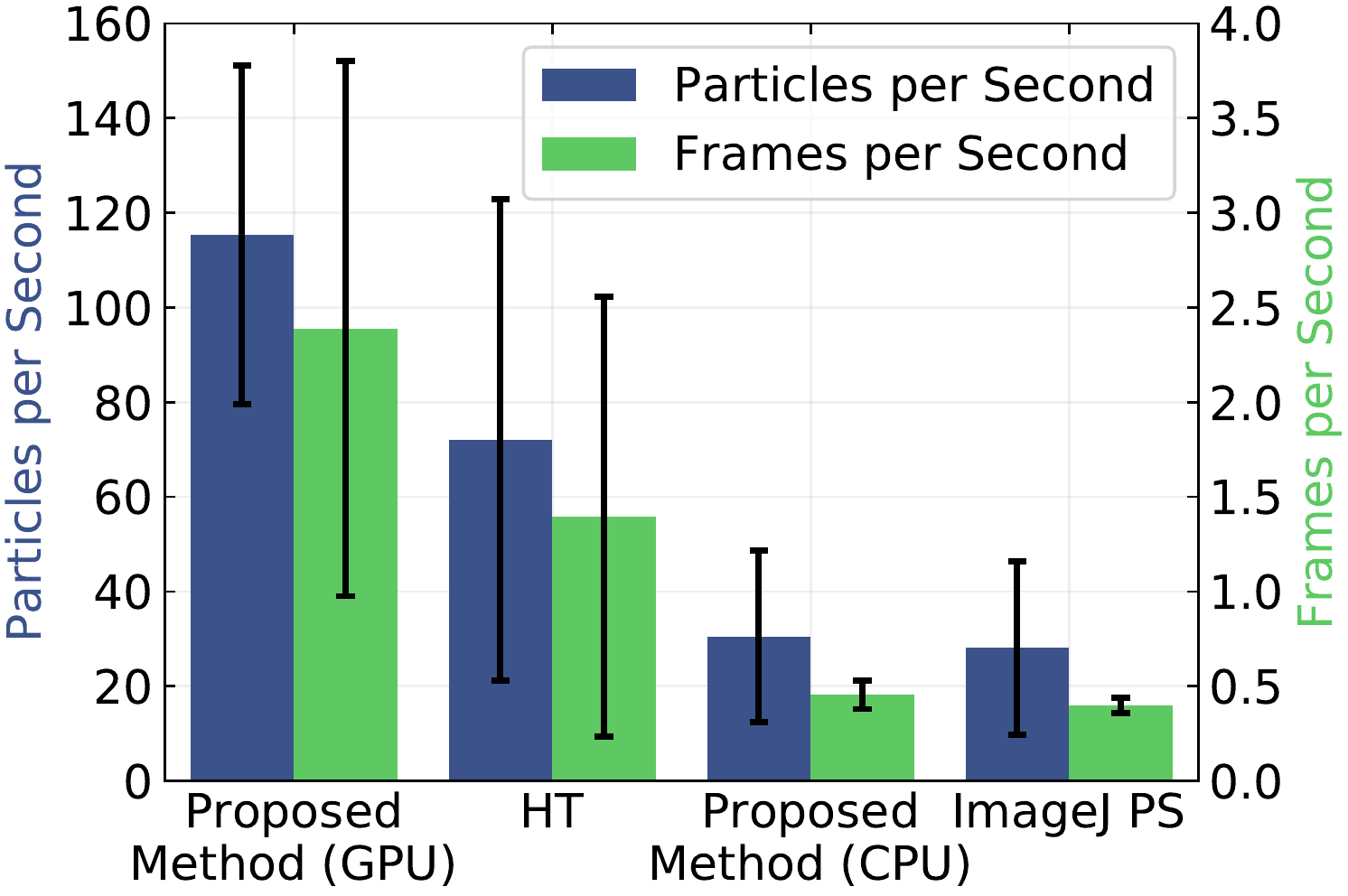}
	\caption{Analysis speeds of the proposed method, the \gls{HT} and the ImageJ \gls{PS}, on the one hand based on the number of analyzed images and on the other hand based on the number of analyzed particles (10 repetitions for each of the 10 test sets, error bars represent $\pm\gls{symb:standard_deviation}$).}
	\label{fig:analysis_speed}
\end{figure}

The analysis speed of the tested methods is a very important property, especially for the use in on-line measurement setups. Therefore, the analysis speed was surveyed by a tenfold measurement of the analysis times of each of the tested methods for each of the 10 test sets. Subsequently, the analysis speeds of the tested methods were calculated, on the one hand based on the number of images and on the other hand based on the number of primary particles in the respective test sets. Ultimately, the mean and the standard deviation of all measurements of each method were calculated and compared (see \cref{fig:analysis_speed}). As a specialty, the analysis speed of the proposed methods was measured twice, once when being run on the \gls{CPU} and once when being run on a single \gls{GPU}.\footnote{For the specifications of the test system, please refer to Appendix \cref{app:tab:Hardware,app:tab:Software}.}

Both, the analysis speed measurement based on the number of images as well as the analysis speed measurement based on the number of primary particles yield very similar results (see \cref{fig:analysis_speed}). The proposed method clearly outperforms the other tested methods when being run on a \gls{GPU}, the reason for its dominance being its extremely high degree of parallelism. When being run on a \gls{CPU}, the proposed method is still slightly faster than the ImageJ \gls{PS}. However, on a \gls{CPU}, both methods are clearly outperformed by the \gls{HT}. As expected, all of the tested automated methods outspeed human operators (\numrange{0.5}{0.1} particles per second) by a large margin. 
	% !TeX spellcheck = en_US
\section{Conclusion and Outlook}
Within this work, a novel, deep learning-based, method for the pixel-perfect detection and sizing of agglomerated, aggregated or occluded primary particles was proposed and validated. Therefore, the very powerful Mask R-\gls{CNN} architecture was carefully adapted to and trained on \gls{SEM} images of nanoparticles, using state-of-the-art training strategies, such as early stopping, transfer learning, a cyclical learning rate schedule and image augmentation. 

As a specialty, the training was carried out using only synthetic \gls{SEM} images, which were produced by the especially implemented \gls{synthPIC} toolbox, to avoid the laborious task of manual annotation and to increase the quality of the ground truth. Despite the training on synthetic images, the proposed method performed excellent on real world samples, belonging to 10 test sets, which resulted from a sintering study of silica nanoparticles and exhibited various sintering degrees and image characteristics, due to varying \gls{SEM} operators as well as changing sample and instrument conditions.

To evaluate the performance of the proposed method, it was compared to two established methods for the determination of primary \glspl{PSD}, the \gls{HT} and the ImageJ \gls{PS} plug-in. The proposed method clearly outperformed both of the other tested methods, exhibiting a \gls{MAPE} as low as \SI{4}{\percent} across all test sets, for both the geometric mean diameter and the geometric standard deviation. It thereby attains human-like performance and reliability.

Also the analysis speed of the proposed method was compared to those of the \gls{HT} and the ImageJ \gls{PS}. Once again, the proposed method clearly outperformed the other tested methods by a large margin, as long as it was run on a \gls{GPU}. However, it might not be fast enough for some image-based on-line measurement applications. Fortunately, the speed of the method can be increased nearly linearly by using multiple \glspl{GPU} in parallel. Nevertheless, future studies will concentrate on achieving higher analysis speeds.
Last but not least, the development of unsupervised deep learning strategies for the reliable and fast measurement of \glspl{PSD} are of utmost interest, because they would lift the requirement of a known ground truth and thereby the necessity of manual annotation or image synthesis. 
	% !TeX spellcheck = en_US
\section*{Acknowledgment}
The authors gratefully acknowledge the support via the project \emph{"20226 N -- Deep Learning Particle Detection"} of the \emph{DECHEMA e.V.} research foundation,
which was funded by the \emph{German Federation of Industrial Research Associations (AiF)} within the program for
\emph{Industrial Corporate Research (IGF)} of the \emph{Federal Ministry
for Economic Affairs and Energy (BMWi)} on the basis of a
decision by the \emph{German Bundestag}. All authors declare that they have no competing interests.\newline

Special thank goes to Dennis Kiesler\footnote{\url{https://orcid.org/0000-0002-2219-9773}} for many helpful suggestions as well as his help with respect to \gls{SEM} analyses and the manual annotation of the test samples.\newline

\setlength{\intextsep}{0pt}
\setlength{\columnsep}{1em}

\setlength{\imagewidth}{3\baselineskip}
\begin{wrapfigure}[4]{l}{\imagewidth}
	\vspace{0.25\baselineskip}
	\includegraphics[width=\imagewidth]{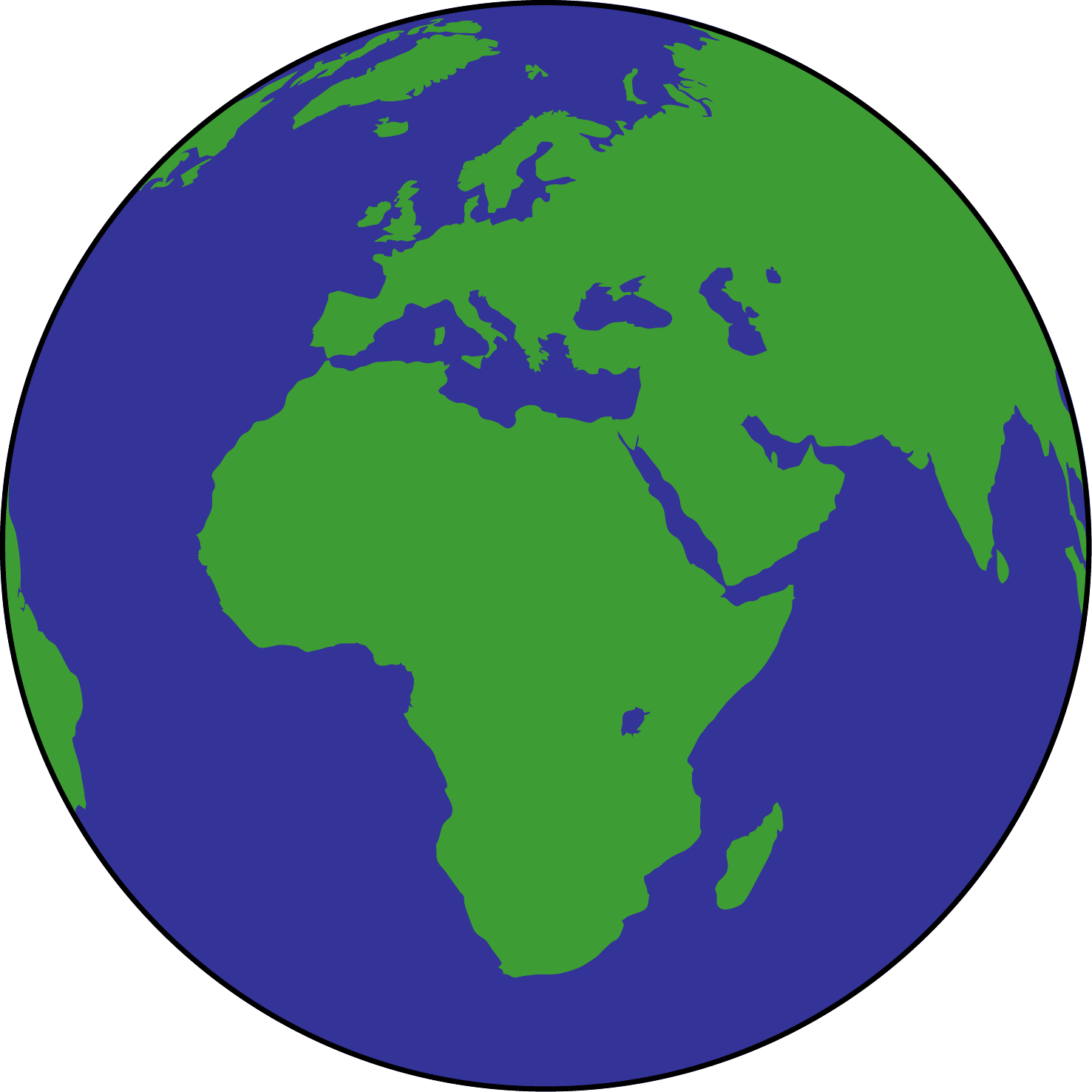}
\end{wrapfigure}

The carbon dioxide, which was emitted due to the training of \glsentrylongpl{ANN} for this publication was fully compensated on behalf of the lead author, by donating to \emph{atmosfair gGmbH}.\footnote{\url{https://www.atmosfair.de/en/}}\textsuperscript{,}\footnote{Neither the authors nor the institutions they represent have any affiliation with \emph{atmosfair gGmbH}.}
	
	% List of Abbreviatons
	\newpage
	\setlength{\glsdescwidth}{67mm}
	\printglossary[type=\acronymtype]
	
	% List of Symbols
	\setlength{\glsdescwidth}{72.75mm}
	\printglossary[type=symbolslist]
	
	% Bibliography
	\newpage
	\bibliographystyle{elsarticle-num}
	\bibliography{bibliography}

\begin{thebibliography}{10}
\expandafter\ifx\csname url\endcsname\relax
  \def\url#1{\texttt{#1}}\fi
\expandafter\ifx\csname urlprefix\endcsname\relax\def\urlprefix{URL }\fi
\expandafter\ifx\csname href\endcsname\relax
  \def\href#1#2{#2} \def\path#1{#1}\fi

\bibitem{Schulze.2014}
D.~Schulze, {Pulver und Schüttgüter: Fließeigenschaften und Handhabung}, 3rd
  Edition, {VDI-Buch}, {Springer Berlin Heidelberg}, Berlin, Heidelberg, 2014.

\bibitem{Schneider.2015}
I.~L. Schneider, E.~C. Teixeira, L.~F. {Silva Oliveira}, F.~Wiegand,
  {Atmospheric particle number concentration and size distribution in a
  traffic–impacted area}, {Atmospheric Pollution Research} 6~(5) (2015)
  877--885.
\newblock \href {https://doi.org/10.5094/APR.2015.097}
  {\path{doi:10.5094/APR.2015.097}}.

\bibitem{Rodrigues.2003}
R.~T. Rodrigues, J.~Rubio, {New basis for measuring the size distribution of
  bubbles}, {Minerals Engineering} 16~(8) (2003) 757--765.
\newblock \href {https://doi.org/10.1016/S0892-6875(03)00181-X}
  {\path{doi:10.1016/S0892-6875(03)00181-X}}.

\bibitem{Azari.2014}
M.~R. Azari, A.~Yazdian, R.~Zendehdel, H.~Souri, S.~Khodakarim, H.~Peirovi,
  D.~Panahi, M.~Kazempour, {Improved Method for Analysis of Airborne Asbestos
  Fibers Using Phase Contrast Microscopy and FTIR Spectrometry}, {Tanaffos}
  13~(3) (2014) 38--45.

\bibitem{Cheng.2009}
J.~Cheng, J.~C. Rajapakse, {Segmentation of clustered nuclei with shape markers
  and marking function}, {IEEE transactions on bio-medical engineering} 56~(3)
  (2009) 741--748.
\newblock \href {https://doi.org/10.1109/TBME.2008.2008635}
  {\path{doi:10.1109/TBME.2008.2008635}}.

\bibitem{Jung.2010}
C.~Jung, C.~Kim, {Segmenting clustered nuclei using H-minima transform-based
  marker extraction and contour parameterization}, {IEEE transactions on
  bio-medical engineering} 57~(10 PART 2) (2010) 2600--2604.
\newblock \href {https://doi.org/10.1109/TBME.2010.2060336}
  {\path{doi:10.1109/TBME.2010.2060336}}.

\bibitem{Shu.2013}
J.~Shu, H.~Fu, G.~Qiu, P.~Kaye, M.~Ilyas, {Segmenting overlapping cell nuclei
  in digital histopathology images}, in: {2013 35th annual international
  conference of the IEEE Engineering in Medicine and Biology Society (EMBC)},
  IEEE, Piscataway, NJ, 2013, pp. 5445--5448.
\newblock \href {https://doi.org/10.1109/EMBC.2013.6610781}
  {\path{doi:10.1109/EMBC.2013.6610781}}.

\bibitem{Park.2013}
C.~Park, J.~Z. Huang, J.~X. Ji, Y.~Ding, {Segmentation, Inference and
  Classification of Partially Overlapping Nanoparticles}, {IEEE transactions on
  pattern analysis and machine intelligence} 35~(3) (2013) 669--681.
\newblock \href {https://doi.org/10.1109/TPAMI.2012.163}
  {\path{doi:10.1109/TPAMI.2012.163}}.

\bibitem{Wang.2016}
Z.~Wang, {A semi-automatic method for robust and efficient identification of
  neighboring muscle cells}, {Pattern Recognition} 53 (2016) 300--312.
\newblock \href {https://doi.org/10.1016/j.patcog.2015.12.009}
  {\path{doi:10.1016/j.patcog.2015.12.009}}.

\bibitem{Kruis.1994}
F.~E. Kruis, J.~van Denderen, H.~Buurman, B.~Scarlett, {Characterization of
  Agglomerated and Aggregated Aerosol Particles Using Image Analysis},
  {Particle {\&} Particle Systems Characterization} 11~(6) (1994) 426--435.
\newblock \href {https://doi.org/10.1002/ppsc.19940110605}
  {\path{doi:10.1002/ppsc.19940110605}}.

\bibitem{Ballard.1981}
D.~Ballard, {Generalizing the Hough transform to detect arbitrary shapes},
  {Pattern Recognition} 13~(2) (1981) 111--122.
\newblock \href {https://doi.org/10.1016/0031-3203(81)90009-1}
  {\path{doi:10.1016/0031-3203(81)90009-1}}.

\bibitem{Merlin.1975}
P.~M. Merlin, D.~J. Farber, {A Parallel Mechanism for Detecting Curves in
  Pictures}, {IEEE Transactions on Computers} C-24~(1) (1975) 96--98.
\newblock \href {https://doi.org/10.1109/T-C.1975.224087}
  {\path{doi:10.1109/T-C.1975.224087}}.

\bibitem{Xu.1990}
L.~Xu, E.~Oja, P.~Kultanen, {A new curve detection method: Randomized Hough
  transform (RHT)}, {Pattern Recognition Letters} 11~(5) (1990) 331--338.
\newblock \href {https://doi.org/10.1016/0167-8655(90)90042-Z}
  {\path{doi:10.1016/0167-8655(90)90042-Z}}.

\bibitem{Frei.2018}
M.~Frei, F.~E. Kruis, {Fully automated primary particle size analysis of
  agglomerates on transmission electron microscopy images via artificial neural
  networks}, {Powder Technology} 332 (2018) 120--130.
\newblock \href {https://doi.org/10.1016/j.powtec.2018.03.032}
  {\path{doi:10.1016/j.powtec.2018.03.032}}.

\bibitem{Hamzeloo.2014}
E.~Hamzeloo, M.~Massinaei, N.~Mehrshad, {Estimation of particle size
  distribution on an industrial conveyor belt using image analysis and neural
  networks}, {Powder Technology} 261 (2014) 185--190.
\newblock \href {https://doi.org/10.1016/j.powtec.2014.04.038}
  {\path{doi:10.1016/j.powtec.2014.04.038}}.

\bibitem{Ko.2011}
Y.-D. Ko, H.~Shang, {A neural network-based soft sensor for particle size
  distribution using image analysis}, {Powder Technology} 212~(2) (2011)
  359--366.
\newblock \href {https://doi.org/10.1016/j.powtec.2011.06.013}
  {\path{doi:10.1016/j.powtec.2011.06.013}}.

\bibitem{Goodfellow.2016}
I.~Goodfellow, Y.~Bengio, A.~Courville, {Deep learning}, {Adaptive computation
  and machine learning}, 2016.

\bibitem{Mehle.2017}
A.~Mehle, B.~Likar, D.~Tomaževič, {In-line recognition of agglomerated
  pharmaceutical pellets with density-based clustering and convolutional neural
  network}, {IPSJ Transactions on Computer Vision and Applications} 9~(1)
  (2017) 7.
\newblock \href {https://doi.org/10.1186/s41074-017-0019-2}
  {\path{doi:10.1186/s41074-017-0019-2}}.

\bibitem{Heimowitz.2018}
A.~Heimowitz, J.~Andén, A.~Singer,
  \href{http://arxiv.org/pdf/1802.00469v2}{{APPLE Picker: Automatic Particle
  Picking, a Low-Effort Cryo-EM Framework}}.
\newline\urlprefix\url{http://arxiv.org/pdf/1802.00469v2}

\bibitem{Kriesel.2007}
D.~Kriesel, \href{http://www.dkriesel.com}{{A Brief Introduction to Neural
  Networks}} (2007).
\newline\urlprefix\url{http://www.dkriesel.com}

\bibitem{He.2017}
K.~He, G.~Gkioxari, P.~Dollár, R.~Girshick,
  \href{http://arxiv.org/pdf/1703.06870v3}{{Mask R-CNN}}.
\newline\urlprefix\url{http://arxiv.org/pdf/1703.06870v3}

\bibitem{Gonzalez.2004}
R.~C. González, R.~E. Woods, S.~L. Eddins, {Digital image processing: Using
  MATLAB}, {Prentice Hall}, Upper Saddle River, NJ, 2004.

\bibitem{Lee.2011}
H.~Lee, R.~Grosse, R.~Ranganath, A.~Y. Ng, {Unsupervised learning of
  hierarchical representations with convolutional deep belief networks},
  {Communications of the ACM} 54~(10) (2011) 95.
\newblock \href {https://doi.org/10.1145/2001269.2001295}
  {\path{doi:10.1145/2001269.2001295}}.

\bibitem{Deng.2013}
L.~Deng, {Deep Learning: Methods and Applications}, {FNT in Signal Processing
  (Foundations and Trends in Signal Processing)} 7~(3-4) (2013) 197--387.
\newblock \href {https://doi.org/10.1561/2000000039}
  {\path{doi:10.1561/2000000039}}.

\bibitem{Hui.2018}
J.~Hui,
  \href{https://medium.com/@jonathan_hui/image-segmentation-with-mask-r-cnn-ebe6d793272}{{Image
  segmentation with Mask R-CNN}} (2018).
\newline\urlprefix\url{https://medium.com/@jonathan_hui/image-segmentation-with-mask-r-cnn-ebe6d793272}

\bibitem{Babick.2018}
F.~Babick, L.~Hillemann, M.~Stintz, T.~Dillenburger, M.~Pitz, A.~Hellmann,
  S.~Antonyuk, S.~Ripperger, F.~J.~T. Huber, S.~Will, R.~Wernet,
  M.~Seipenbusch, M.~Gensch, A.~Weber, D.~Kiesler, E.~Kruis, R.~Friehmelt,
  B.~Sachweh,
  \href{https://onlinelibrary.wiley.com/doi/pdf/10.1002/cite.201700094}{{Multiparameter
  Characterization of Aerosols}}, {Chemie Ingenieur Technik} 90~(7) (2018)
  923--936.
\newblock \href {https://doi.org/10.1002/cite.201700094}
  {\path{doi:10.1002/cite.201700094}}.
\newline\urlprefix\url{https://onlinelibrary.wiley.com/doi/pdf/10.1002/cite.201700094}

\bibitem{MacKay.2016}
D.~J.~C. MacKay, {Information theory, inference, and learning algorithms}, 17th
  Edition, 2016.

\bibitem{Abdulla.2017}
W.~Abdulla, {Mask R-CNN for object detection and instance segmentation on Keras
  and TensorFlow} (2017).

\bibitem{Chollet.2015}
F.~Chollet, et~al., {Keras} (2015).

\bibitem{Abadi.2015}
M.~Abadi, A.~Agarwal, P.~Barham, E.~Brevdo, Z.~Chen, C.~Citro, G.~S. Corrado,
  A.~Davis, J.~Dean, M.~Devin, S.~Ghemawat, I.~Goodfellow, A.~Harp, G.~Irving,
  M.~Isard, Y.~Jia, R.~Jozefowicz, L.~Kaiser, M.~Kudlur, J.~Levenberg,
  D.~Mané, R.~Monga, S.~Moore, D.~Murray, C.~Olah, M.~Schuster, J.~Shlens,
  B.~Steiner, I.~Sutskever, K.~Talwar, P.~Tucker, V.~Vanhoucke, V.~Vasudevan,
  F.~Viégas, O.~Vinyals, P.~Warden, M.~Wattenberg, M.~Wicke, Y.~Yu, X.~Zheng,
  \href{https://www.tensorflow.org/}{{TensorFlow: Large-Scale Machine Learning
  on Heterogeneous Systems}} (2015).
\newline\urlprefix\url{https://www.tensorflow.org/}

\bibitem{Python.2019}
{Python Software Foundation}, \href{http://www.python.org/}{{Python 3.5.2}}.
\newline\urlprefix\url{http://www.python.org/}

\bibitem{He.2015}
K.~He, X.~Zhang, S.~Ren, J.~Sun, \href{http://arxiv.org/pdf/1512.03385v1}{{Deep
  Residual Learning for Image Recognition}}.
\newline\urlprefix\url{http://arxiv.org/pdf/1512.03385v1}

\bibitem{Canziani.2016}
A.~Canziani, A.~Paszke, E.~Culurciello,
  \href{http://arxiv.org/pdf/1605.07678v4}{{An Analysis of Deep Neural Network
  Models for Practical Applications}}.
\newline\urlprefix\url{http://arxiv.org/pdf/1605.07678v4}

\bibitem{Russakovsky.2014}
O.~Russakovsky, J.~Deng, H.~Su, J.~Krause, S.~Satheesh, S.~Ma, Z.~Huang,
  A.~Karpathy, A.~Khosla, M.~Bernstein, A.~C. Berg, L.~Fei-Fei,
  \href{http://arxiv.org/pdf/1409.0575v3}{{ImageNet Large Scale Visual
  Recognition Challenge}}.
\newline\urlprefix\url{http://arxiv.org/pdf/1409.0575v3}

\bibitem{Lin.2014}
T.-Y. Lin, M.~Maire, S.~Belongie, L.~Bourdev, R.~Girshick, J.~Hays, P.~Perona,
  D.~Ramanan, C.~L. Zitnick, P.~Dollár,
  \href{http://arxiv.org/pdf/1405.0312v3}{{Microsoft COCO: Common Objects in
  Context}}.
\newline\urlprefix\url{http://arxiv.org/pdf/1405.0312v3}

\bibitem{Smith.2015}
L.~N. Smith, \href{http://arxiv.org/pdf/1506.01186v6}{{Cyclical Learning Rates
  for Training Neural Networks}}.
\newline\urlprefix\url{http://arxiv.org/pdf/1506.01186v6}

\bibitem{Kenstler.2017}
B.~Kenstler, {Cyclical Learning Rate (CLR)} (2017).

\bibitem{MATLAB.2018b}
\href{https://www.mathworks.com/help/images/ref/imfindcircles.html}{{MATLAB and
  Image Processing Toolbox: Function: imfindcircles}} (2018).
\newline\urlprefix\url{https://www.mathworks.com/help/images/ref/imfindcircles.html}

\bibitem{Schneider.2012}
C.~A. Schneider, W.~S. Rasband, K.~W. Eliceiri, {NIH Image to ImageJ: 25 years
  of image analysis}, {Nature methods} 9~(7) (2012) 671--675.
\newblock \href {https://doi.org/10.1038/nmeth.2089}
  {\path{doi:10.1038/nmeth.2089}}.

\bibitem{Wagner.2017}
{Thorsten Wagner}, {Jan Eglinger}, {Thorstenwagner/Ij-Particlesizer: V1.0.9
  Snapshot Release} (2017).
\newblock \href {https://doi.org/10.5281/zenodo.820296}
  {\path{doi:10.5281/zenodo.820296}}.

\bibitem{MATLAB.2018a}
\href{https://www.mathworks.com/help/images/ref/regionprops.html}{{MATLAB and
  Image Processing Toolbox: Function: regionprops}} (2018).
\newline\urlprefix\url{https://www.mathworks.com/help/images/ref/regionprops.html}

\bibitem{COCOConsortium.2019}
\href{http://cocodataset.org/#detection-eval}{{COCO - Common Objects in
  Context: Metrics}} (2019).
\newline\urlprefix\url{http://cocodataset.org/#detection-eval}

\bibitem{Hyndman.2006}
R.~J. Hyndman, A.~B. Koehler, {Another look at measures of forecast accuracy},
  {International Journal of Forecasting} 22~(4) (2006) 679--688.
\newblock \href {https://doi.org/10.1016/j.ijforecast.2006.03.001}
  {\path{doi:10.1016/j.ijforecast.2006.03.001}}.

\bibitem{Murphy.2012}
K.~P. Murphy, {Machine learning: A probabilistic perspective}, {Adaptive
  computation and machine learning series}, 2012.

\end{thebibliography}
	 
	%%%%%%%%%%%%%%%%%%%%%%%%%%%%%%%%%%%%%%%%%%%%%%%%%%%%%%
	% Appendix
	%%%%%%%%%%%%%%%%%%%%%%%%%%%%%%%%%%%%%%%%%%%%%%%%%%%%%%
	\beginappendix
	% !TeX spellcheck = en_US
\section*{Appendix}

\begin{table}
	\centering
	\caption{Relevant hardware of the utilized \gls{GPU} server.}	
	\label{app:tab:Hardware}
	\begin{tabularx}{\figurewidth}{Yc}
		\toprule
		\toprule
		Mainboard & Supermicro X11DPG-QT\\
		CPU & 2 x Intel Xeon Gold 5118\\
		GPU & 4 x NVIDIA GeForce RTX 2080 Ti\\
		RAM & 12 x \SI{8}{\giga\byte} DDR4 PC2666 ECC reg.\\
		SSD (OS) & Micron SSD 5100 PRO \SI{960}{\giga\byte}, SATA\\
		SSD (data) & Samsung SSD 960 EVO \SI{1}{\tera\byte}, M.2\\
		\bottomrule
	\end{tabularx}
\end{table}%\\

\setlength{\belowcaptionskip}{9ex}

\begin{table}
	\centering
	\caption{Relevant software of the utilized \gls{GPU} server.}	
	\label{app:tab:Software}
	\begin{tabularx}{\figurewidth}{YZ}
		\toprule
		\toprule
		OS (host) & Ubuntu 18.04.2 LTS\\
		OS (docker) & Ubuntu 16.04.5 LTS\\
		Python (docker) & 3.5.2\\
		Tensorflow (docker) & 1.13.1\\
		\bottomrule
	\end{tabularx}
\end{table}%

\begin{table}
	\centering
	\caption{Possible image augmentations.}
	\label{app:tab:Augmentation}
	\begin{tabularx}{\figurewidth}{YZ}
		\toprule
		\toprule
		Flip: left-right& \SI{50}{\percent}\\
		Flip: up-down& \SI{50}{\percent}\\
		Rotation& \ang{90}, \ang{180} or \ang{270}\\
		Multiplication& \numrange{0.8}{1.5}\\
		Gaussian Blur & $\gls{symb:standard_deviation}=\text{\numrange{0}{0.5}}$\\
		\bottomrule
	\end{tabularx}
\end{table}%

\begin{table}
	\centering
	\caption{Final training conditions.}
	\label{app:tab:FinalTrainingConditions}
	\begin{tabularx}{\figurewidth}{Yc}
		\toprule
		\toprule
		Backbone & ResNet-50 \\
		Initial Weights & \gls{COCO} \\
		\midrule
		Number of Training Samples & 400\\
		Number of Validation Samples & 100\\
		Image Augmentation & \numrange{0}{2} of \cref{app:tab:Augmentation}\\
		Batch Size & 4\\
		Iterations per Epoch & 100\\
		\midrule
		Optimizer & \gls{SGDM} \cite{Murphy.2012}\\
		Learning Rate Cycle Policy & triangular\\
		Minimum Learning Rate & 0.0005\\
		Maximum Learning Rate & 0.0037\\
		Learning Rate Cycle Length & 4 epochs\\
		Momentum & 0.9\\
		Weight Decay & $10^{-4}$\\
		Early Stopping Patience & 20 epochs\\
		\bottomrule
	\end{tabularx}
\end{table}%

\end{document}